\documentclass[iicol]{sn-jnl}

\usepackage{apacite}
\usepackage{colortbl}
\usepackage{xspace}
\usepackage{graphicx}%
\usepackage{multirow}%
\usepackage{amsmath,amssymb,amsfonts}%
\usepackage{amsthm}%
\usepackage{mathrsfs}%
\usepackage[title]{appendix}%
\usepackage{xcolor}%
\usepackage{textcomp}%
\usepackage{manyfoot}%
\usepackage{booktabs}%
\usepackage{algorithm}%
\usepackage{algorithmicx}%
\usepackage{algpseudocode}%
\usepackage{listings}%

\usepackage[capitalize]{cleveref}
\crefname{section}{Sec.}{Secs.}
\Crefname{section}{Section}{Sections}
\Crefname{table}{Table}{Tables}
\crefname{table}{Tab.}{Tabs.}

\newcommand{\subheading}[1]{\textbf{#1}.}

\newcolumntype{a}{>{\columncolor{verylightgray}}c}
\newcommand{\rowNormal}{$\;\;$}
\newcommand{\rowMarked}{\rowcolor{verylightgray}\rowNormal}

\definecolor{verylightgray}{HTML}{E0E0E0}

\renewcommand{\eqref}[1]{Eq.~(\ref{#1})}

\newcommand{\papertitle}{3D Multi-Pigeon Pose Estimation and Tracking\xspace}
\newcommand{\paperabr}{3D-MuPPET\xspace}
\newcommand{\dataabr}{Wild-MuPPET\xspace}

\begin{document}

\title[\paperabr: \papertitle]{\paperabr: \papertitle}


\author*[1,2]{\fnm{Urs} \sur{Waldmann}}\email{urs.waldmann@uni-konstanz.de}
\equalcont{These authors contributed equally to this work.}

\author*[2,3,4]{\fnm{Alex Hoi Hang} \sur{Chan}}\email{hoi-hang.chan@uni-konstanz.de}
\equalcont{These authors contributed equally to this work.}

\author[2,4,5]{\fnm{Hemal} \sur{Naik}}\email{hnaik@ab.mpg.de}

\author[2,3,4,6,7]{\fnm{M\'{a}t\'{e}} \sur{Nagy}}\email{nagymate@hal.elte.hu}

\author[2,3,4]{\fnm{Iain D.} \sur{Couzin}}\email{icouzin@ab.mpg.de}

\author[1,2]{\fnm{Oliver} \sur{Deussen}}\email{Oliver.Deussen@uni-konstanz.de}

\author[1,2]{\fnm{Bastian} \sur{Goldluecke}}\email{bastian.goldluecke@uni-konstanz.de}

\author[2,3,4]{\fnm{Fumihiro} \sur{Kano}}\email{fumihiro.kano@uni-konstanz.de}

\affil[1]{\orgdiv{Department of Computer and Information Science}, \orgname{University of Konstanz}, \orgaddress{\country{Germany}}}

\affil[2]{\orgdiv{Centre for the Advanced Study of Collective Behaviour}, \orgname{University of Konstanz}, \orgaddress{\country{Germany}}}

\affil[3]{\orgdiv{Department of Collective Behavior}, \orgname{Max Planck Institute of Animal Behavior}, \orgaddress{\city{Konstanz}, \country{Germany}}}

\affil[4]{\orgdiv{Department of Biology}, \orgname{University of Konstanz}, \orgaddress{\country{Germany}}}

\affil[5]{\orgdiv{Department of Ecology of Animal Societies}, \orgname{Max Planck Institute of Animal Behavior}, \orgaddress{\city{Konstanz}, \country{Germany}}}

\affil[6]{\orgdiv{Department of Biological Physics}, \orgname{E\"otv\"os Lor\'and University}, \orgaddress{\city{Budapest}, \country{Hungary}}}

\affil[7]{\orgdiv{MTA-ELTE ‘Lend\"ulet’ Collective Behaviour Research Group}, \orgname{Hungarian Academy of Sciences}, \orgaddress{\city{Budapest}, \country{Hungary}}}

\abstract{
Markerless methods for animal posture tracking have been rapidly developing recently, but frameworks and benchmarks for tracking large animal groups in 3D are still lacking.
To overcome this gap in the literature, we present~\paperabr, a framework to estimate and track 3D poses of up to 10 pigeons at interactive speed using multiple camera views. 
We train a pose estimator to infer 2D keypoints and bounding boxes of multiple pigeons, then triangulate the keypoints to 3D.
For identity matching of individuals in all views, we first dynamically match 2D detections to global identities in the first frame, then use a 2D tracker to maintain IDs across views in subsequent frames.
We achieve comparable accuracy to a state of the art 3D pose estimator in terms of median error and Percentage of Correct Keypoints.
Additionally, we benchmark the inference speed of~\paperabr, with up to 9.45 fps in 2D and 1.89 fps in 3D, and perform quantitative tracking evaluation, which yields encouraging results.
Finally, we showcase two novel applications for 3D-MuPPET.
First, we train a model with data of single pigeons and achieve comparable results in 2D and 3D posture estimation for up to 5 pigeons.
Second, we show that~\paperabr also works in outdoors without additional annotations from natural environments.
Both use cases simplify the domain shift to new species and environments, largely reducing annotation effort needed for 3D posture tracking.
To the best of our knowledge we are the first to present a framework for 2D/3D animal posture and trajectory tracking that works in both indoor and outdoor environments for up to 10 individuals.
We hope that the framework can open up new opportunities in studying animal collective behaviour and encourages further developments in 3D multi-animal posture tracking.
}




\maketitle

\begin{figure*}[!ht]
  \centering
  \includegraphics[width=0.49\linewidth]{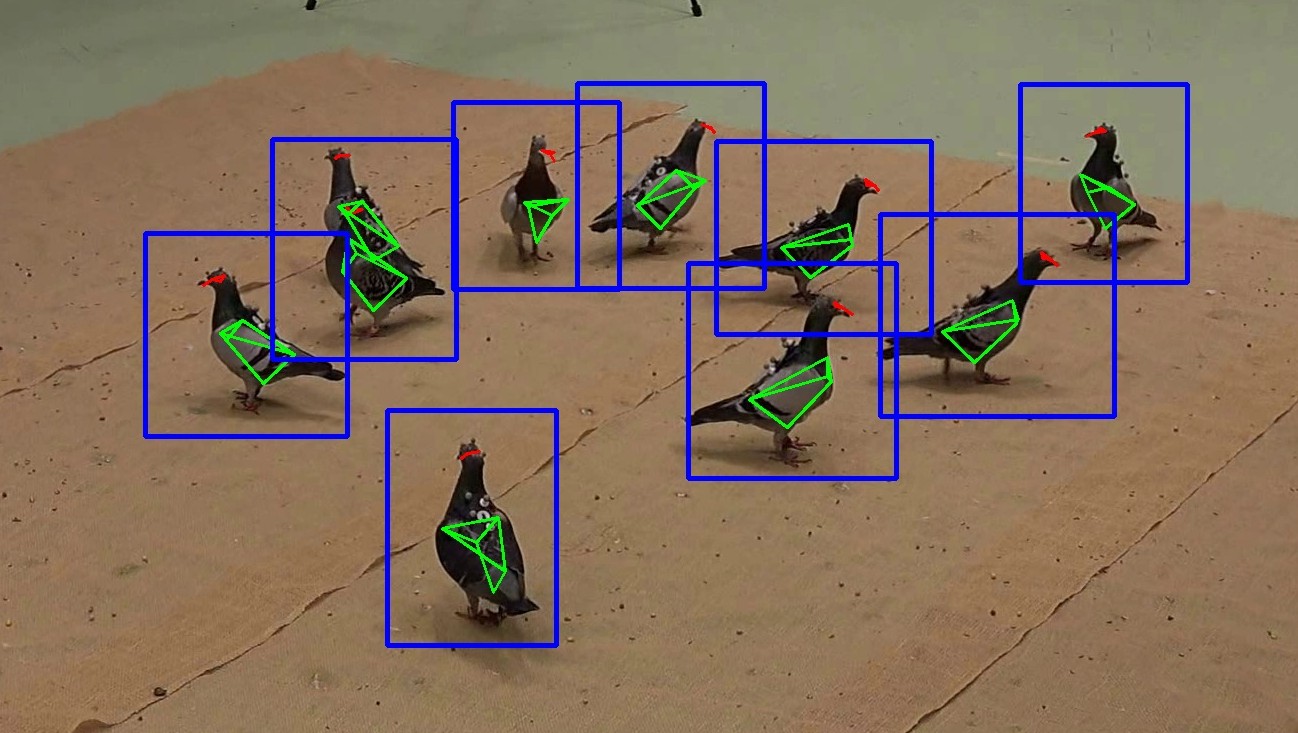}
  \includegraphics[width=0.49\linewidth]{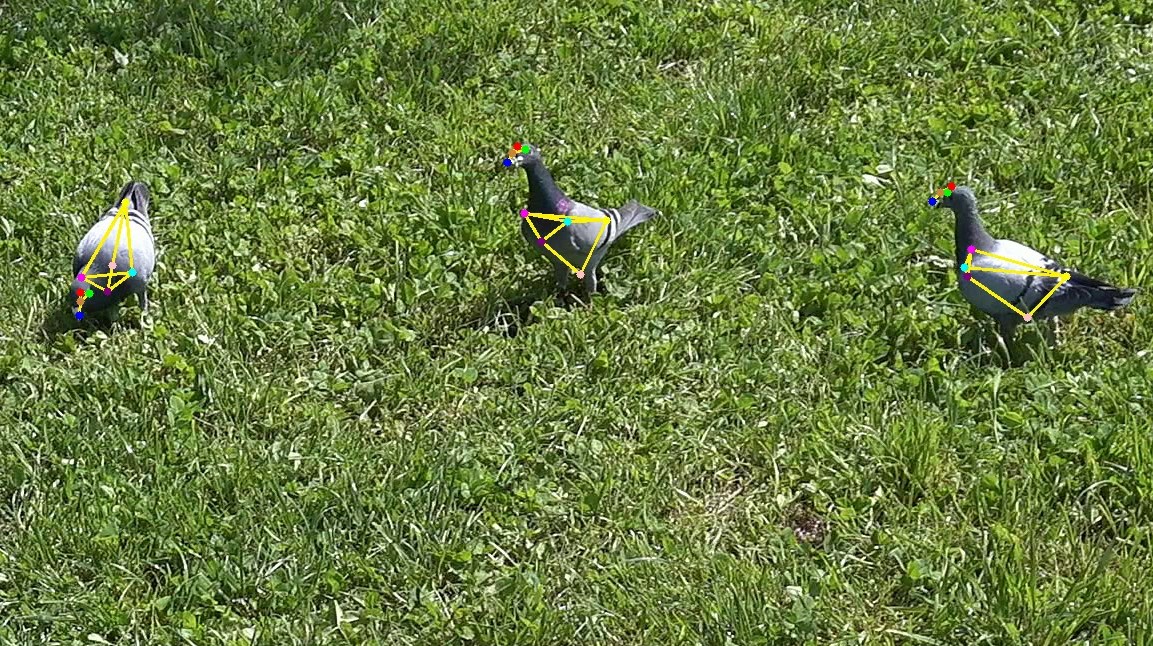}
  \caption[\papertitle (\paperabr)]{\emph{\papertitle (\paperabr)} is a framework for multi-animal pose estimation and tracking for lab (left) and outdoor data (right).
  \textit{Left}: Estimated complex pose (beak, nose, left and right eye, left and right shoulder, top and bottom keel and tail) of pigeons recorded in a captive environment.
  \textit{Right}: The image shows an example with three pigeons recorded outdoors with estimated 3D keypoints reprojected to camera view (colored dots).
  }
  \label{fig:teaser}
\end{figure*}
%

\section{Introduction}
\label{sec:introduction}
Pose estimation and tracking are among the fundamental problems in computer vision and a crucial task in many visual tracking applications ranging from sports in humans~\cite{bridgeman2019multi-person} to the study of collective behaviour in nonhuman animals~\cite{couzin2022emerging,koger2023quantifying}.
For the latter, accurate quantification of behavior is critical to understand the underlying principles of social interaction and the neural and cognitive underpinnings of animal behaviour~\cite{bernshtein1967co,ObservationalStudyofBehaviorSamplingMethods,Berman2018,DLC,Kays2015te}.
While researchers conventionally analyzed animal behaviour manually using a predefined catalogue of behaviours using ethograms, recent advances in computer vision, as well as the increasing demands for large datasets involving the analysis of the fine-scaled and rapidly-changing behaviours of animals, encouraged the development of automated tracking methods \cite{DELL2014417,gomez2014bi,ANDERSON201418,DLC}.
In such applications, multi-object pose estimation is essential to observe the dynamics of socially interacting animals because individuals in a group tend to be partially occluded.
Notably, with the recent advances in hardware and computer vision, marker-based motion capture systems have enabled posture tracking of single and multiple animals in controlled captive environments~\cite{SMART-BARN,kano2022head,itahara2022corvid,minano2023through,Itahara2023}.
Such marker-based motion capture systems also facilitated the curation of large-scale animal posture datasets~\cite{3D-POP,marshall2021pair} to develop markerless methods for posture tracking of single~\cite{DLC,LEAP,DANNCE,DPK} and multiple animals~\cite{maDLC,SLEAP,I-MuPPET}.
A crucial advantage of markerless over marker-based methods is that individuals do not have to be equipped with markers, thus opening possibilities for posture tracking and behavioural quantification of unhabituated animals even in the wild (i.e., natural habitat).
Recently, with the success of 2D single animal markerless pose estimation methods like LEAP~\cite{LEAP} and DeepLabCut (DLC,~\citeA{DLC}), this research area has received increased attention in method development for 2D tracking multiple animals~\cite{maDLC,SLEAP,DPK,I-MuPPET} and 3D postures ~\cite{DeepFly3D,AcinoSet,DANNCE,NePu,han2023social}.
This recent progress in markerless pose estimation also boosted the research area of computer vision for animals, as exemplified by the fact that the CVPR workshop on ``Computer Vision for Animal Behavior Tracking and Modeling''~\cite{CV4An2023} has been taking place every year since 2021.
Topics of this workshop range from object detection \cite{Duporge2021us}, behavior analysis \cite{EthoLoop,bolanos2021a3}, object segmentation~\cite{Chen2020si,unlabprop}, 3D shape and pose fitting \cite{CreaturesGreatAndSMAL,3DBirdReconstruction} to pose estimation \cite{labuguen2021ma,gosztolai2021li,I-MuPPET} and tracking \cite{Paperidtrackerai,Pedersen20203d,I-MuPPET}.

Despite recent progress in the field of computer vision for animals, reliable tracking of multiple moving animals in real-time and estimating their 3D pose to measure behaviours in a group remain an open challenge.
While frameworks for multi-animal pose estimation and tracking in 2D~\cite{maDLC,SLEAP,I-MuPPET} are common, frameworks for 3D multi-animal pose estimation are generally lacking, with a few notable exceptions. 
We are aware of only three frameworks that estimate the 3D pose of more than one individual (two macaques~\cite{OpenMonkeyStudio}, two rats/parrots~\cite{han2023social}, and four/two pigs/dogs~\cite{MAMMAL}) in controlled captive environments, and finally one framework~\cite{AcinoSet,3DDLC} that estimates 3D poses of single Cheetahs in the wild.
%

One limiting factor for the development of animal pose estimation methods is the limited amount of annotated data as ground truth for training and evaluation, especially compared to human datasets (for example $3.6$ million in Human 3.6M~\cite{H36M}), cf. also~\citeA{Sanakoyeu_2020_CVPR}.
Using birds as an example, we are aware of only four datasets for birds across different bird species~\cite{CaltechBirdDataset,NAbirds,3DBirdReconstruction,3D-POP}. 
The lack of annotated datasets not only limits the ability to do thorough quantitative evaluation for new proposed methods, but biologists who want to make use of these methods also require a large amount of laborious manual annotations.
DeepLabCut~\cite{DLC}, LEAP~\cite{LEAP} and DeepPoseKit~\cite{DPK} overcome this lack of training data using a human in loop approach where a small manually labelled dataset is used to train a neural network, then predict body parts (pre-labeling) of previously unlabeled material to generate larger training datasets.
Creatures Great and SMAL~\cite{CreaturesGreatAndSMAL} instead creates synthetic silhouettes for training and extracts
silhouettes with~\citeA{wang2015se,wang2015jo} from real data for inference.
Hence, one way to circumvent the lack of available annotated large-scale datasets for many animal species is to develop methods that exploit few training data in an efficient way.
However, the drawback of this approach is that these methods cannot be evaluated quantitatively in detail due to the few annotated data that they leverage.

We choose pigeons as an example species not only because it is a common model species for animal collective behaviour (e.g \citeA{yomosa2015coordinated,nagy2010hierarchical,nagy2013context,papadopoulou2022self,sasaki2017cumulative}), but also because of the recent introduction of a large scale multi-animal 2D/3D posture dataset in 3D-POP \cite{3D-POP}. This dataset opens up possibilities to propose and benchmark methods for 3D posture estimation and tracking due to its size. Here, we extend I-MuPPET~\cite{I-MuPPET}, a recent framework proposed for interactive 2D posture estimation and tracking of multiple pigeons, by incoporating multiple views to obtain 3D information. We will first evaluate the 2D framework proposed in I-MuPPET \cite{I-MuPPET} on the 3D-POP dataset, then introduce and evaluate our extension to 3D. We also highlight the applicability of the framework to data recorded in outdoor settings without any further annotations.

\subheading{Contributions}
In this paper, we present~\paperabr, a flexible framework for interactive tracking and 3D pose estimation of multiple pigeons that works for data recorded both in captivity and the wild.
The framework is based on triangulating 2D poses from multiple views to 3D, allowing 3D reconstruction if a 2D posture estimation model and a multi-view setup is available.
Compared to a state of the art 3D pose estimation method (Learnable Triangulation of Human Pose; LToHP,~\citeA{LToHP}) that requires
ground truth in 3D for training, \paperabr is less accurate (Root Mean Square Error; RMSE of $24.0$ mm vs. $14.8$ mm, and Percentage of Correct Keypoints; PCK05 of $71.0\%$ vs. $76.7\%$ for ours and LToHP respectively), but comparable in terms of median error ($7.0$ mm vs. $5.8$ mm for LToHP) and Percentage of Correct Keypoints (PCK10 of $92.5\%$ vs. $94.3\%$ for LToHP).
We track up to ten pigeons (the upper limit in~\citeA{3D-POP}) with up to $9.45$ fps in 2D and $1.89$ fps in 3D, and report detailed results for speed and accuracy.
Additionally, we highlight two use cases that showcases the flexibility of our framework.
\begin{enumerate}
    \item  We demonstrate that it is possible to train on an annotated dataset containing only a single pigeon to predict keypoints of a complex pose for multiple pigeons in a stable and accurate way. 
    \item We demonstrate the ability to estimate 3D poses of pigeons recorded outdoors, cf.~\cref{fig:teaser}, without any additional annotations.
\end{enumerate}
Both applications provide alternatives for the domain shift to other species or applications in the wild by reducing annotation effort required for multi-animal posture estimation.

Finally, to evaluate pose estimation from data recorded outdoors, we also present~\dataabr, a novel 3D posture dataset of~$500$ manually annotated frames from $4$ camera views collected in the wild.

To the best of our knowledge, we are the first to present a markerless 2D and 3D animal pose estimation framework for more than four individuals.
Our approach is also not limited to pigeons and can be applied to any other species, given 2D posture annotations and a calibrated multi-camera system are available.
In our supplemental material we also showcase the applicability to other species like mice from~\citeA{DLC} and cowbirds from~\citeA{3DBirdReconstruction} where 2D posture annotations from one camera view are available.
The source code and data to reproduce the results of this paper are publicly available at~\url{https://alexhang212.github.io/3D-MuPPET/}.
We think that~\paperabr offers a promising framework for automated 3D multi-animal pose estimation and identity tracking, opening up new ways for biologists to study animal collective behaviour in a fine-scaled way.
%

\section{Related Work}
\label{sec:realtedwork}
In this section, we explore existing work on both 2D and 3D posture estimation and multi-animal tracking, since \paperabr makes use of 2D detections and triangulation for 3D poses. We identify existing methods, then major gaps that we hope \paperabr can fill.

\subsection{Animal Pose Estimation}
\subheading{2D Single Animal Pose Estimation}
With the success of DeepLabCut \cite{DLC} and LEAP \cite{LEAP}, animal pose estimation has been developing into its own research branch parallel to human pose estimation. DeepLabCut and LEAP both introduce a method for labelling animal body parts and training a deep neural network for predicting 2D body part positions. DeepPoseKit~\cite{DPK} improved the inference speed by a factor of approximately two while maintaining the accuracy of DeepLabCut. 3D Bird Reconstruction \cite{3DBirdReconstruction} predicts 2D keypoints and silhouettes to estimate the 3D shape of cowbirds from a single view.
However, other than the extension of DeepLabCut in DeepLabCut-live \cite{DLC-live}, most applications have focused on offline post-hoc analysis, which limits any application that might require posture estimation at interactive speeds ($\geq 1$ fps) to perform stimulus driven behavior experiments e.g. VR for animals \cite{naik2020animals, naik2021xrforall}.

\subheading{2D Multi-Animal Pose Estimation}
DeepLabCut~\cite{DLC} is extended in~\citeA{maDLC} to predict 2D body parts of multiple animals and maintain identity by temporal tracking.
This extension uses training data with annotations of multiple animals. The authors released four datasets with annotations containing mice ($n=3$), mouse with pups ($n=2$), marmosets ($n=2$) and fish ($n=14$).
Similarly SLEAP~\cite{SLEAP} provides several architectures to estimate 2D body parts of multiple animals.
These two approaches~\cite{maDLC,SLEAP} can track the poses of multiple animals and are trained on multi-animal annotated data. 
However, manual annotations for multi-animal data is often challenging and time consuming to obtain, largely constraining the development of multi-animal methods. 

\subheading{3D Animal Pose Estimation}
To infer 3D poses of single rodents from multi-view data, \citeA{DANNCE} developed DANNCE, a method similar to~\citeA{LToHP} by learning the triangulation process from multiple views using a 3D CNN.
Similar to~\citeA{LToHP},~\citeA{DANNCE} has a cost of longer run times due to its 3D CNN architecture.
Neural Puppeteer~\cite{NePu} is a keypoint based neural rendering pipeline.
By inverse rendering the authors estimate 3D keypoints from multi-view silhouettes.
While this method is independent from variations in texture and lighting, most of their evaluation is performed using synthetic data, and thus its applicability to real-world animal data has not been extensively tested.
\citeA{BKinD-3D} proposes a self-supervised method for 3D keypoint discovery in animals filmed from multiple views without reliance on 2D/3D annotated data.
This method uses joint length constraints and a similarity measure for spatio-temporal differences across multiple views.
While there is no need for annotated data, this method comes with a cost of lower accuracy.
\citeA{MAMMAL} fits a mesh model, which must be prepared for each species, to $10$ camera views for 3D pose estimation of four pigs, two dogs and one mouse captured in indoor environments.
For~\citeA{DeepFly3D,3DDLC,AcinoSet,OpenMonkeyStudio,Anipose,3D-UPPER,han2023social,3D-POP} the procedure to obtain 3D poses is to use a 2D pose estimator (e.g. \citeA{StackedHourglass,DLC}) and to triangulate to 3D using the 2D keypoint predictions of multiple views.
Just like the proposed method, these 3D frameworks exploit 2D keypoints and trigulation from multiple views.

All these methods are limited to the pose tracking of up to four individuals, and no framework has been shown to track the 3D poses of larger animal groups.

\subsection{Multi-Animal Identity Tracking}
Multiple animal tracking~\cite{AnimalTrack}, a variation of multi-object tracking (MOT,~\citeA{dendorfer2021motchallenge}), is important in order to maintain identities of animals throughout behavioural experiments. 

\citeA{Paperidtrackerai} and~\citeA{Zebrafish} use the software idtracker.ai \cite{idtrackerai} to track up to $100$ zebrafish in 2D at once. The software needs to know the number of individuals beforehand since it performs individual identification in each frame. TRex \cite{TRex} is capable of tracking up to $256$ individuals while estimating the 2D head and rear positions of animals. It achieves real-time tracking using background subtraction.
\citeA{AnimalTrack} provides a multi-animal tracking benchmark in the wild. The benchmark includes $58$ sequences with around $25K$ frames containing ten common animal categories with $33$ target objects on average for tracking.
\citeA{Pedersen20203d} provides a zebrafish tracking benchmark in 3D. The benchmark includes 3D data of up to ten zebrafish recorded in an aquarium.

\subheading{Frameworks for Animal Pose Estimation and Identity Tracking}
For applications in biological experiments of multiple individuals, the problem of posture estimation and tracking often goes hand in hand, because the posture of multiple individuals alone will not be meaningful if the identities are not maintained. Existing posture estimation frameworks also provide identity tracking, but are often limited to 2D.

DeepLabCut~\cite{maDLC} splits the workflow in local and global animal tracking. For local animal tracking they build on SORT \cite{SORT}, a simple online tracking approach. For animals that are closely interacting or in case of occlusions they introduce a global tracking method by optimizing the local tracklets with a global minimization problem using multiple cost functions on the basis of the animals' shape or motion.
SLEAP \cite{SLEAP} also uses a tracker based on Kalman filter or flow shift inspired by \citeA{xiao2018si} for candidate generation to track multiple individuals. 

In contrast to the previous two works~\cite{maDLC,SLEAP}, we propose a posture estimation and tracking framework in 2D and 3D, that focuses on online tracking. We first initialize correspondences between cameras using the first frame and then use a 2D tracker from each view to maintain correspondences to reduce computation time. 
In addition, our framework works both on data recorded in captive and outdoor environments.
%

\section{Technical Framework}
We first discuss the datasets that we use for this study, describe the technical framework behind~\paperabr,
explain how we extend the framework to two further use cases, and finally discuss ablation studies and network training.

\subsection{Datasets}
\label{sec:dataset}
We describe the indoor dataset~\cite{3D-POP} and the additional datasets that we use for our two domain shifts including our novel outdoor pigeon dataset containing multi-pigeon annotations.

\subsubsection{3D-POP}
\label{sec:dataset:3d-pop}
For this study, we use the 3D-POP dataset~\cite{3D-POP}, a multi-view multi-individual dataset of freely-moving (i.e. walking) pigeons filmed by both RGB and motion-capture cameras.
This dataset contains RGB video sequences from $4$ views (4K, $3840\times 2160$ px) of $1$, $2$, $5$ and $10$ pigeons.
The ground truth provided by the dataset for each individual is a bounding box (on average $215$ px wide and $218$ px high in 2D), $9$ distinct keypoints in 2D and 3D (beak, nose, left and right eye, left and right shoulder, top and bottom keel and tail), and individual identities. 
For more details on the curation and features of the dataset, we refer to~\citeA{3D-POP}.

From this dataset, we adopt a $60/30/10$ (training/validation/test) split based on 3D-POP~\cite{3D-POP}, by sampling a total of $6036$ random images as our training set from the sequences of $1$, $2$, $5$ and $10$ pigeons ($25\%$ for each type). We ensure that an equal number of frames were sampled from each sequence to avoid bias.
As our validation
set, we sample $3040$
frames separately from the training set following the same sampling method.
As our test set for posture estimation, we use $1000$ frames, across four test sequences of different individual numbers ($1$, $2$, $5$, $10$ pigeons), each $250$ frames long. We choose temporal sequences as the test set to evaluate the complete~\paperabr pipeline (cf.~\cref{fig:framework,sec:framework}).

Finally, to perform quantitative evaluation on multi-object tracking in 2D and 3D, we use the $5$ test sequences containing $10$ pigeons provided in 3D-POP \cite{3D-POP}, ranging between 1 to 1.5 minutes in length.

\subsubsection{Additional Datasets}
\label{sec:dataset:use-cases}
We also extend~\paperabr in two applications of domain shifts of training a single individual model and tracking outdoors, which corresponds to two additional datasets.
For discussion of the implementation of the two use cases, we refer to \cref{sec:framework:application}.

\subheading{Single Pigeon Dataset}
To test if training a model on $1$ pigeon can be used to track multiple pigeons, we sample a single pigeon training set from 3D-POP, using the same sampling method as the multiple pigeon dataset (cf.~\cref{sec:dataset:3d-pop}) but only from single pigeon sequences. The dataset contains $6006$ and $3012$ images for training and validation respectively. We use the same $1000$ frames of test sequences (cf.~\cref{sec:dataset:3d-pop}) that contains both single and multi-individual data for quantitative evaluation.

\subheading{\dataabr}
To evaluate tracking in the wild, we provide a novel dataset collected from pigeons foraging in an outdoor environment. The data is collected from 4 synchronized and calibrated cameras (4K, 30fps) mounted on tripods in a rectangular formation, similar to 3D-POP \cite{3D-POP}. We hope to mirror the 3D-POP setup to minimize the differences between the indoor and outdoor datasets, with the only difference being the outdoor environment.

The dataset consists of short sequences featuring between 1 to 3 pigeons under natural sunlight conditions. To provide a quantitative evaluation of pose estimation performance in the wild, we also sample and manually annotate $500$ frames from a single individual sequence, taken from all 4 views ($2000$ frames in 2D). 
These annotated keypoints are then triangulated to obtain 3D ground truth data. 
To the best of our knowledge, this is the first calibrated multi-view video dataset of more than one animal that is captured in fully outdoor settings (cf.~\citeA{AcinoSet} for a 3D single Cheetah dataset).

Finally, for additional network training and fine-tuning (cf.~\cref{sec:applications:pigeons-in-the-wild}), we further separated the dataset into an $80\%/20\%$ train-test split, resulting in $100$ 3D test frames for evaluation.

For more details on data collection and calibration procedure used for the dataset, we refer to the supplementary information.

\subsection{Pose Estimation and Identity Tracking}
\label{sec:framework}
\begin{figure*}[!ht]
  \centering
  \includegraphics[width=\linewidth,trim={5cm 9cm 13cm 14cm},clip]{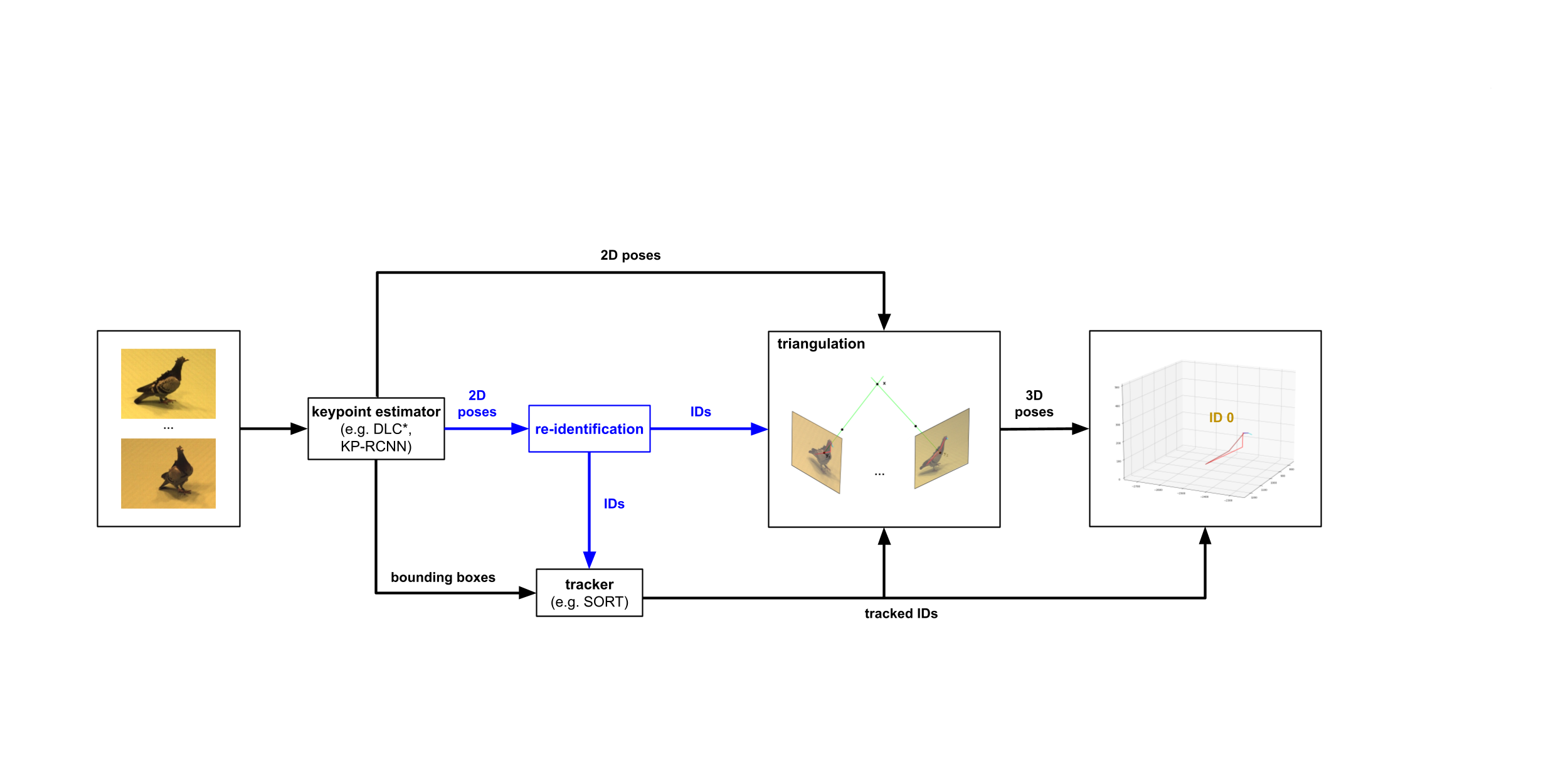}
  \caption[\paperabr]{\emph{\paperabr}.
  The framework consists of a pose estimation and tracking module, into which we can readily slot any state of the art pose estimator and tracking method.
  We identify all individuals in all views (blue part) based on~\citeA{huang2020end} in the first frame only.
  In the subsequent frames we track the identities (IDs) with SORT~\cite{SORT}.
  \paperabr predicts 3D poses together with IDs from multi-view image input using triangulation.
  For details we refer to~\cref{sec:framework}.
  }
  \label{fig:framework}
\end{figure*}
This work extends upon I-MuPPET~\cite{I-MuPPET} and thus the core components of our framework are a pose estimation module and a tracking module, into which we can readily slot any state of the art pose estimator or tracking method, see \cref{fig:framework}.
In the pose estimation module we use three methods for comparison, i.e. a KeypointRCNN~\cite{MaskRCNN}, a modified DeepLabCut (DLC,~\citeA{DLC}) and a modified ViTPose~\cite{ViTPose}.
We choose DLC and ViTPose because they are state of the art frameworks for animal and human pose estimation respectively. 
The choice of the KeypointRCNN allows for the domain shift from single to multiple individuals, cf.~\cref{sec:single-multi-domain-shift}.
In addition KeypointRCNN achieves the fastest inference speed for multiple individuals (on average $7.5$ fps and $1.76$ fps for 2D and 3D poses respectively, cf.~\cref{table:2DError,table:3DError} respectively).
In this way we present options for the pose estimator module in terms of accuracy and speed, allowing researchers to choose based on their application.

For the modified DLC and ViTPose, we adopt a top-down approach, by first using YOLOv8~\cite{YOLOv8} to detect the individual pigeons in each frame and
then pass the cropped pigeon images into the single individual DLC~\cite{DLC} and ViTPose~\cite{ViTPose} pipeline.
For details, we refer to~\citeA{DLC} and~\citeA{ViTPose}.
In the following, we denote
these models by DLC* and ViTPose* (with an asterisk).

The KeypointRCNN is a PyTorch \cite{pytorch} implementation of a Mask R-CNN \cite{MaskRCNN}, which is modified to output
nine keypoints for each detected instance (individual),
in addition to a confidence score (confidence of the model about its prediction), label (background vs. object) and bounding box.
Like DLC \cite{DLC}, this network has a ResNet-$50$-FPN~\cite{ResNet,FPN} backbone that was pre-trained on ImageNet \cite{ImageNet}.
For details, we refer to~\citeA{MaskRCNN}.
The input to the KeypointRCNN are RGB images (cf. \cref{fig:framework}) normalized to mean and standard deviation of~$0.5$.

\subheading{3D Posture Estimation}
We use the 2D postures of all four camera views obtained from KeypointRCNN, DLC* and ViTPose* to acquire 3D keypoint estimates using triangulation with sparse bundle adjustment.
Since correspondence matching errors during triangulation can lead to inflated error metrics in terms of RMSE which do not reflect the actual accuracy of the methods,
we apply a Kalman filter~\cite{KalmanFilter} to smooth our pose estimates.
In the following, we denote the three~\paperabr posture estimation modules by 3D-KeypointRCNN, 3D-DLC* and 3D-ViTPose*.

\subheading{3D Mutli-Animal Identity Tracking}
For multi-animal tracking, we first use SORT~\cite{SORT} to track the identity of individuals in each of the four camera views in 2D. 
We chose this method since we are primarily interested in online tracking and high inference speed, and SORT~\cite{SORT} can run up to $260$~fps. We use standard parameters and a maximum age of~$10$ frames (refer to~\citeA{SORT} for details). 

To match each individual across views, we use a dynamic matching algorithm based on~\citeA{huang2020end} in the first frame to assign each SORT ID from each view to a global ID (cf. blue part in~\cref{fig:framework}). After the assignment, we maintain the identities based on SORT tracking in 2D. We choose to do identity matching in the first frame only to allow the whole framework to be used in an online manner.

The dynamic matching algorithm first generates 3D pose estimates for each possible pair of 2D poses, creating a large 3D pose subspace. 
Within the 3D pose subspace, we match 3D poses that are close together based on the Euclidean distance, and assign 2D poses that contribute to the matched 3D poses as the same individual.
We match until the pairwise distance threshold of $200$ mm is reached.
Since the algorithm does not know the number of individuals in the scene, we choose a conservative threshold of $200$ mm to ensure all individuals are matched. Note that the algorithm prioritizes matches with lower distance, hence a larger threshold doesn't lead to worse performance.
For more details we refer to~\citeA{huang2020end}.
After the dynamic matching is completed, we maintain the global ID in subsequent frames and triangulate based on the 2D tracklets from SORT. 
Finally, if a 2D tracklet in a certain camera view is lost or switched, we skip the detections of the given camera.

\subsection{Further Applications}
\label{sec:framework:application}
Here, we discuss how we adapt our framework for the two use cases of training a single pigeon model and posture tracking outdoors.

\subheading{Single to Multi-Animal Domain Shift}
Annotating frames of multiple individuals is often more labour intensive than labelling frames with a single animal.
Here, we explore this idea by training a model using our single pigeon dataset (cf.~\cref{sec:dataset:use-cases}).
For trianing and evaluation, we use the same framework as for indoor posture tracking from~\cref{fig:framework,sec:framework}.
However, in our pose estimation module we use the KeypointRCNN because the YOLOv8 object detection model in DLC* and ViTPose* cannot reliably generalize to multiple pigeons when only trained on images of a single pigeon.

\subheading{Pigeons in the Wild}
Usually, the difference in the background between different datasets is one of the biggest hurdles for generalizing a keypoint detection model trained on an annotated dataset to other data of the same species. 
Here, we propose a methodology to eliminate the effect of the background to estimate postures of pigeons in the wild without further annotation and fine-tuning.
For training, we make use of the same multi-animal training set sampled from 3D-POP, cf.~\cref{sec:dataset:3d-pop}. But as an extra processing step, we remove the influence of the background by using the Segment-Anything-Model (SAM,~\citeA{SAM}), a model that allows objects in an image to be segmented based on a prompt of the object location (ground truth bounding box). We then train our framework to predict keypoints on masked images instead of crops from bounding boxes that contains both the object and background.

For the choice of pose estimator module, we train both ViTPose* and DLC* on the masked images because they perform similarly well on the 3D-POP dataset (cf.~\cref{table:3DError}). To remove confusion from the 3D-POP benchmarking results, we refer to these 2 models as Wild-VitPose and Wild-DLC.

Finally, we evaluate the models on the 100 test frames of our novel Wild-MuPPET dataset, cf.~\cref{sec:dataset:use-cases}.
We first use a pre-trained MaskRCNN~\cite{MaskRCNN} to localize and segment all objects with the ``bird'' class in the frame and then pass them to the pose estimator.
We do not use SAM during inference because it does not provide category labels.
Unlike the evaluation on 3D-POP, we also do not perform any temporal filtering since the Wild-MuPPET test set only contains individually sampled frames.

\subsection{Network Training and Ablation Studies}
\label{sec:besttrainingconfiguration}
\subheading{Data Augmentation}
In I-MuPPET~\cite{I-MuPPET}, we performed ablation studies on data augmentation for pigeons. These ablation studies can be found in our supplemental material.
In this work, we use the same data augmentation parameters to train the KeypointRCNN model (cf.~\cref{sec:framework}).
This includes changing the sharpness with a probability of $0.2$, blurring the input image with a small probability of $0.2$, randomly jittering the brightness by a factor chosen uniformly from $[0.4, 1.6]$, a flipping probability of $0.5$ and a small scaling range of $\pm 5\%$.

For DLC* (cf.~\cref{sec:framework}), we use their default augmentation parameters~\cite{DLC,YOLOv8} that also include blurring and jittering.

And finally for ViTPose*, we also use the default augmentation implementation~\cite{ViTPose} for our animal posture tracking.

\subheading{Training Hyperparameters}
To find out the best network configuration for KeypointRCNN (cf.~\cref{sec:framework}) we perform several experiments (see supplemental material).
From this analysis we find that using a learning rate of $0.005$ and reducing it by $\gamma=0.5$ every given step size to reach a final learning rate of $0.0003$ at the end of training works best.

For DLC* (cf.~\cref{sec:framework}), we use a custom learning rate schedule from $0.0001$ to $0.00001$ over $30000$ iterations for DLC, and default hyperparamters for all others~\cite{DLC,YOLOv8}.

For ViTPose*, we use default hyperparamters and training configuration~\cite{ViTPose}, with a custom learning rate of $0.00005$.

\subheading{Training Procedure} For all trained neural networks, we monitor the validation loss when training, with the final weights chosen based on the epoch with the lowest validation loss overall to ensure the best performance and least over-fitting. For DeepLabCut, we instead use RMSE accuracy provided by the package~\cite{DLC}, and for ViTPose, we use the highest mean average precision ($mAP$) score.

This procedure can lead to a different number of training epochs in each experiment. Nevertheless experiments are comparable in the sense that each model is trained to perform best without over-fitting to the training data.
%

\section{Evaluation}
\label{sec:evaluation}
We evaluate each module of~\paperabr on test sequences of the 3D-POP dataset.
We separate our evaluation into three parts, to provide an idea of how each component of the framework performs.
First, we evaluate keypoint estimation accuracy in~\cref{sec:keypoint_estimation}. Second, we evaluate identity tracking accuracy and third, we evaluate inference speed.
The latter two evaluations are both in~\cref{sec:i-muppet-tracking-performance}.
We first briefly discuss the evaluation metrics we use in~\cref{sec:metrics}, then report quantitative results on each of the components above. Finally, we also show qualitative results on all tasks.

Since the current framework extends the work of I-MuPPET~\cite{I-MuPPET}, which relies on triangulating 2D posture estimates into 3D, in~\cref{sec:keypoint_estimation} 
we evaluate both 2D performance and 3D performance for all tasks to provide insights into how errors propagate. 

\subsection{Metrics}
\label{sec:metrics}
\subheading{Pose Estimation}
Two widely used metrics, also in human pose estimation, are the Root Mean Square Error (RMSE), in human pose estimation better known as Mean Per Joint Position Error (MPJPE, cf. e.g. \citeA{LToHP}), and the Percentage of Correct Keypoints (PCK, cf. e.g. \citeA{yang2013ar}). DeepLabCut~\cite{DLC} uses the former, 3D Bird Reconstruction~\cite{3DBirdReconstruction} the latter to evaluate their animal pose estimation, hence we use both here.

RMSE is calculated by taking the root mean squared of the Euclidean distance between each predicted point and the ground truth point, while
PCK is the percentage of predicted keypoints that fall within a given threshold~\cite{3DBirdReconstruction}. We compute PCK05 and PCK10, where the threshold is a fraction ($0.05$ and $0.1$) of the largest dimension of the ground truth bounding box for 2D and the maximum distance between any two ground truth keypoints in 3D.
Compared to RMSE, the PCK takes into account the size and scale of the tracked animal, providing a more meaningful estimate of keypoint accuracy compared to the RMSE.

\subheading{Tracking}
There are three sets of tracking performance measures that are widely used in the literature \cite{dendorfer2021motchallenge}:
the CLEAR-MOT metrics introduced in \citeA{CLEARMOT},
the metrics introduced in \citeA{nevatia2009learning} to measure track quality,
and the trajectory-based metrics proposed in \citeA{ristani2016performance}.
Here, we also report the novel Higher Order Tracking Accuracy (HOTA), introduced in~\citeA{HOTA} because the other metrics overemphasize the importance of either detection or association.
HOTA measures how well the trajectories of matching detections align, and averages this over all matching detections, while also penalising detections that do not match \cite{HOTA}.

For further details on the tracking metrics we refer to~\citeA{dendorfer2021motchallenge,HOTA}.
A detailed description of each reported metric is also available in the supplementary material.
For the evaluation, we use code provided by~\citeA{luiten2020trackeval,dendorfer2020motchallengeevalkit}.

\subheading{Inference Speed}
We also benchmark the inference speed of our framework in 2D and 3D with all $1000$ frames in the test set
from 3D-POP~\cite{3D-POP}, cf.~\cref{sec:dataset:3d-pop}.
For this evaluation, we use a workstation with a 16GB Nvidia Geforce RTX 3070 GPU, 11th Gen Intel(R) Core(TM) i9-11900H @ 2.50GHz CPU, and Sandisk 2TB SSD.

Since each pose estimation module of~\paperabr (cf.~\cref{fig:framework}) has different data and model loading procedures, we
include all processes (data loading, model loading, inference, data saving) to get a realistic comparison of the processing time.
We loop three times over each inference script and report the average speed in frames per second (fps).
We consider the framework as interactive if the inference speed is $\geq 1$ fps.

\subsection{Pose Estimation}
\label{sec:keypoint_estimation}
We report quantitative and qualitative results of 2D and 3D poses on the indoor pigeon data (cf.~\cref{sec:dataset:3d-pop}) and compare~\paperabr to a 3D baseline based on 3D CNNs~\cite{LToHP}.
Furthermore, to illustrate the applicability to other species, we also compare the KeypointRCNN (cf.~\cref{sec:framework}) to DLC~\cite{DLC} on their 2D odor trail tracking data
and
to 3D Bird Reconstruction~\cite{3DBirdReconstruction}
on their 2D cowbird keypoint dataset, both available in the supplementary materials.

\subheading{3D Baseline}
For a 3D comparison, we train the ``Learnable Triangulation of Human Pose'' framework (LToHP,~\citeA{LToHP}), 
on the same training dataset specified in~\cref{sec:dataset:3d-pop}.
We perform this comparison because the framework is state of the art for human 3D posture estimation, and uses a 3D CNN architecture, which is shown to be more accurate than simple triangulation~\cite{LToHP}.
With this comparison we can evaluate how well the triangulation based~\paperabr performs, since models like LToHP rely on 3D ground truth datasets, which are rare in animal posture tracking.

The framework predicts a 2D heatmap from each view that is projected into a 3D voxel grid, then learns to predict 3D keypoints using a 3D CNN architecture.
Since the model requires a 3D root point
as input, we train both an algebraic
and volumetric triangulation model by providing cropped images of pigeon individuals based on ground truth bounding boxes.
During inference, we follow the same workflow as in~\citeA{LToHP} by first obtaining a root point estimate (top keel) using the algebraic model, then run the volumetric model to obtain 3D keypoint estimates.
We refer to~\citeA{LToHP} for more details.

For model training, we train the algebraic model for $292$ epochs and the volumetric model for $782$ epochs with default augmentation parameters, both models having lowest validation loss.

Finally, note that since LToHP is a single subject framework, we make use of ground truth bounding boxes to crop the image inputs during training and inference, with the goal of providing a baseline for 3D posture estimation accuracy, but not as a complete pipeline. Implementing a complete pipeline for multi-animal 3D CNN based posture estimation is outside the scope of this study, and can be a
further application of~\paperabr, where it can replace the algebraic model together with the ground truth bounding boxes by providing root point estimate, bounding boxes and identities to the volumetric model of LToHP.

\begin{table*}[t]
    \centering
    \caption[]{
        {\em Quantitative Evaluation of 2D Pigeon Poses.}
        We report the RMSE and its median (px),
        PCK05 ($\%$) and PCK10 ($\%$) for estimated 2D poses on the 3D-POP test sequences.
        Comparison between KeypointRCNN (KP-RCNN, cf.~\cref{sec:framework}), modified DeepLabCut (DLC*) and modified ViTPose (ViTPose*).
        *: We combine YOLOv8~\protect\cite{YOLOv8} for instance detection with single-object DLC~\protect\cite{DLC} and ViTPose~\protect\cite{ViTPose}.
        We also report the mean 2D inference speed for the complete pipelines in fps. For details on the inference speed we refer to~\cref{sec:i-muppet-tracking-performance}.
        Upwards and downwards arrows represent whether a higher or lower value is better, respectively.
        Best results per row in bold.
        }
    \begin{tabular}{ c | c | c | c}
        \toprule
        Metric / Method & KP-RCNN & DLC* & ViTPose*\\
        \midrule
        RMSE ($px$) $\downarrow$ & $\mathbf{28.1}$ & $39.0$ & $38.9$\\
        Median ($px$) $\downarrow$ & $5.7$ & $4.7$ & $\mathbf{4.4}$\\
        PCK05 ($\%$) $\uparrow$ &  $82.4$ & $89.1$ & $\mathbf{91.1}$\\
        PCK10 ($\%$) $\uparrow$ &  $95.4$ & $\mathbf{96.8}$ & $\mathbf{96.8}$\\
        Mean Speed (fps) $\uparrow$ & $\mathbf{7.5}$ & $3.0$ & $2.1$\\
        \bottomrule
    \end{tabular}
    \label{table:2DError}
\end{table*}
\begin{table*}[t]
\centering
\caption{
    {\em Quantitative Evaluation of 3D Pigeon Poses.}
    We report the filtered (cf.~\cref{sec:framework}) RMSE and its median ($mm$),
    PCK05 ($\%$) and PCK10 ($\%$) for the 3D poses on the 3D-POP test sequences.
    Comparison between LToHP~\protect\cite{LToHP} and~\paperabr (highlighted in gray).
    *: We combine YOLOv8~\protect\cite{YOLOv8} for instance detection with single-object DLC~\protect\cite{DLC} and ViTPose~\protect\cite{ViTPose}.
    We also report the mean 3D inference speed for the complete pipeline in fps. For details on the inference speed we refer to~\cref{sec:i-muppet-tracking-performance}.
    Upwards and downwards arrows represent whether a higher or lower value is better, respectively.
    Best results per row in bold.
    See text for a discussion of the results.
    }
\begin{tabular}{c|a|a|a|c}
\toprule
Metric / Method & 3D-KP-RCNN & 3D-DLC* & 3D-ViTPose* & LToHP~\protect\cite{LToHP} \\
\midrule
RMSE ($mm$) $\downarrow$  & $25.0 $ & $25.0$ & $24.0$ & $\mathbf{14.8}$ \\
Median ($mm$) $\downarrow$  & $9.4$ & $7.5$ & $7.0$ & $\mathbf{5.8}$ \\
PCK05 ($\%$) $\uparrow$  & $53.2$ & $66.1$ & $71.0$ & $\mathbf{76.7}$ \\
PCK10 ($\%$) $\uparrow$  & $85.4$ & $90.9$ & $92.5$ & $\mathbf{94.3}$ \\
Mean Speed (fps) $\uparrow$  & $\mathbf{1.76}$ & $0.72$ & $0.51$ & $0.38$ \\
\bottomrule
\end{tabular}
\label{table:3DError}
\end{table*}
\begin{figure}[t]
    \centering
    \includegraphics[width=\columnwidth]{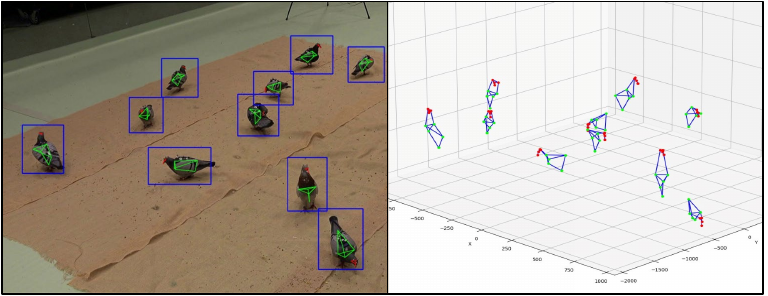}
    \caption{
        {\em Qualitative 3D Results}:
        Example frame from 3D-POP, cf.~\cref{sec:dataset:3d-pop}.
        2D (left side) and 3D (right side) pose estimates using~\paperabr.
        }
    \label{fig:BarnTrack}
\end{figure}
\begin{figure*}[t]
  \centering
  \includegraphics[width=0.199\linewidth]{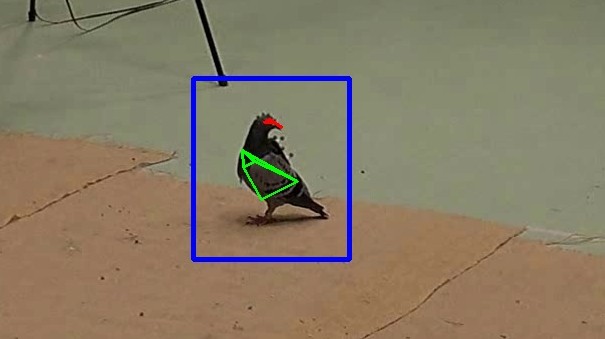}\hfill\includegraphics[width=0.199\linewidth]{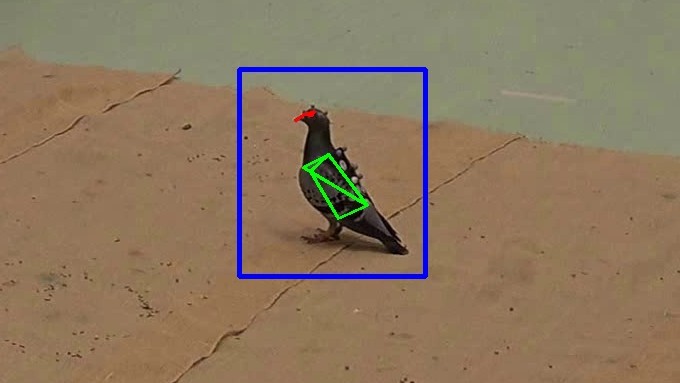}\hfill\includegraphics[width=0.199\linewidth]{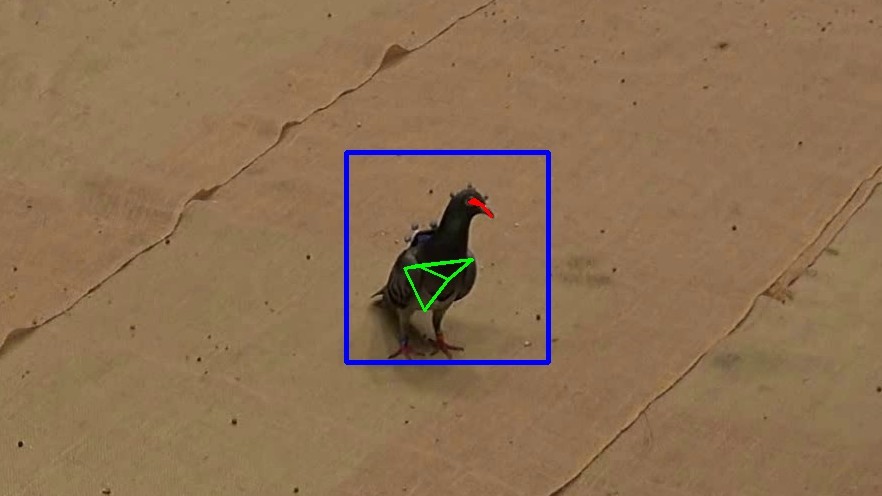}\hfill\includegraphics[width=0.199\linewidth]{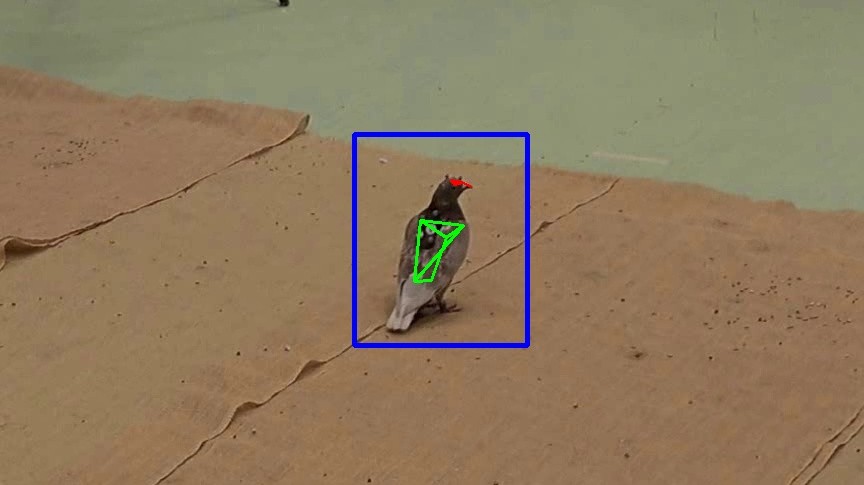}\hfill\includegraphics[width=0.199\linewidth]{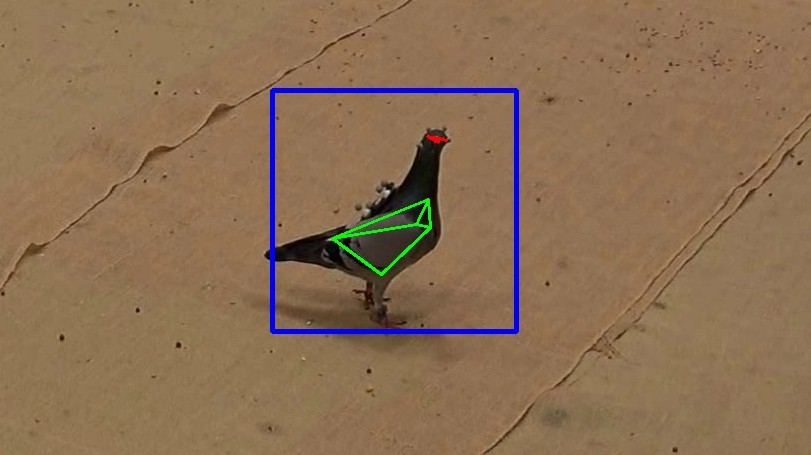}
  \includegraphics[width=0.199\linewidth]{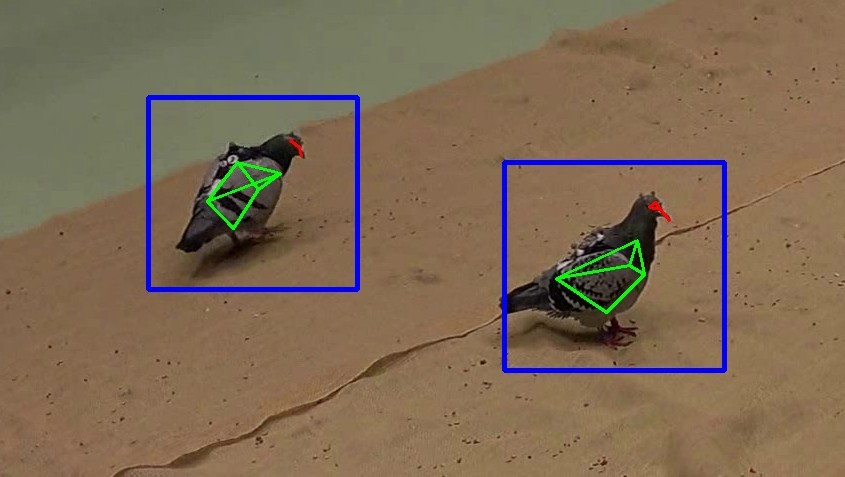}\hfill\includegraphics[width=0.199\linewidth]{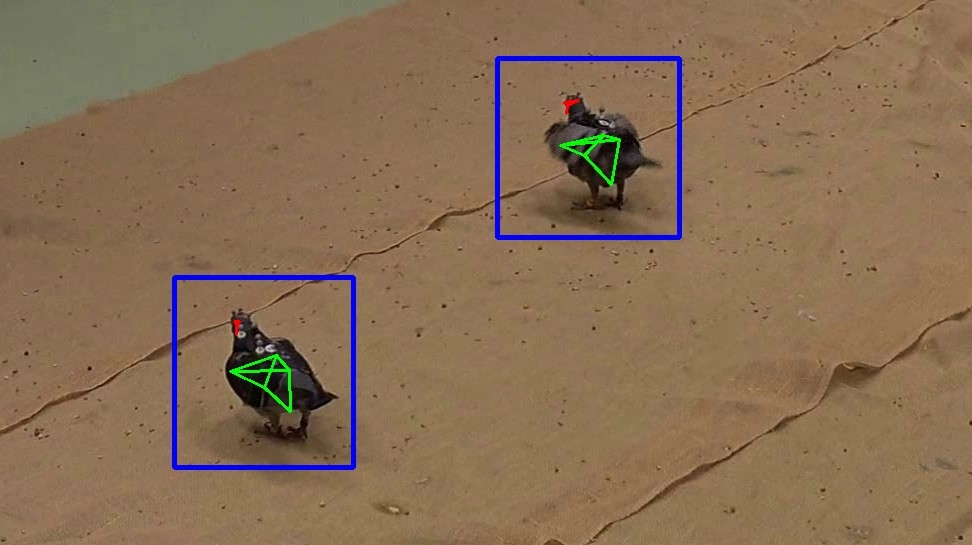}\hfill\includegraphics[width=0.199\linewidth]{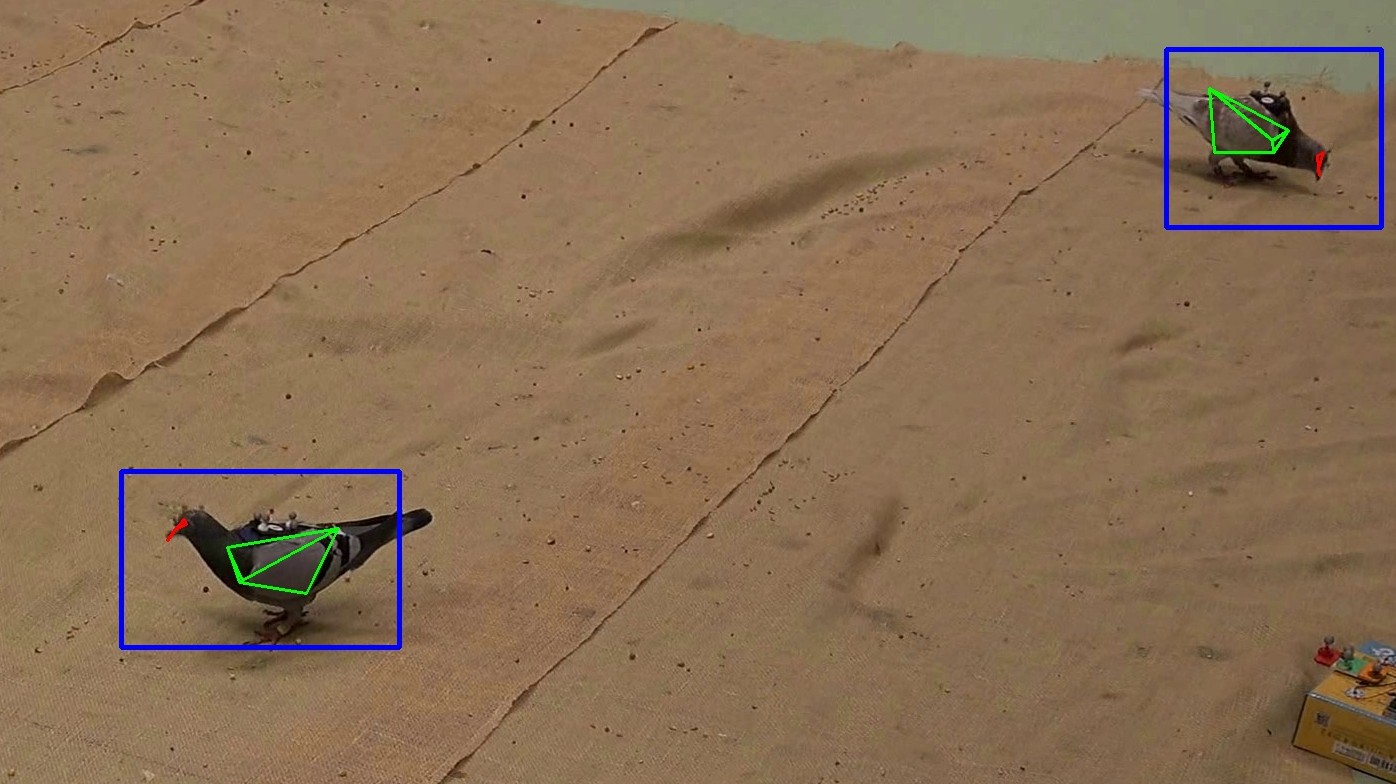}\hfill\includegraphics[width=0.199\linewidth]{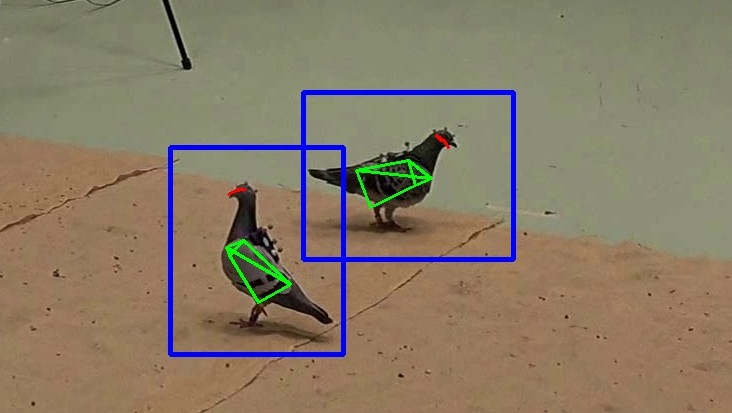}\hfill\includegraphics[width=0.199\linewidth]{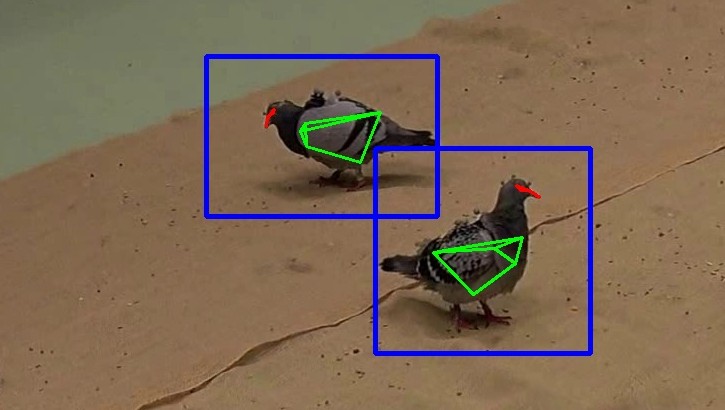}
  \includegraphics[width=0.199\linewidth]{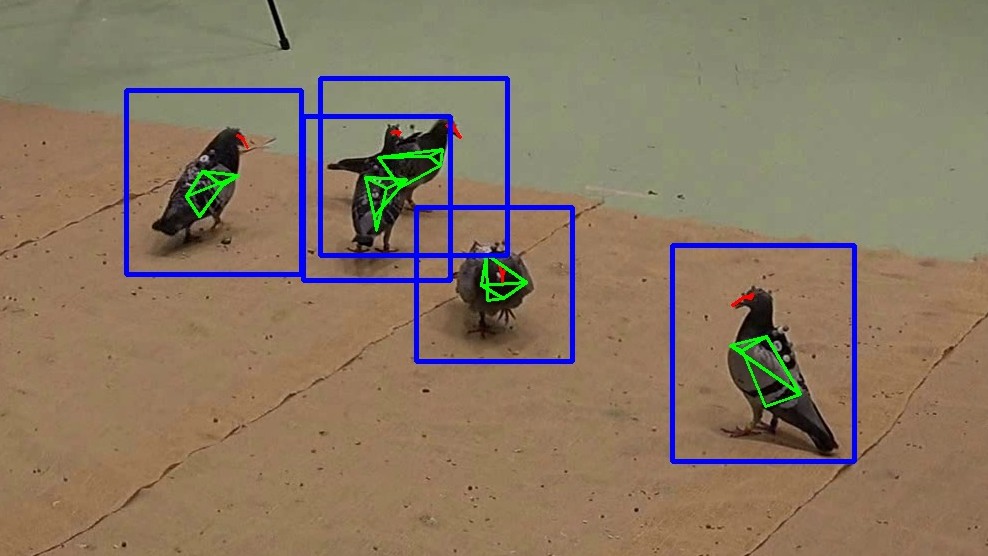}\hfill\includegraphics[width=0.199\linewidth]{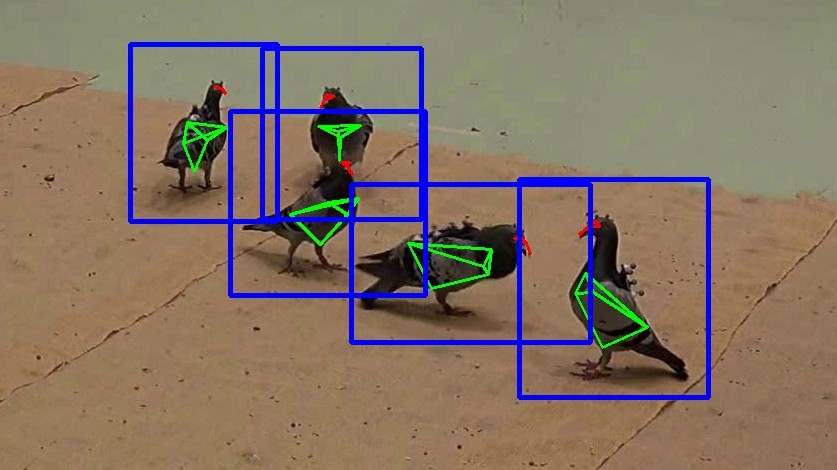}\hfill\includegraphics[width=0.199\linewidth]{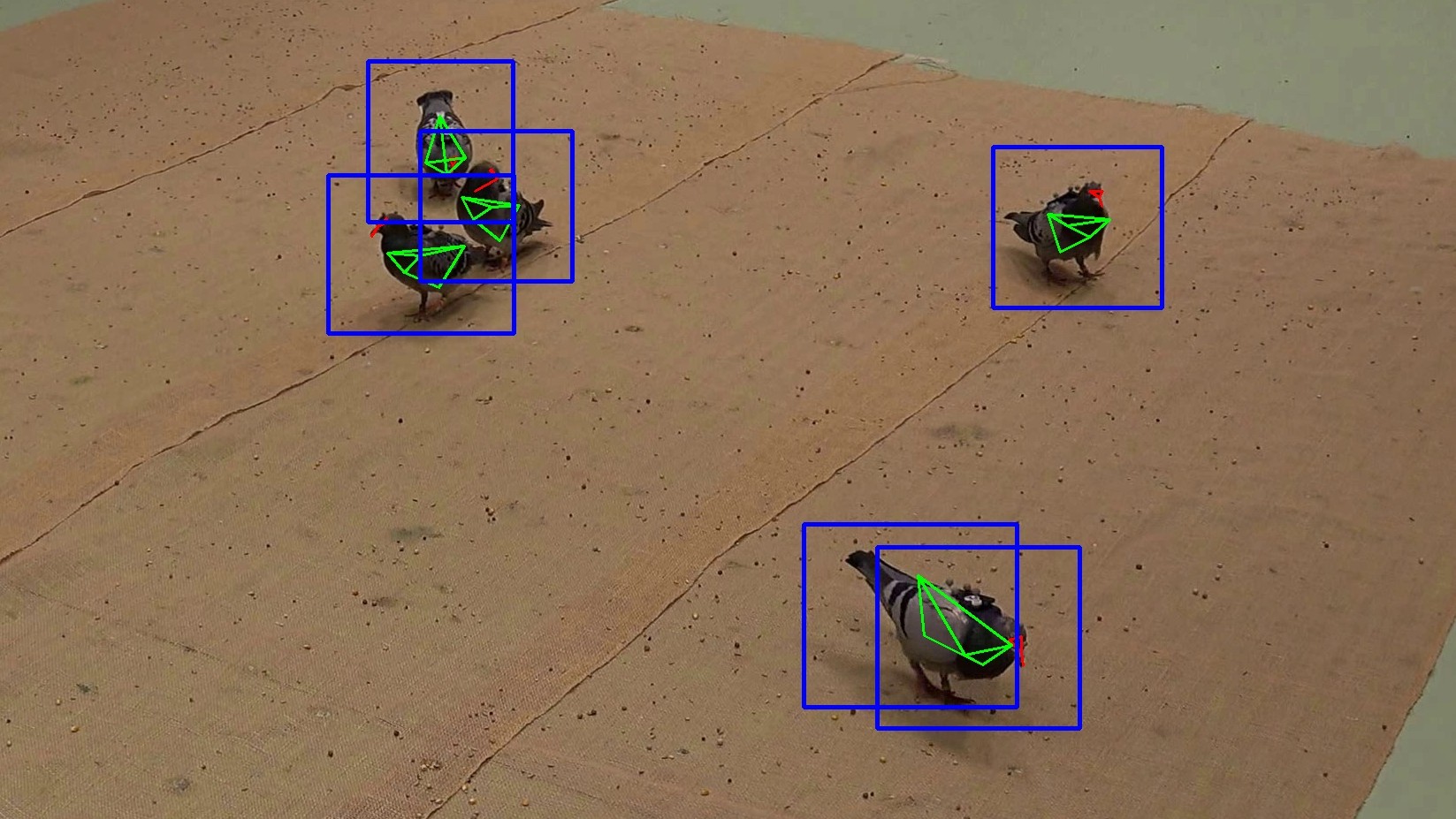}\hfill\includegraphics[width=0.199\linewidth]{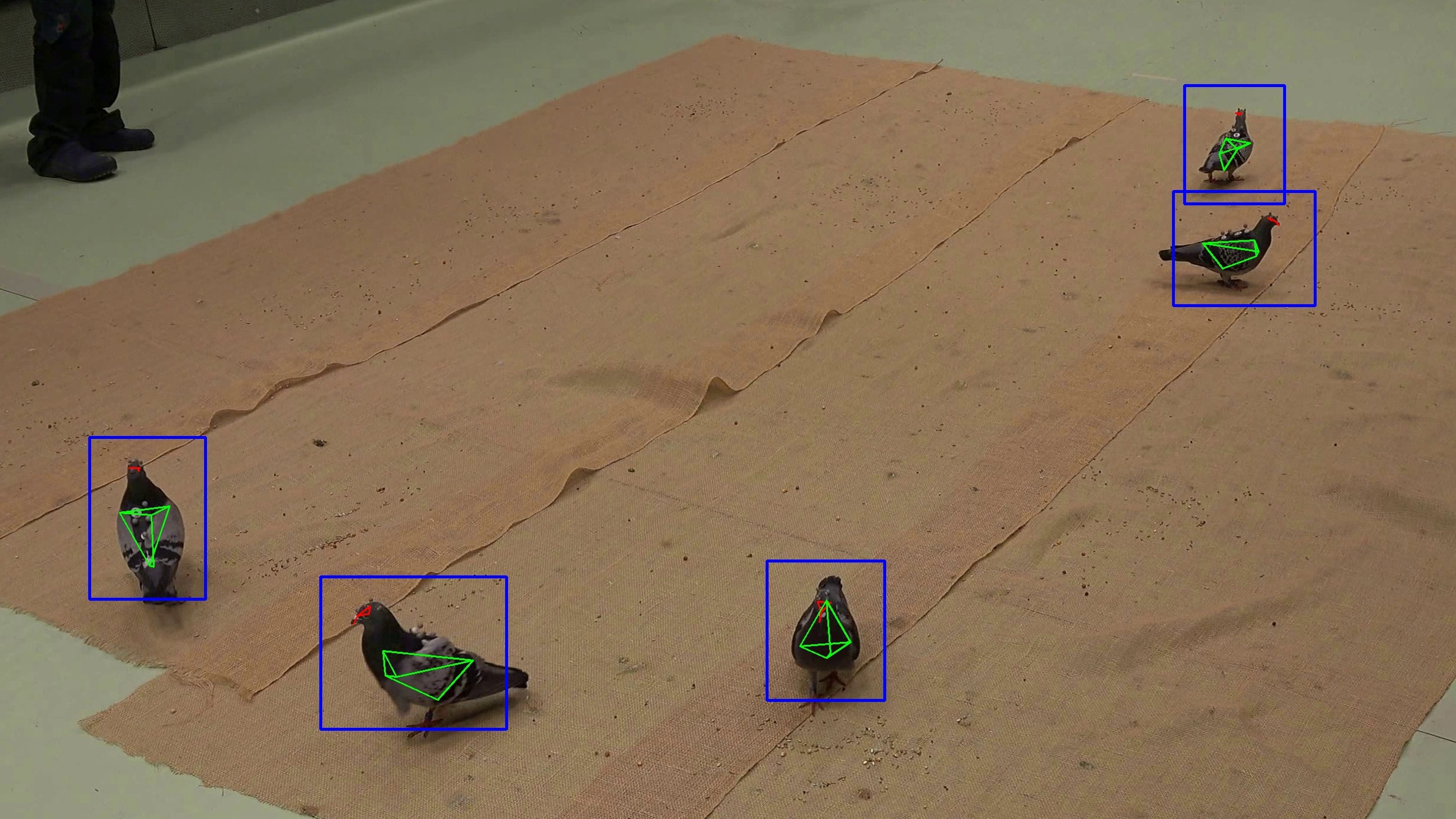}\hfill\includegraphics[width=0.199\linewidth]{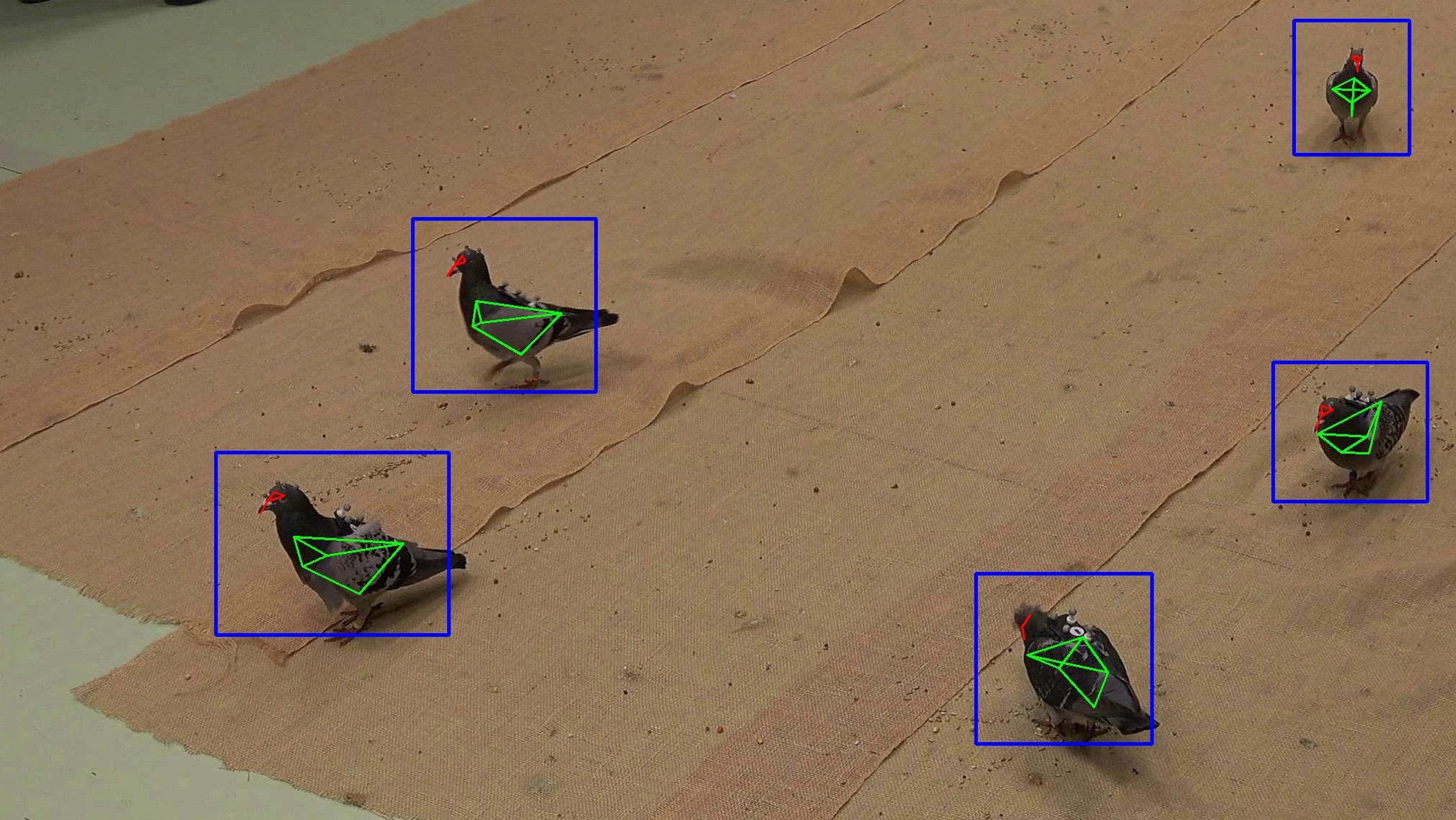}
  \includegraphics[width=0.199\linewidth]{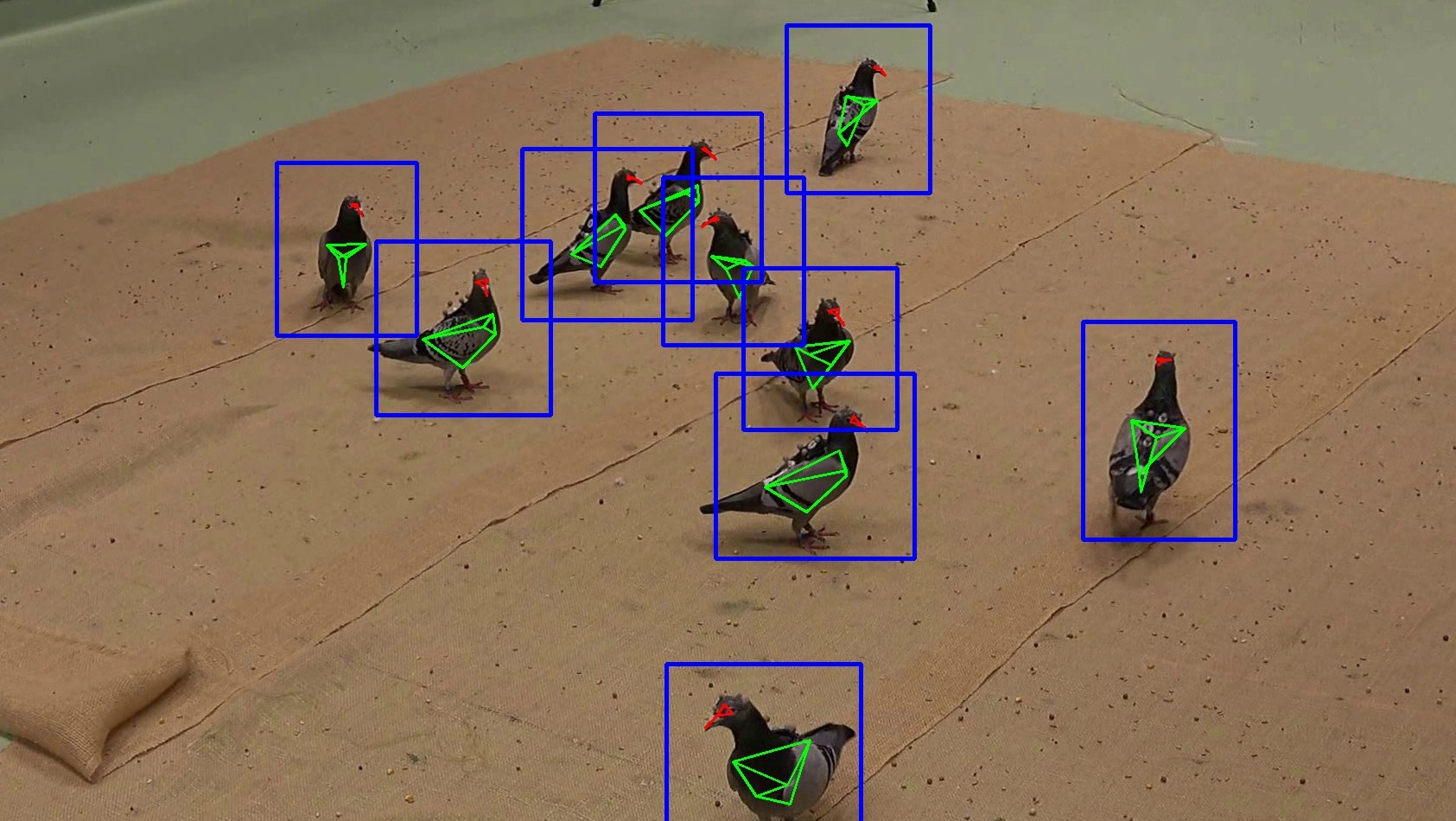}\hfill\includegraphics[width=0.199\linewidth]{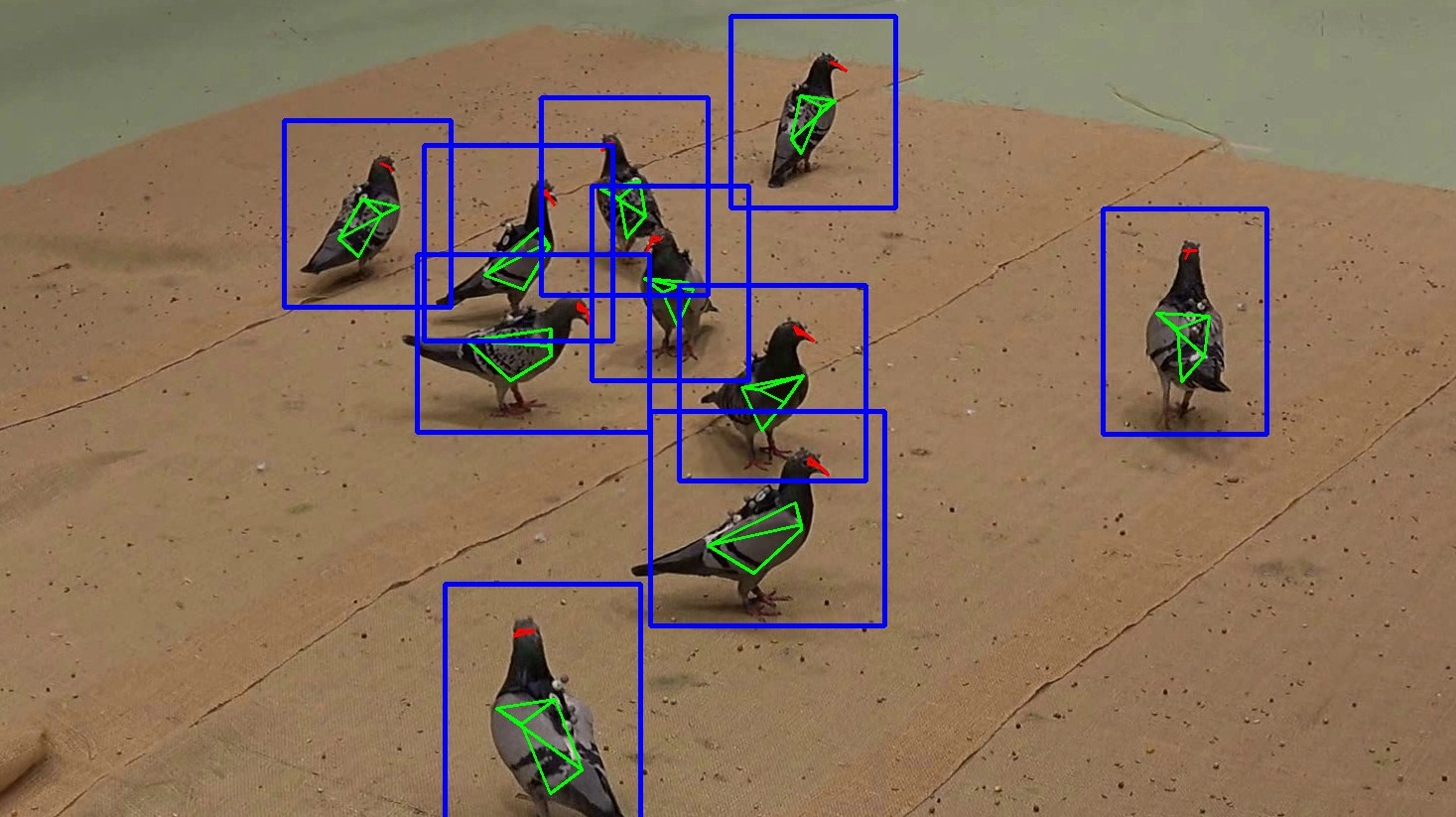}\hfill\includegraphics[width=0.199\linewidth]{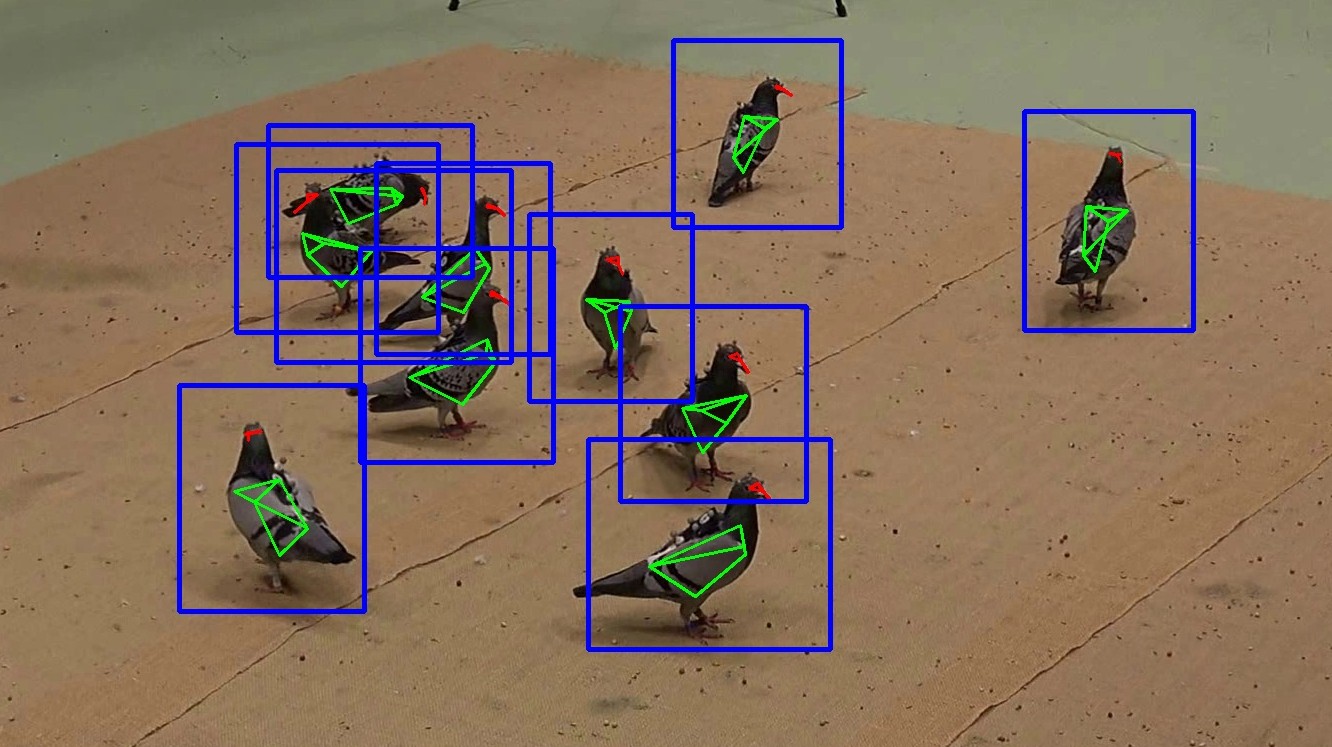}\hfill\includegraphics[width=0.199\linewidth]{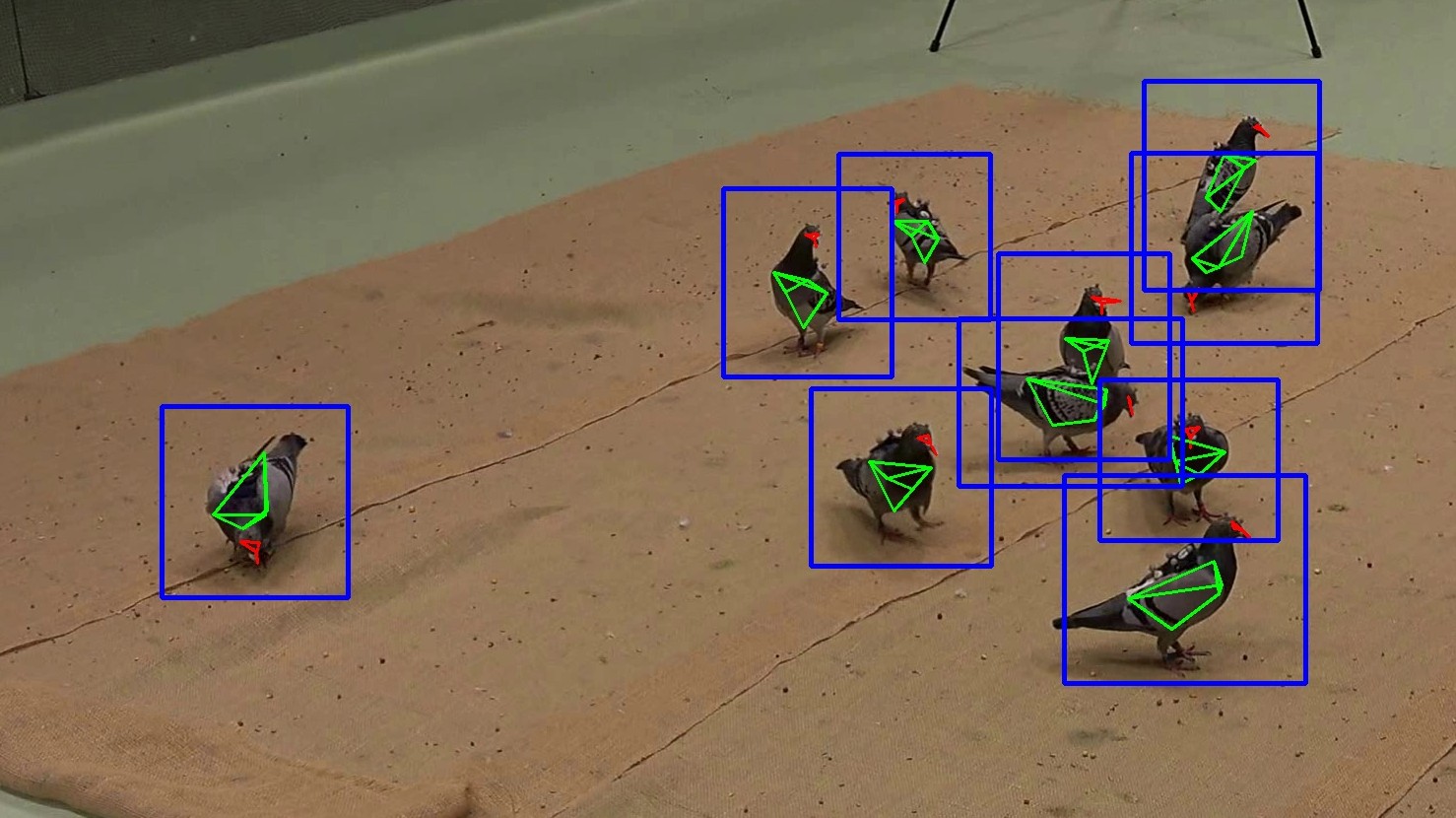}\hfill\includegraphics[width=0.199\linewidth]{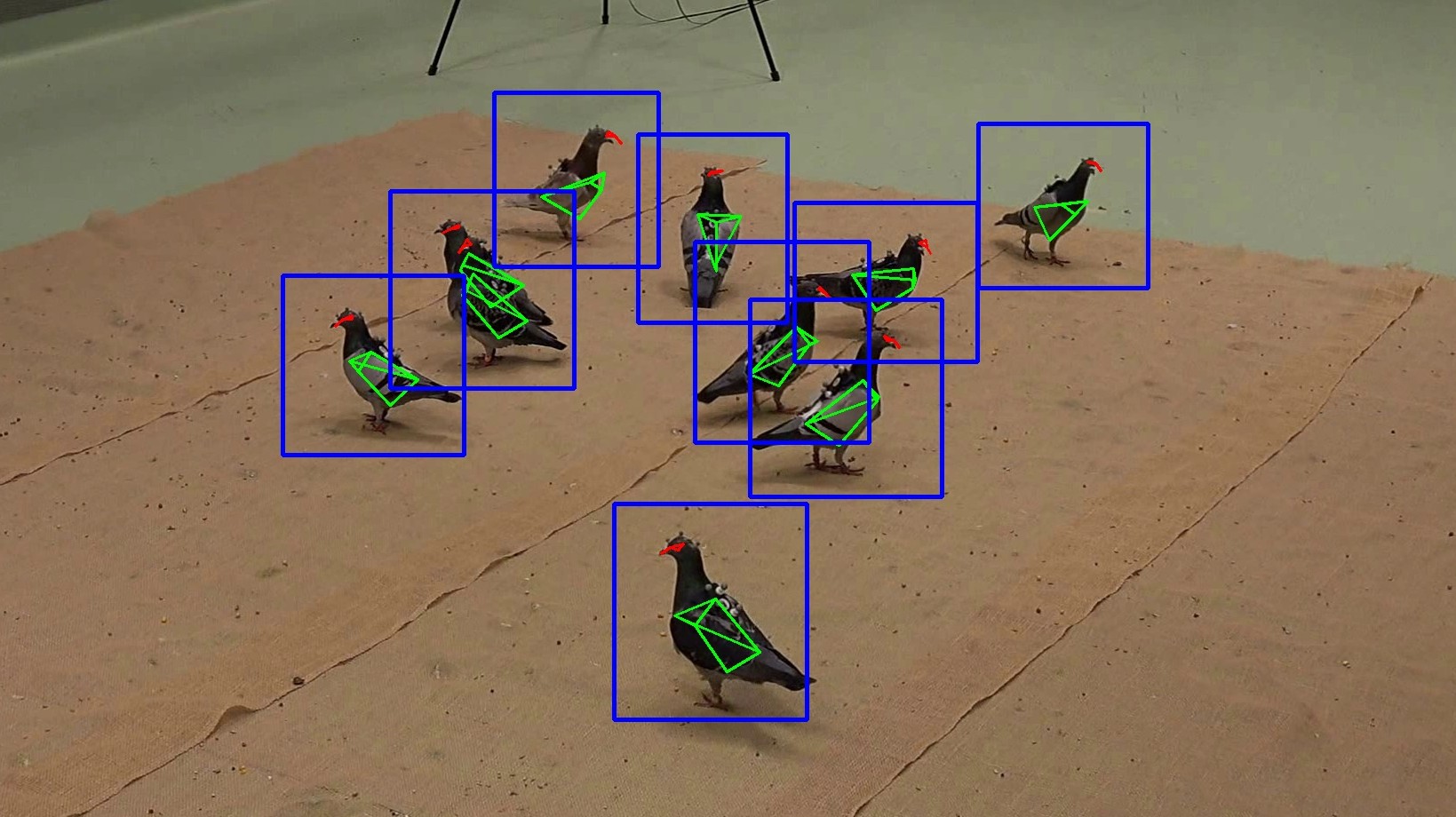}
  \caption[Qualitative Results]{\emph{Qualitative Results}. Example frames from 3D-POP~\cite{3D-POP} for multi-pigeon pose estimation and tracking in 3D, reprojected to 2D view. Green lines connect the body, red lines the head keypoints. Some frames are cropped for a better view.
  }
  \label{fig:3d_muppet}
\end{figure*}
\subheading{Results}
We train the different posture estimation modules of~\paperabr on
multi-pigeon data from~\citeA{3D-POP}, cf.~\cref{sec:dataset:3d-pop}, and choose the best weights with the lowest validation loss.
We train the KeypointRCNN (cf.~\cref{sec:framework}) for $44$ epochs.
In the case of DLC* and ViTPose* (cf.~\cref{sec:framework}), we train YOLOv8~\cite{YOLOv8} for $27$ epochs, ViTPose~\cite{ViTPose} for $175$ epochs and DLC~\cite{DLC} for $86 000$ iterations.

Quantitative results for 2D pose estimation are in~\cref{table:2DError}.
We find that ViTPose* performs best across most metrics like median error ($4.4$ px) and PCK (PCK05 $91.1\%$, PCK10 $96.8\%$).
When a more generous threshold is considered in PCK10, both DLC* and ViTPose* are equally accurate (PCK10 $96.8\%$).
KP-RCNN has the lowest RMSE,
likely due to reduced outliers since the RMSE metric is quite sensitive to large outliers which is also reflected in a relatively small median error compared to RMSE (RMSE $28.1$ px, median $5.7$ px).
This difference is likely due to bounding box detection errors in
the YOLOv8 model within DLC* and ViTPose*.

For 3D, when comparing between models in the posture estimation module of~\paperabr, 3D-ViTPose* performs the best across all evaluation metrics with a RMSE of $24.0$ mm, its median of $7.0$ mm, PCK05 of $71.0\%$ and PCK10 of $92.5\%$, cf.~\cref{table:3DError}.
This is not surprising since ViTPose* already performs the best in 2D, cf.~\cref{table:2DError}, and shows that 2D accuracy propagates into 3D.

We conclude that in applications where high accuracy is needed, researchers should prefer 3D-ViTPose* for the pose estimation module of~\paperabr.

Comparing~\paperabr with the 3D baseline in LToHP~\cite{LToHP}, we find that LToHP has the best performance across all metrics with a RMSE of $14.8$ mm, its median of $5.8$ mm, PCK05 of $76.7\%$ and PCK10 of $94.3\%$, cf.~\cref{table:3DError}.
One of the reasons is that the bounding boxes of the subjects
are provided from the ground truth for LToHP, removing the reliance on 2D and 3D multi-animal identity tracking.
In addition, the model can also learn the general 3D structure of a pigeon instead of relying on 2D detection and triangulation.

Nevertheless, we show that~\paperabr produces comparable estimates compared to LToHP (cf.~\cref{fig:BarnTrack,fig:3d_muppet}), given a median difference of only $1.2$ mm between the best model in~\paperabr (3D-ViTPose*) and LToHP, cf.~\cref{table:3DError}. This difference in error is very small in the context of keypoints on a pigeon, and will likely not affect any downstream behavioural experiments. For example, the diameter of the eye of a pigeon is on average around $10-13$ mm \cite{chard1938structure}, which is much larger than the difference between the model estimates.

\subsection{Tracking Performance}
\label{sec:i-muppet-tracking-performance}
\cref{fig:3d_muppet,fig:field_data} show results of the 3D pose estimation and tracking task for multiple pigeons.
Further qualitative results
can be found in our supplementary video at \url{https://youtu.be/GZZ_u53UpfQ}.

\begin{table*}[!ht]
  \centering
  \caption[Quantitative Tracking Evaluation in 2D]{\emph{Quantitative Tracking Evaluation in 2D}. We test~$20$ video sequences quantitatively with the metrics specified in~\cref{sec:metrics} and our supplementary materials. Upwards and downwards arrows represent whether a higher or lower value is better, respectively. The threshold for the confidence score of ViTPose* (cf.~\cref{sec:framework}) is set to~$0.5$.
  }
  \small
  \begin{tabular}{c|cccccccccccc}
    \toprule
    Test seq. & HOTA \hspace{-1.5mm}$\uparrow$ & MOTA \hspace{-1.5mm}$\uparrow$ & MOTP \hspace{-1.5mm}$\uparrow$ & Rcll \hspace{-1.5mm}$\uparrow$ & Prcn \hspace{-1.5mm}$\uparrow$ & MT \hspace{-1.5mm}$\uparrow$ & ML \hspace{-1.5mm}$\downarrow$ & FPF \hspace{-1.5mm}$\downarrow$ & IDS \hspace{-1.5mm}$\downarrow$ & Frag \hspace{-1.5mm}$\downarrow$ & IDF1 \hspace{-1.5mm}$\uparrow$ \\
    \midrule
    $11$, view $1$ & $0.82$ & $0.92$ & $0.90$ & $0.96$ & $0.96$ & $0.90$ & $0$ & $0.39$ & $2$ & $14$ & $0.92$ \\
    $11$, view $2$ & $0.84$ & $0.92$ & $0.88$ & $0.96$ & $0.96$ & $0.90$ & $0$ & $0.41$ & $0$ & $7$ & $0.96$ \\
    $11$, view $3$ & $0.84$ & $0.92$ & $0.89$ & $0.96$ & $0.96$ & $0.90$ & $0$ & $0.41$ & $0$ & $11$ & $0.96$ \\
    $11$, view $4$ & $0.85$ & $0.94$ & $0.90$ & $0.97$ & $0.97$ & $1$ & $0$ & $0.26$ & $3$ & $29$ & $0.95$ \\
    \midrule
    $19$, view $1$ & $0.90$ & $0.99$ & $0.92$ & $0.99$ & $1$ & $1$ & $0$ & $0$ & $2$ & $13$ & $0.97$ \\
    $19$, view $2$ & $0.93$ & $1$ & $0.92$ & $1$ & $1$ & $1$ & $0$ & $0$ & $0$ & $1$ & $1$ \\
    $19$, view $3$ & $0.92$ & $1$ & $0.91$ & $1$ & $1$ & $1$ & $0$ & $0$ & $0$ & $4$ & $1$ \\
    $19$, view $4$ & $0.89$ & $0.99$ & $0.92$ & $0.99$ & $1$ & $1$ & $0$ & $0$ & $4$ & $11$ & $0.94$ \\
    \midrule
    $30$, view $1$ & $0.83$ & $0.96$ & $0.92$ & $0.97$ & $1$ & $1$ & $0$ & $0.03$ & $9$ & $25$ & $0.88$ \\
    $30$, view $2$ & $0.90$ & $0.99$ & $0.93$ & $0.99$ & $1$ & $1$ & $0$ & $0.03$ & $8$ & $15$ & $0.96$ \\
    $30$, view $3$ & $0.89$ & $0.99$ & $0.89$ & $0.99$ & $1$ & $1$ & $0$ & $0.03$ & $2$ & $7$ & $0.99$ \\
    $30$, view $4$ & $0.87$ & $0.99$ & $0.91$ & $0.99$ & $1$ & $1$ & $0$ & $0.02$ & $6$ & $13$ & $0.95$ \\
    \midrule
    $48$, view $1$ & $0.87$ & $1$ & $0.89$ & $1$ & $1$ & $1$ & $0$ & $0$ & $1$ & $14$ & $0.96$ \\
    $48$, view $2$ & $0.90$ & $1$ & $0.90$ & $1$ & $1$ & $1$ & $0$ & $0.02$ & $0$ & $6$ & $1$ \\
    $48$, view $3$ & $0.91$ & $1$ & $0.91$ & $1$ & $1$ & $1$ & $0$ & $0$ & $0$ & $4$ & $1$ \\
    $48$, view $4$ & $0.91$ & $1$ & $0.90$ & $1$ & $1$ & $1$ & $0$ & $0$ & $0$ & $3$ & $1$ \\
    \midrule
    $59$, view $1$ & $0.77$ & $0.98$ & $0.89$ & $0.98$ & $1$ & $1$ & $0$ & $0.02$ & $8$ & $33$ & $0.82$ \\
    $59$, view $2$ & $0.80$ & $0.97$ & $0.90$ & $0.97$ & $1$ & $1$ & $0$ & $0.02$ & $12$ & $40$ & $0.84$ \\
    $59$, view $3$ & $0.79$ & $0.98$ & $0.89$ & $0.98$ & $1$ & $1$ & $0$ & $0.02$ & $8$ & $28$ & $0.87$ \\
    $59$, view $4$ & $0.80$ & $0.97$ & $0.89$ & $0.97$ & $1$ & $1$ & $0$ & $0.02$ & $8$ & $40$ & $0.89$ \\
    \midrule
    \rowcolor{verylightgray} \textbf{Combined} & $0.86$ & $0.98$ & $0.90$ & $0.98$ & $0.99$ & $0.99$ & $0$ & $0.08$ & $73$ & $318$ & $0.94$ \\
    \bottomrule
  \end{tabular}
  \label{tab:quantitative-tracking-evaluation}
\end{table*}

\begin{table}[t]
  \centering
  \caption[Quantitative Tracking Evaluation in 3D]{\emph{Quantitative Tracking Evaluation in 3D}. We test five sequences quantitatively with the metrics specified in~\cref{sec:metrics}. For detailed explanations on abbreviations and metrics, 
  please refer to our supplemental material.
  Upwards and downwards arrows represent whether a higher or lower value is better, respectively.
  See text for a discussion of the results.
  }
  \small
  \begin{tabular}{c|ccccc}
    \toprule
    Seq. & MOTA \hspace{-1.5mm}$\uparrow$ & MT \hspace{-1.5mm}$\uparrow$ & ML \hspace{-1.5mm}$\downarrow$ & IDS \hspace{-1.5mm}$\downarrow$ & Frag \hspace{-1.5mm}$\downarrow$ \\
    \midrule
    $11$ & $0.92$ & $1$     & $0$ & $0$ & $173$ \\
    $19$ & $0.89$ & $0.90$  & $0$ & $0$ & $214$ \\
    $30$ & $0.92$ & $1$  & $0$ & $0$ & $225$ \\
    $48$ & $0.93$ & $1$     & $0$ & $0$ & $245$ \\
    $59$ & $0.57$ & $0.60$  & $0$ & $8$ & $334$ \\
    \midrule
    \rowcolor{verylightgray} \textbf{Comb.} & $0.85$ & $0.90$ & $0$ & $8$ & $1191$ \\
    \bottomrule
  \end{tabular}
  \label{tab:quantitative-tracking-evaluation-3D}
\end{table}
\subheading{Quantitative Tracking Evaluation}
We test our framework quantitatively in 2D and 3D on five video sequences from 3D-POP, cf.~\cref{sec:dataset:3d-pop}.
Each sequence contains ten pigeons ($50$ objects in total, $200$ in 2D) and~$10053$ frames ($40212$ frames in 2D).
Since the sequences contain small gaps due to missed detections in motion capture (see~\citeA{3D-POP} for more details), we use linear interpolation to fill all gaps before evaluation.
For evaluation we use ViTPose* (cf.~\cref{sec:framework}; the most accurate model from~\cref{sec:keypoint_estimation}) and the metrics specified in~\cref{sec:metrics}.
Note that for sequence 59, we remove the first 3 seconds (90 frames) since 2 pigeons are initially outside the tracking volume which causes the first frame identity matching (see~\cref{sec:framework}) to fail.

Detailed 2D results for a detection confidence threshold of~$0.5$ are shown in~\cref{tab:quantitative-tracking-evaluation}.
Overall, we achieve good results with our framework on the 2D video sequences including a HOTA of~$86\%$, $98\%$ multi-object tracking accuracy (MOTA), $90\%$ multi-object tracking precision (MOTP), a recall of~$98\%$, $99\%$ precision, $99\%$ mostly tracked (MT), and $0\%$ mostly lost (ML) trajectories, $0.08$ false positives per frame (FPF), and a IDF1 of~$94\%$ (metrics specified in~\cref{sec:metrics} and our supplemental material).

In~\cref{tab:quantitative-tracking-evaluation-3D} we report detailed 3D tracking results of the bottom keel joint for the five sequences where we set the maximum allowed distance between detections and ground truth positions in~\citeauthor{dendorfer2020motchallengeevalkit} to $30$ mm.
We choose $30$ mm as this threshold is well within the body size of a pigeon, while taking into account the possible distance an individual can move within one frame.
Overall, we achieve good 3D results with~\paperabr including~$85\%$ multi-object tracking accuracy (MOTA), $90\%$ mostly tracked (MT), and~$0\%$ mostly lost (ML) trajectories (metrics specified in~\cref{sec:metrics} and our supplemental material).

\begin{table}[t]
\centering
\caption{
    {\em 2D Inference Speed.}
    Benchmark for the complete pipelines (including data loading, model loading, inference, data saving).
    We report the inference speed (fps) for the 2D models, cf.~\cref{sec:framework}.
    Best results per column in bold.
    See text for a discussion of the results.
    }
\begin{tabular}{c|c|c|c|c}
\toprule
Method / Num. of Ind. &  $1$  & $2$  & $5$  & $10$   \\
\midrule
KP-RCNN & $\mathbf{7.50}$ & $\mathbf{8.07}$ & $\mathbf{8.02}$ & $\mathbf{6.22}$ \\
DLC* & $3.85$ & $3.42$ & $2.75$ & $2.03$\\
ViTPose* & $3.22$ & $2.61$ & $1.57$ & $0.99$\\
\bottomrule
\end{tabular}
\label{table:InferenceSpeed_2D}
\end{table}
\begin{table}[t]
  \centering
  \caption[2D Inference Speed]{\emph{2D Inference Speed}. Benchmark for our in-memory pipeline using the KeypointRCNN, cf.~\cref{sec:framework}. We benchmark our pipeline with our video sequences preloaded in memory and report values for different batch sizes.
  }
  \begin{tabular}{c|c}
    \toprule
    batch & frame rate [fps] \\
        
    \begin{tabular}{c}
    size \\
    \midrule
    $1$\\
    $2$\\
    $4$\\
    $8$\\
    $16$\\
    \end{tabular} &
        
    \begin{tabular}{cccc}
    $1$ pigeon & $2$ pigeons & $5$ pigeons & $10$ pigeons \\
    \midrule
    $8.24$ & $8.13$ & $8.03$ & $6.77$ \\
    $8.70$ & $8.54$ & $8.27$ & $6.91$ \\
    $8.90$ & $8.81$ & $8.43$ & $7.09$ \\
    $9.10$ & $8.96$ & $8.61$ & $7.17$ \\
    $\textbf{9.45}$ & $\textbf{9.29}$ & $\textbf{8.88}$ & $\textbf{7.29}$ \\
    \end{tabular} \\
        
    \bottomrule
  \end{tabular}
        
  \label{tab:benchmark_in-memory_batch_size}
\end{table}
\begin{table}[t]
\centering
\caption{
    {\em 3D Inference Speed.}
    Benchmark for the complete pipelines (including data loading, model loading, inference, data saving).
    We report the inference speed (fps) for the 3D models.
    Best results per column in bold, \paperabr versions highlighted in gray.
    See text for a discussion of the results.
    }
\begin{tabular}{c|c|c|c|c}
\toprule
Method / Num. of Ind. &  $1$  & $2$  & $5$  & $10$   \\
\midrule
\rowMarked 3D-KP-RCNN & $\mathbf{1.89}$ & $\mathbf{1.84}$ & $\mathbf{1.73}$ & $\mathbf{1.59}$ \\
\rowMarked 3D-DLC* & $0.93$ & $0.84$ & $0.65$ & $0.47$ \\
\rowMarked 3D-ViTPose* & $0.79$ & $0.64$ & $0.38$ & $0.24$ \\
\midrule
LToHP (cf.~\cref{sec:keypoint_estimation}) & $0.83$ & $0.44$ &  $0.17$ &  $0.08$\\
\bottomrule
\end{tabular}
\label{table:InferenceSpeed}
\end{table}
\subheading{Inference Speed}
Finally, we benchmark the inference speed of the pipeline, and we show that~\paperabr can estimate 2D and 3D postures at interactive speeds (defined by $\geq1$ fps).
\cref{table:InferenceSpeed_2D,table:InferenceSpeed} provide detailed inference speed estimates for different numbers of individuals for 2D and 3D respectively, and we see that inference speed decreases with increasing number of individuals across all models (at most by $2.23$ fps for ViTPose* in 2D, cf.~\cref{table:InferenceSpeed_2D}, and $0.75$ fps for LToHP in 3D, cf.~\cref{table:InferenceSpeed}).
Overall, we see that the mean inference speed is the fastest for the KeypointRCNN, reaching $7.5$ fps in 2D and $1.76$ fps in 3D, cf.~\cref{table:2DError,table:3DError} respectively.

To push the inference speed of the KeypointRCNN even further, 
we also benchmark the scenario where we pre-load the video sequence in memory and are thus independent of disk I/O, with otherwise the same procedure, see~\cref{tab:benchmark_in-memory_batch_size} for results. We report values for batch sizes up to $16$, restricted by the hardware that we use, cf.~\cref{sec:metrics}. 
The maximum speed is at a batch size of $16$ with an interactive speed of about $7-9$ fps depending on the number of pigeons present in the video sequence.

We conclude that researchers prioritizing inference speed for multi-animal posture estimation and tracking may consider the KeypointRCNN for the pose estimation module in~\paperabr.

The speed evaluation shows that our pipeline can potentially be applied to closed-loop experiments (see~\citeA{naik2021xrforall}), based on the requirements of the researcher. For example, if an experiment requires general position and orientation of pigeons in closed-loop, inference speeds of $1.76$ fps (cf.~\cref{table:3DError}; can be pushed even further by preloading the data in memory and processing batches, cf.~\cref{tab:benchmark_in-memory_batch_size}) might be sufficient. However, we do note that the current inference speed estimates do not include video acquisition time, so researchers considering such applications will need to develop a multi-view video acquisition framework independently. 

There is another framework that also performs 2D keypoint prediction of complex poses and tracking:
SLEAP~\cite{SLEAP}.
Their inference speed benchmark procedure and hardware are comparable to~\paperabr, cf.~\cref{sec:metrics}.
A rough comparison yields that
SLEAP \cite{SLEAP} is
about an order of magnitude faster than the KeypointRCNN (SLEAP up to $\sim800$ fps; numbers read off from \citeA{SLEAP}, Figs. 2b, 3e and Extended Data Fig. 6c). Considering the fact that the image resolution provided in 3D-POP is higher than the one of the flies and mice ($3840\times2160$ px vs. $1280\times1024$ px) and thus we process more data through the whole pipeline.
While our framework solves the substantially harder task of a `generalist' approach of training a single model that works on all datasets, SLEAP uses a `specialist' paradigm where small, lightweight models have just enough representational capacity to generalize to the low variability typically found in scientific data \cite{SLEAP}.
The approach of our framework comes with an additional cost of computing resource requirements.
However, we hope to offer a framework that works with both low and high variability data at the same time. Depending on the application, one can easily change the pose estimator of our framework (cf.~\cref{sec:framework,fig:framework}) to achieve frame rates comparable to SLEAP.
%

\section{Applications}
\label{sec:Applications}
We showcase the flexibility of~\paperabr by presenting two domain shifts.
First we show that~\paperabr can be trained on annotated data that contains only single individuals and applied to multi-animal data which can reduce the annotation effort needed for new species or experimental setups (also see our supplemental material for 2D single mouse and cowbird pose estimation).
Secondly, we show that~\paperabr is robust to an indoor to outdoor environment domain shift by applying a model trained on indoor data to data from outdoors without further fine-tuning.

\begin{table}[t]
\centering
\caption{
    {\em Quantitative Results for Our Single to Multi-Aninal Domain Shift.}
    We report RMSE and its median ($px$ and $mm$ in 2D and 3D respectively), PCK05 ($\%$) and PCK10 ($\%$) for estimated 2D and 3D posture from the 3D-POP dataset using the KeypointRCNN trained with single pigeon data.
    Upwards and downwards arrows represent whether a higher or lower value is better, respectively.
    We report results for sequences containing different number of individuals ($1$, $2$, $5$, and $10$), cf.~\cref{sec:dataset:use-cases}.
    }
\begin{tabular}{c|c|c|c|c}
\toprule
Metric / Num. of Ind. & $1$  & $2$  & $5$  & $10$   \\
\midrule
 2D \\
\midrule
RMSE ($px$) $\downarrow$   & $8.6$ & $20.1$ & $57.2$ & $272.5$ \\
Median ($px$) $\downarrow$  & $4.3$ & $6.0$ & $7.7$ & $17.9$ \\
PCK05 ($\%$) $\uparrow$  & $90.5$ & $76.9$ & $66.7$ & $42.9$ \\
PCK10 ($\%$) $\uparrow$  & $98.7$ & $93.4$ & $83.6$ & $53.9$ \\
\midrule
 3D \\
 \midrule
RMSE ($mm$) $\downarrow$  & $11.1$ & $26.9$ & $93.2$ & $434.3$ \\
Median ($mm$) $\downarrow$  & $6.9$ & $6.0$ & $15.4$ & $246.7$ \\
PCK05 ($\%$) $\uparrow$  & $70.4$ & $54.1$ & $30.2$ & $11.4$ \\
PCK10 ($\%$) $\uparrow$  & $94.9$ & $82.4$ & $60.3$ & $19.7$ \\

\bottomrule
\end{tabular}
\label{table:SinglePigeon}
\end{table}
\subsection{Single to Multi-Animal Domain Shift}
\label{sec:single-multi-domain-shift}

We train the KeypointRCNN (cf.~\cref{sec:framework:application}) for $30$ epochs
on the single-pigeon dataset specified in~\cref{sec:dataset:use-cases}.
Results can be found in~\cref{table:SinglePigeon}, showing difference in error across different number of individuals.

Overall, the single pigeon model performs well in 2D, but not as well in 3D, with the model not being able to generalize for 3D tracking of $10$ pigeons.
For sequences with $1$ and $2$ individuals, the performance is similar to using a multi-animal dataset for both 2D and 3D (cf.~\cref{table:2DError,table:3DError,table:SinglePigeon}). 
For example, when comparing results of $2$ individuals using the single pigeon model (\cref{table:SinglePigeon}) with the KeypointRCNN trained with multi-pigeon data (averaged over $1$, $2$, $5$, $10$ individuals, \cref{table:3DError}), we achieve a RMSE for the single-pigeon model of $26.9$ mm vs. multi-pigeon of $25.0$ mm with a median for single-pigeon of $6.0$ mm vs. multi-pigeon of $9.4$ mm.

For sequences with $5$ and $10$ individuals, performance differs.
In 2D, we observe outliers as evident from the large RMSE values ($5$ individuals: $57.2$ px, $10$ individuals: $272.5$ px), but from the median and PCK values from the multi-pigeon model (median of $5.7$ px, PCK10 of $95.4\%$), the single-pigeon model show comparable accuracy for $5$ individuals (median of $7.7$ px, PCK10 of $83.6\%$), and good accuracy for $10$ individuals (median of $17.9$ px, PCK10 of $53.9\%$). 

For 3D posture estimation, we expect accuracy to propagate from 2D estimation errors, as shown in the multi-animal model evaluation (cf.~\cref{table:2DError,table:3DError}), but we show that while 3D error is still low at $15.4$ mm (median error) for 5 individuals, the model fails to generalize in 3D for 10 pigeons (median of $246.7$ mm).

We think there are two main reasons that the model fails to generalize to 10 pigeons.
Firstly, the detection of the bird individuals is less robust with the single pigeon model, where 10 pigeons are not always detected from all frames, and can affect the first frame identity matching and subsequent 2D tracking in the~\paperabr pipeline.
So, an incorrect 2D tracklet in one view can already increase the 3D error while additional ID switches in further camera views further deteriorate the 3D accuracy.
This is reflected in~\cref{table:SinglePigeon} where the median error is $\sim16\times$ higher for 10 compared to 5 individuals in 3D while it is ``only'' $\sim2\times$ higher in 2D; the 2D errors from different views potentiate in 3D.
Another reason is occlusions, where the model struggles to predict keypoints when the objects are too occluded, which is often the case in the 10 pigeon sequences.
This shortcoming is also expected since the model was only trained on single pigeon data.

Nevertheless, we highlight that training a model with only single pigeon data can allow 2D and 3D posture estimation of up to 5 pigeons, which can simplify the domain shift to new species or systems, because annotating single animal data is less labour intensive than multi-animal annotations.

While less reliable in 3D, we show that the single-pigeon can predict keypoints in 2D reliably, so if researchers wish to annotate multi-individual data, the single-individual model can also be used as a pre-labelling tool. This can further reduce annotation time by first predicting keypoints from the 2D frame and manually correcting faulty detections, similar to methodologies provided in~\citeA{SLEAP,DLC,DPK}.

\begin{table*}[t]
\centering
\caption{
    {\em Quantitative Evaluation of 3D Pigeon Poses in Our Novel Wild-MuPPET dataset.}
    We report RMSE and its median ($mm$),
    PCK05 ($\%$) and PCK10 ($\%$) for the 3D poses of pigeons in the wild, on the 100 test frames in the Wild-MuPPET dataset cf.~\cref{sec:dataset:use-cases}.
    Wild-ViTPose and Wild-DLC are models trained on masked images from 3D-POP using ViTPose~\protect\cite{ViTPose} and DLC~\protect\cite{DLC} respectively, without additional annotations from the wild. DLC-Fine-tuned and DLC-Scratch are trained on sampled images from Wild-MuPPET training set (cf.~\cref{sec:dataset:use-cases}), with DLC-Fine-tuned using Wild-DLC as initial weights.
    See text for a discussion of the results.
}
\begin{tabular}{c|c|c|c|c}
\toprule
Metric / Method & Wild-ViTPose & Wild-DLC & DLC-Fine-tuned& DLC-Scratch\\
\midrule
RMSE ($mm$) $\downarrow$ & $166.0$ & $53.4$ & $58.2$ & $\mathbf{45.0}$\\
Median ($mm$) $\downarrow$ & $146.0$ & $15.0$ & $\mathbf{11.4}$ & $12.7$\\
PCK05 ($\%$) $\uparrow$ & $0$ & $25.1$ & $\mathbf{44.7}$ & $40.1$ \\
PCK10 ($\%$) $\uparrow$ & $0.2$ & $74.4$ & $\mathbf{81.3}$ & $77.4$ \\
\bottomrule
\end{tabular}
\label{table:3DError_in_the_wild}
\end{table*}
\begin{figure*}[ht!]
  \centering
  \includegraphics[width=0.499\linewidth]{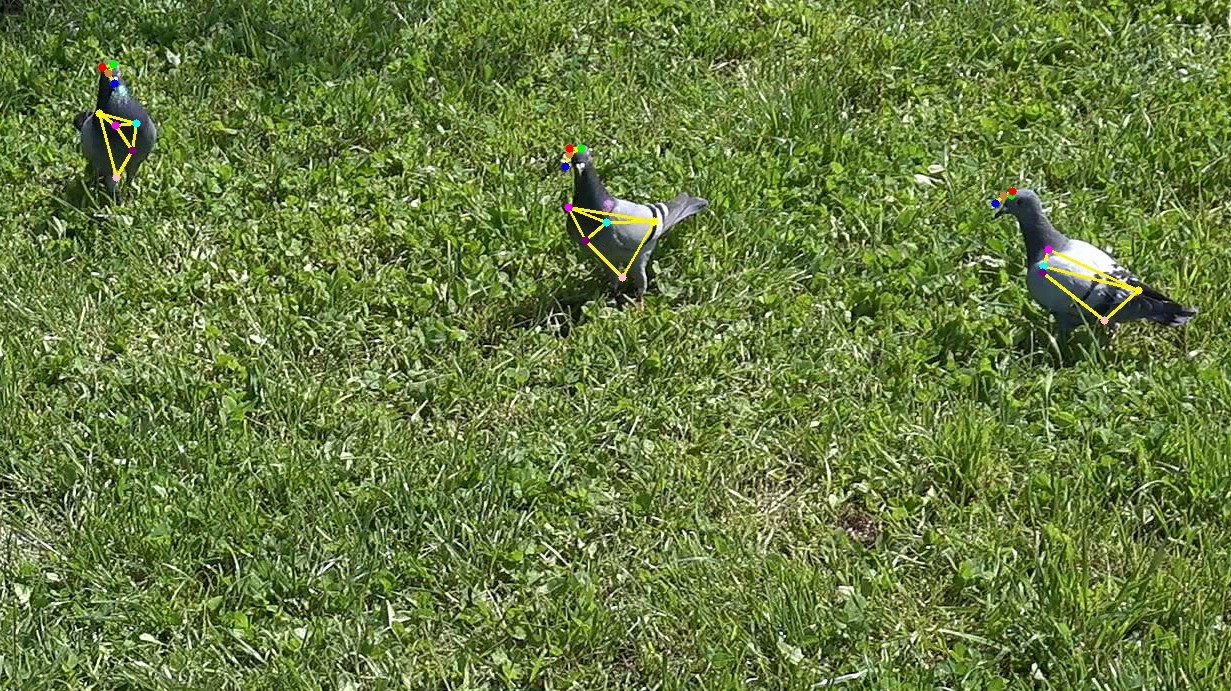}\hfill\includegraphics[width=0.499\linewidth]{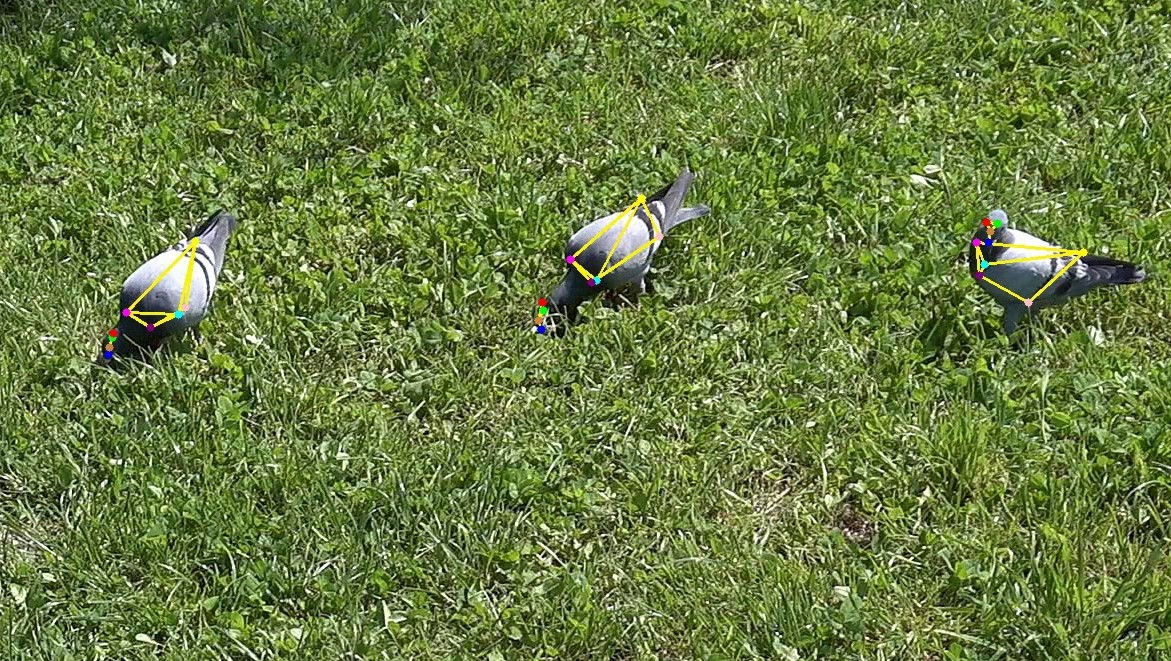}
  \includegraphics[width=0.499\linewidth]{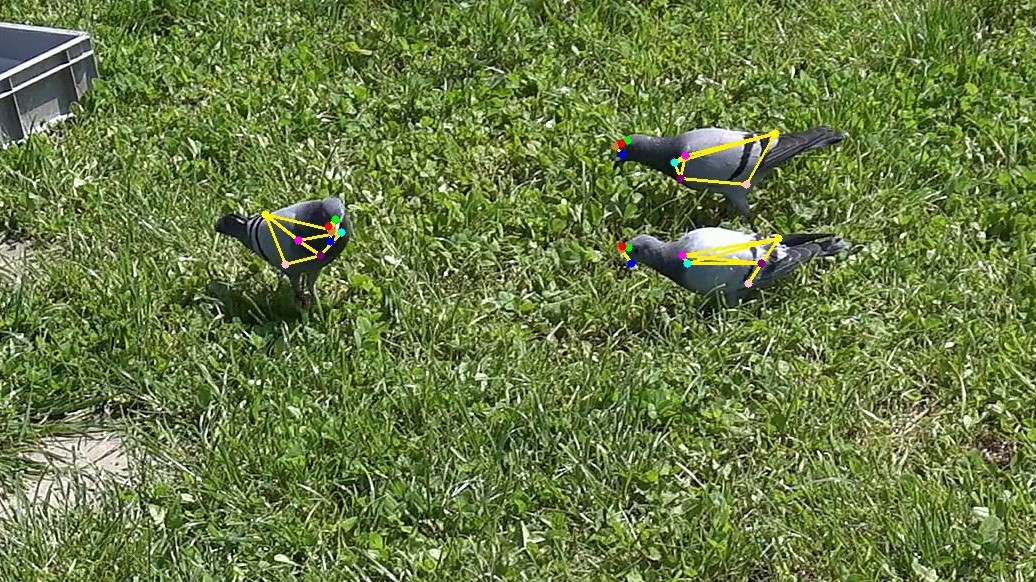}\hfill\includegraphics[width=0.499\linewidth]{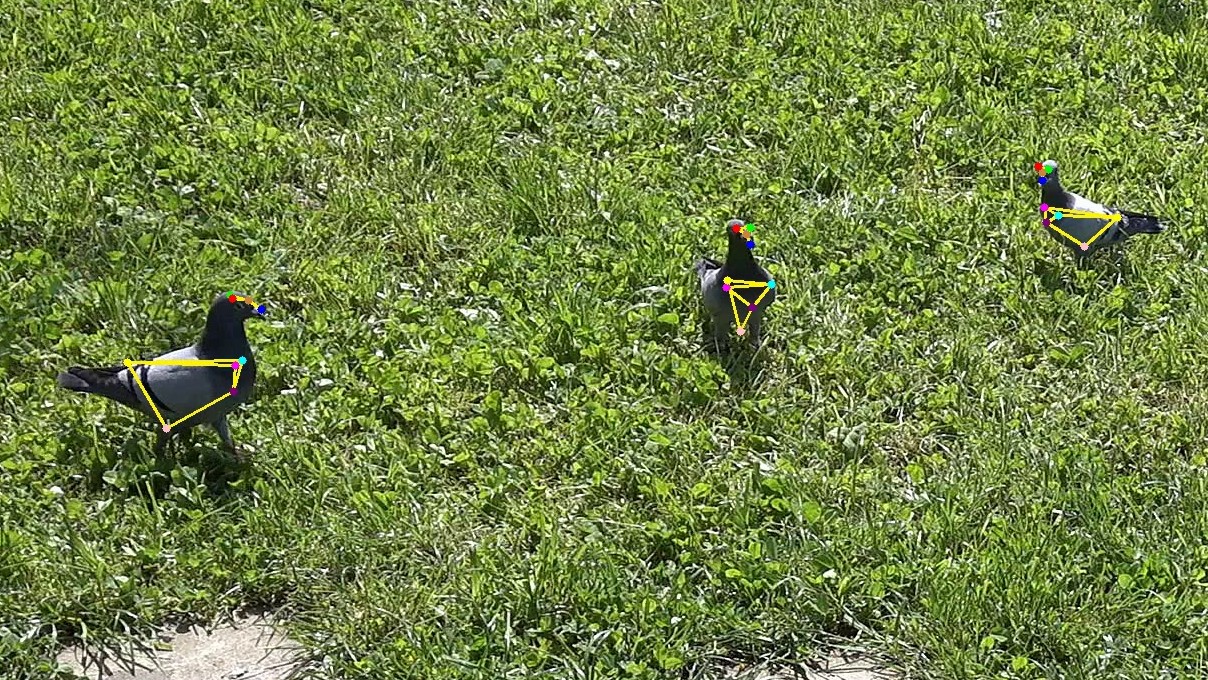}
  \caption[Qualitative Results of Pigeons in the Wild]{\emph{Qualitative Results of Pigeons in the Wild}.
  Example frames for 3D multi-pigeon pose estimation and tracking in the wild, reprojected to 2D view.
  Notably, we did not fine-tune~\paperabr (here with Wild-DLC) on our novel~\dataabr data recorded in the wild, cf.~\cref{sec:framework:application}.
  }
  \label{fig:field_data}
\end{figure*}

\subsection{Pigeons in the Wild}
\label{sec:applications:pigeons-in-the-wild}
We train the Wild-ViTPose model for $124$ epochs and Wild-DLC for $93000$ iterations.

In~\cref{table:3DError_in_the_wild} we report quantitative results on the test set of our novel Wild-MuPPET dataset. 
We first show that Wild-ViTPose (ViTPose* is the most accurate model in~\cref{sec:keypoint_estimation}) does not generalize well for pigeons in the wild, compared to Wild-DLC, likely due to differences in augmentation parameters (median error of $146.0$ mm and $15.0$ mm respectively).
However, for Wild-DLC, we show that the model performs well on Wild-MuPPET, with a median accuracy of $15.0$ mm, only with training data of pigeons indoors, cf.~\cref{sec:dataset:use-cases}.
Additionally, we also use Wild-DLC for inference in a 3 pigeon sequence in the wild, which reflects our promising quantitative results, cf.~\cref{fig:field_data} and supplementary video.

To further explore how a model trained on pigeons indoors can aid the domain shift to the wild, we also fine-tune the Wild-DLC model (named DLC-Fine-tuned) using sampled 2D frames from the training set of Wild-MuPPET (see cf.~\cref{sec:dataset:use-cases}). 
To compare whether initializing model weights using data of pigeons indoors can lead to better accuracy in the wild, we also trained a DLC model from scratch, without fine-tuning (named DLC-Scratch), using the same outdoor image dataset, cf.~\cref{sec:dataset:use-cases}.
Fine-tuning takes $61 000$ iterations, and training from scratch takes $99 000$ iterations to reach lowest validation loss.

We show that both fine-tuning and training from scratch improves the performance of Wild-DLC (cf.~\cref{table:3DError_in_the_wild}), and both methods yield comparable accuracy.
However, the fine-tuned model performs slightly better than the model trained from scratch (median of $11.4$ mm vs. $12.7$ mm respectively). 
Finally, we note that while keypoint estimation accuracy in the latter two cases is comparable, fine-tuning requires less iterations for the model to converge, allowing reduced training time for domain shifts across datasets.

All together, our two applications show that~\paperabr is flexible and robust, promising to open up new ways for biologists to study animal collective behaviour in a fine-scaled way with multi-animal 3D posture tracking.
%

%
\begin{figure}[t]
  \centering
  \includegraphics[width=\columnwidth]{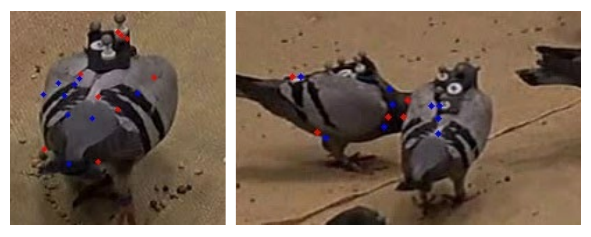}
  \caption[Limitations]{\emph{Limitations}.
  Cropped frames of failure cases from 3D-POP~\cite{3D-POP} data for 2D pose estimation using the KeypointRCNN (cf.~\cref{sec:framework}), due to occlusions. Blue denotes the ground truth, red denotes the prediction.}
  \label{fig:futurework}
\end{figure}
\section{Limitations and Future Work}
\label{sec:futurework}
Keypoint detection can fail e.g. due to self-occlusions or occlusions from other individuals (cf.~\cref{fig:futurework}),
which can affect the triangulation procedure.
This may have caused outliers present in 2D and 3D keypoint evaluation, as indicated by the high RMSE values in contrast to their median errors.
While we use a Kalman filter to smooth 3D posture estimates, the method can fail when there are multiple consecutive frames of error. Other filtering and smoothing methods that consider temporal consistency in an offline fashion can alleviate this problem if online processing is not required (e.g.~\citeA{maDLC,AcinoSet}).

For pigeons in the wild, we limit the pigeon segmentation to~\citeA{SAM} and~\citeA{MaskRCNN} and the tracking to~\citeA{SORT}, other methods available like~\citeA{Bekuzarov_2023_ICCV} and~\citeA{yang2023track} for segmentation and~\citeA{karaev2023cotracker} for tracking might boost the performance. 
Using another tracker might also boost our single to multi-animal domain shift when dealing with 10 individuals.

Finally, our current tracking approach relies on all subjects being present in the first frame for first frame re-identification, as well as all subjects staying in frame for the whole sequence. Future work can improve upon the tracking algorithm e.g. by using visual features for re-identification~\cite{Wojke2018deep,ferreira2020deep,NeTePu}.
%

\section{Conclusion}
\label{sec:conclusion}
In this work we present~\paperabr, a framework to estimate 3D poses of multiple pigeons from a multi-view setup.
We show that our framework allows complex poses and trajectories of multiple pigeons to be tracked reliably in 2D and 3D (cf.~\cref{table:2DError,table:3DError}) at interactive speeds with up to~$9.45$ fps in 2D and $1.89$ fps in 3D.
While our results are comparable to a state of the art 3D pose estimator in terms of median error and Percentage of Correct Keypoints, cf.~\cref{table:3DError}, \paperabr achieves a faster inference speed, cf.~\cref{table:InferenceSpeed_2D,table:InferenceSpeed}, and only relies on training a 2D posture estimation model.
Additionally, we perform the first quantitative tracking evaluation on 3D-POP and obtain good results, cf.~\cref{tab:quantitative-tracking-evaluation,tab:quantitative-tracking-evaluation-3D}.

In applications where a higher accuracy is needed, researchers should prefer 3D-ViTPose* for the pose estimation module of~\paperabr, cf.~\cref{fig:framework}.
Researchers that prioritize inference speed for multi-animal posture estimation and tracking or are interested in the single to multi-animal domain shift may consider the KeypointRCNN for the pose estimation module in~\paperabr.

Finally, we demonstrate that training a pose estimation module on single pigeon training data yields comparable results compared to a model trained on multi-pigeon data for up to 5 pigeons (cf.~\cref{sec:single-multi-domain-shift}), as well as showing that a model trained with indoor data can be generalized to data in the wild, cf.~\cref{sec:applications:pigeons-in-the-wild}.
This highlights the potential of a domain shift to new species and environments without the need for laborious manual annotation. 

\paperabr is the first 3D pose estimation framework for more than four animals that also works with data recorded in the wild, cf.~\cref{sec:framework}.
While previous work~\cite{OpenMonkeyStudio,han2023social,MAMMAL} has demonstrated 3D pose estimation for up to four animals,
\paperabr shows that it is possible to track the 3D poses of up to $10$ pigeons if a 2D posture estimation model and a multi-camera setup is available.
Our work offers a promising and flexible framework opening up new ways for biologists to study animal collective behaviour and we hope that this leads to further systematic progress in the field.

\section{Declarations}
\subheading{Data availability}
During the review process, the datasets generated during and/or analysed during the current study are available from the corresponding author on reasonable request.
Upon acceptance, the datasets generated during and/or analysed during the current study are available in the GitHub repository, \url{https://github.com/alexhang212/3D-MuPPET}.

\subheading{Funding and Competing Interests}
Funded by the Deutsche Forschungsgemeinschaft (DFG, German Research Foundation) under Germany's Excellence Strategy – EXC 2117 – 422037984, and the Federal Ministry of Education and Research (BMBF) within the research program KI4KMU under grant number 01IS23046B (ARGUS).
All authors certify that they have no affiliations with or involvement in any organization or entity with any financial interest or non-financial interest in the subject matter or materials discussed in this manuscript.

\subheading{Acknowledgements}
We thank Lili Karashchuk for the valuable feedback and suggestions in CVPR 2023.
%

\newpage
\bibliographystyle{sn-apacite}
\bibliography{refs}

\begin{thebibliography}{}
\renewcommand{\doi}[1]{\url{https://doi.org/#1}}
\bibcommenthead

\bibitem [\protect \citeauthoryear {%
Altmann%
}{%
Altmann%
}{%
{\protect \APACyear {1974}}%
}]{%
ObservationalStudyofBehaviorSamplingMethods}
\APACinsertmetastar {%
ObservationalStudyofBehaviorSamplingMethods}%
\begin{APACrefauthors}%
Altmann, J.%
\end{APACrefauthors}%
\unskip\
\newblock
\APACrefYearMonthDay{1974}{}{}.
\newblock
{\BBOQ}\APACrefatitle {Observational Study of Behavior: Sampling Methods}
  {Observational study of behavior: Sampling methods}.{\BBCQ}
\newblock
\APACjournalVolNumPages{Behaviour}{49}{3-4}{227 - 266,}
\newblock

\newblock

\PrintBackRefs{\CurrentBib}

\bibitem [\protect \citeauthoryear {%
An%
\ \protect \BOthers {.}}{%
An%
\ \protect \BOthers {.}}{%
{\protect \APACyear {2023}}%
}]{%
MAMMAL}
\APACinsertmetastar {%
MAMMAL}%
\begin{APACrefauthors}%
An, L.%
, Ren, J.%
, Yu, T.%
, Hai, T.%
, Jia, Y.%
\BCBL {} Liu, Y.%
\end{APACrefauthors}%
\unskip\
\newblock
\APACrefYearMonthDay{2023}{}{}.
\newblock
{\BBOQ}\APACrefatitle {Three-dimensional surface motion capture of multiple
  freely moving pigs using MAMMAL} {Three-dimensional surface motion capture of
  multiple freely moving pigs using mammal}.{\BBCQ}
\newblock
\APACjournalVolNumPages{Nature Communications}{14}{1}{7727,}
\newblock

\newblock

\PrintBackRefs{\CurrentBib}

\bibitem [\protect \citeauthoryear {%
Anderson%
\ \BBA {} Perona%
}{%
Anderson%
\ \BBA {} Perona%
}{%
{\protect \APACyear {2014}}%
}]{%
ANDERSON201418}
\APACinsertmetastar {%
ANDERSON201418}%
\begin{APACrefauthors}%
Anderson, D.%
\BCBT {}\ \BBA {} Perona, P.%
\end{APACrefauthors}%
\unskip\
\newblock
\APACrefYearMonthDay{2014}{}{}.
\newblock
{\BBOQ}\APACrefatitle {Toward a Science of Computational Ethology} {Toward a
  science of computational ethology}.{\BBCQ}
\newblock
\APACjournalVolNumPages{Neuron}{84}{1}{18-31,}
\newblock

\newblock

\PrintBackRefs{\CurrentBib}

\bibitem [\protect \citeauthoryear {%
Badger%
\ \protect \BOthers {.}}{%
Badger%
\ \protect \BOthers {.}}{%
{\protect \APACyear {2020}}%
}]{%
3DBirdReconstruction}
\APACinsertmetastar {%
3DBirdReconstruction}%
\begin{APACrefauthors}%
Badger, M.%
, Wang, Y.%
, Modh, A.%
, Perkes, A.%
, Kolotouros, N.%
, Pfrommer, B.G.%
\BDBL {}Daniilidis, K.%
\end{APACrefauthors}%
\unskip\
\newblock
\APACrefYearMonthDay{2020}{}{}.
\newblock
{\BBOQ}\APACrefatitle {3D Bird Reconstruction: A Dataset, Model, and Shape
  Recovery from a Single View} {3d bird reconstruction: A dataset, model, and
  shape recovery from a single view}.{\BBCQ}
\newblock
 \APACrefbtitle {Eur. Conf. Comput. Vis.} {Eur. conf. comput. vis.}\ (\BPGS\
  1--17).
\PrintBackRefs{\CurrentBib}

\bibitem [\protect \citeauthoryear {%
Bala%
\ \protect \BOthers {.}}{%
Bala%
\ \protect \BOthers {.}}{%
{\protect \APACyear {2020}}%
}]{%
OpenMonkeyStudio}
\APACinsertmetastar {%
OpenMonkeyStudio}%
\begin{APACrefauthors}%
Bala, P.C.%
, Eisenreich, B.R.%
, Yoo, S.B.M.%
, Hayden, B.Y.%
, Park, H.S.%
\BCBL {} Zimmermann, J.%
\end{APACrefauthors}%
\unskip\
\newblock
\APACrefYearMonthDay{2020}{}{}.
\newblock
{\BBOQ}\APACrefatitle {Automated markerless pose estimation in freely moving
  macaques with OpenMonkeyStudio} {Automated markerless pose estimation in
  freely moving macaques with openmonkeystudio}.{\BBCQ}
\newblock
\APACjournalVolNumPages{Nat. Commun.}{11}{}{4560,}
\newblock

\newblock

\PrintBackRefs{\CurrentBib}

\bibitem [\protect \citeauthoryear {%
Bekuzarov%
, Bermudez%
, Lee%
\BCBL {}\ \BBA {} Li%
}{%
Bekuzarov%
\ \protect \BOthers {.}}{%
{\protect \APACyear {2023}}%
}]{%
Bekuzarov_2023_ICCV}
\APACinsertmetastar {%
Bekuzarov_2023_ICCV}%
\begin{APACrefauthors}%
Bekuzarov, M.%
, Bermudez, A.%
, Lee, J\BHBI Y.%
\BCBL {} Li, H.%
\end{APACrefauthors}%
\unskip\
\newblock
\APACrefYearMonthDay{2023}{October}{}.
\newblock
{\BBOQ}\APACrefatitle {XMem++: Production-level Video Segmentation From Few
  Annotated Frames} {Xmem++: Production-level video segmentation from few
  annotated frames}.{\BBCQ}
\newblock
 \APACrefbtitle {Proceedings of the IEEE/CVF International Conference on
  Computer Vision (ICCV)} {Proceedings of the ieee/cvf international conference
  on computer vision (iccv)}\ (\BPG~635-644).
\PrintBackRefs{\CurrentBib}

\bibitem [\protect \citeauthoryear {%
Berman%
}{%
Berman%
}{%
{\protect \APACyear {2018}}%
}]{%
Berman2018}
\APACinsertmetastar {%
Berman2018}%
\begin{APACrefauthors}%
Berman, G.J.%
\end{APACrefauthors}%
\unskip\
\newblock
\APACrefYearMonthDay{2018}{}{}.
\newblock
{\BBOQ}\APACrefatitle {Measuring behavior across scales} {Measuring behavior
  across scales}.{\BBCQ}
\newblock
\APACjournalVolNumPages{BMC Biol.}{16}{23}{,}
\newblock

\newblock

\PrintBackRefs{\CurrentBib}

\bibitem [\protect \citeauthoryear {%
Bernardin%
\ \BBA {} Stiefelhagen%
}{%
Bernardin%
\ \BBA {} Stiefelhagen%
}{%
{\protect \APACyear {2008}}%
}]{%
CLEARMOT}
\APACinsertmetastar {%
CLEARMOT}%
\begin{APACrefauthors}%
Bernardin, K.%
\BCBT {}\ \BBA {} Stiefelhagen, R.%
\end{APACrefauthors}%
\unskip\
\newblock
\APACrefYearMonthDay{2008}{}{}.
\newblock
{\BBOQ}\APACrefatitle {Evaluating multiple object tracking performance: the
  clear mot metrics} {Evaluating multiple object tracking performance: the
  clear mot metrics}.{\BBCQ}
\newblock
\APACjournalVolNumPages{EURASIP Journal on Image and Video
  Processing}{2008}{}{1--10,}
\newblock

\newblock

\PrintBackRefs{\CurrentBib}

\bibitem [\protect \citeauthoryear {%
Bernshtein%
}{%
Bernshtein%
}{%
{\protect \APACyear {1967}}%
}]{%
bernshtein1967co}
\APACinsertmetastar {%
bernshtein1967co}%
\begin{APACrefauthors}%
Bernshtein, N.%
\end{APACrefauthors}%
\unskip\
\newblock
\APACrefYear{1967}.
\newblock
\APACrefbtitle {The Co-ordination and Regulation of Movements} {The
  co-ordination and regulation of movements}.
\newblock
\APACaddressPublisher{}{Pergamon Press}.
\PrintBackRefs{\CurrentBib}

\bibitem [\protect \citeauthoryear {%
Bewley%
, Ge%
, Ott%
, Ramos%
\BCBL {}\ \BBA {} Upcroft%
}{%
Bewley%
\ \protect \BOthers {.}}{%
{\protect \APACyear {2016}}%
}]{%
SORT}
\APACinsertmetastar {%
SORT}%
\begin{APACrefauthors}%
Bewley, A.%
, Ge, Z.%
, Ott, L.%
, Ramos, F.%
\BCBL {} Upcroft, B.%
\end{APACrefauthors}%
\unskip\
\newblock
\APACrefYearMonthDay{2016}{}{}.
\newblock
{\BBOQ}\APACrefatitle {Simple online and realtime tracking} {Simple online and
  realtime tracking}.{\BBCQ}
\newblock
 \APACrefbtitle {IEEE Int. Conf. Image Process.} {Ieee int. conf. image
  process.}\ (\BPG~3464-3468).
\PrintBackRefs{\CurrentBib}

\bibitem [\protect \citeauthoryear {%
Biggs%
, Roddick%
, Fitzgibbon%
\BCBL {}\ \BBA {} Cipolla%
}{%
Biggs%
\ \protect \BOthers {.}}{%
{\protect \APACyear {2019}}%
}]{%
CreaturesGreatAndSMAL}
\APACinsertmetastar {%
CreaturesGreatAndSMAL}%
\begin{APACrefauthors}%
Biggs, B.%
, Roddick, T.%
, Fitzgibbon, A.%
\BCBL {} Cipolla, R.%
\end{APACrefauthors}%
\unskip\
\newblock
\APACrefYearMonthDay{2019}{}{}.
\newblock
{\BBOQ}\APACrefatitle {Creatures Great and SMAL: Recovering the Shape and
  Motion of Animals from Video} {Creatures great and smal: Recovering the shape
  and motion of animals from video}.{\BBCQ}
\newblock
 \APACrefbtitle {Proceedings of the Asian Conference on Computer Vision}
  {Proceedings of the asian conference on computer vision}\ (\BPGS\ 3--19).
\PrintBackRefs{\CurrentBib}

\bibitem [\protect \citeauthoryear {%
Bolaños%
\ \protect \BOthers {.}}{%
Bolaños%
\ \protect \BOthers {.}}{%
{\protect \APACyear {2021}}%
}]{%
bolanos2021a3}
\APACinsertmetastar {%
bolanos2021a3}%
\begin{APACrefauthors}%
Bolaños, L.A.%
, Xiao, D.%
, Ford, N.L.%
, LeDue, J.M.%
, Gupta, P.K.%
, Doebeli, C.%
\BDBL {}Murphy, T.H.%
\end{APACrefauthors}%
\unskip\
\newblock
\APACrefYearMonthDay{2021}{}{}.
\newblock
{\BBOQ}\APACrefatitle {A three-dimensional virtual mouse generates synthetic
  training data for behavioral analysis} {A three-dimensional virtual mouse
  generates synthetic training data for behavioral analysis}.{\BBCQ}
\newblock
\APACjournalVolNumPages{Nat. Methods}{18}{}{378--381,}
\newblock

\newblock

\PrintBackRefs{\CurrentBib}

\bibitem [\protect \citeauthoryear {%
Bridgeman%
, Volino%
, Guillemaut%
\BCBL {}\ \BBA {} Hilton%
}{%
Bridgeman%
\ \protect \BOthers {.}}{%
{\protect \APACyear {2019}}%
}]{%
bridgeman2019multi-person}
\APACinsertmetastar {%
bridgeman2019multi-person}%
\begin{APACrefauthors}%
Bridgeman, L.%
, Volino, M.%
, Guillemaut, J\BHBI Y.%
\BCBL {} Hilton, A.%
\end{APACrefauthors}%
\unskip\
\newblock
\APACrefYearMonthDay{2019}{June}{}.
\newblock
{\BBOQ}\APACrefatitle {Multi-Person 3D Pose Estimation and Tracking in Sports}
  {Multi-person 3d pose estimation and tracking in sports}.{\BBCQ}
\newblock
 \APACrefbtitle {IEEE Conf. Comput. Vis. Pattern Recog. Worksh.} {Ieee conf.
  comput. vis. pattern recog. worksh.}
\PrintBackRefs{\CurrentBib}

\bibitem [\protect \citeauthoryear {%
Chard%
\ \BBA {} Gundlach%
}{%
Chard%
\ \BBA {} Gundlach%
}{%
{\protect \APACyear {1938}}%
}]{%
chard1938structure}
\APACinsertmetastar {%
chard1938structure}%
\begin{APACrefauthors}%
Chard, R.D.%
\BCBT {}\ \BBA {} Gundlach, R.H.%
\end{APACrefauthors}%
\unskip\
\newblock
\APACrefYearMonthDay{1938}{}{}.
\newblock
{\BBOQ}\APACrefatitle {The structure of the eye of the homing pigeon.} {The
  structure of the eye of the homing pigeon.}{\BBCQ}
\newblock
\APACjournalVolNumPages{Journal of Comparative Psychology}{25}{2}{249,}
\newblock

\newblock

\PrintBackRefs{\CurrentBib}

\bibitem [\protect \citeauthoryear {%
Chen%
\ \protect \BOthers {.}}{%
Chen%
\ \protect \BOthers {.}}{%
{\protect \APACyear {2020}}%
}]{%
Chen2020si}
\APACinsertmetastar {%
Chen2020si}%
\begin{APACrefauthors}%
Chen, X.%
, Zhai, H.%
, Liu, D.%
, Li, W.%
, Ding, C.%
, Xie, Q.%
\BCBL {} Han, H.%
\end{APACrefauthors}%
\unskip\
\newblock
\APACrefYearMonthDay{2020}{}{}.
\newblock
{\BBOQ}\APACrefatitle {SiamBOMB: A Real-time AI-based System for Home-cage
  Animal Tracking, Segmentation and Behavioral Analysis} {Siambomb: A real-time
  ai-based system for home-cage animal tracking, segmentation and behavioral
  analysis}.{\BBCQ}
\newblock
 \APACrefbtitle {Proceedings of the Twenty-Ninth International Conference on
  International Joint Conferences on Artificial Intelligence} {Proceedings of
  the twenty-ninth international conference on international joint conferences
  on artificial intelligence}\ (\BPGS\ 5300--5302).
\PrintBackRefs{\CurrentBib}

\bibitem [\protect \citeauthoryear {%
Couzin%
\ \BBA {} Heins%
}{%
Couzin%
\ \BBA {} Heins%
}{%
{\protect \APACyear {2023}}%
}]{%
couzin2022emerging}
\APACinsertmetastar {%
couzin2022emerging}%
\begin{APACrefauthors}%
Couzin, I.D.%
\BCBT {}\ \BBA {} Heins, C.%
\end{APACrefauthors}%
\unskip\
\newblock
\APACrefYearMonthDay{2023}{}{}.
\newblock
{\BBOQ}\APACrefatitle {Emerging technologies for behavioral research in
  changing environments} {Emerging technologies for behavioral research in
  changing environments}.{\BBCQ}
\newblock
\APACjournalVolNumPages{Trends in Ecology \& Evolution}{}{}{,}
\newblock

\newblock

\PrintBackRefs{\CurrentBib}

\bibitem [\protect \citeauthoryear {%
Dell%
\ \protect \BOthers {.}}{%
Dell%
\ \protect \BOthers {.}}{%
{\protect \APACyear {2014}}%
}]{%
DELL2014417}
\APACinsertmetastar {%
DELL2014417}%
\begin{APACrefauthors}%
Dell, A.I.%
, Bender, J.A.%
, Branson, K.%
, Couzin, I.D.%
, {de Polavieja}, G.G.%
, Noldus, L.P.%
\BDBL {}Brose, U.%
\end{APACrefauthors}%
\unskip\
\newblock
\APACrefYearMonthDay{2014}{}{}.
\newblock
{\BBOQ}\APACrefatitle {Automated image-based tracking and its application in
  ecology} {Automated image-based tracking and its application in
  ecology}.{\BBCQ}
\newblock
\APACjournalVolNumPages{Trends in Ecology \& Evolution}{29}{7}{417-428,}
\newblock

\newblock

\PrintBackRefs{\CurrentBib}

\bibitem [\protect \citeauthoryear {%
Dendorfer%
}{%
Dendorfer%
}{%
{\protect \APACyear {2020}}%
}]{%
dendorfer2020motchallengeevalkit}
\APACinsertmetastar {%
dendorfer2020motchallengeevalkit}%
\begin{APACrefauthors}%
Dendorfer, P.%
\end{APACrefauthors}%
\unskip\
\newblock
\APACrefYearMonthDay{2020}{}{}.
\newblock
\APACrefbtitle {MOTChallengeEvalKit.} {Motchallengeevalkit.}
\newblock
\APAChowpublished
  {\url{https://github.com/dendorferpatrick/MOTChallengeEvalKit}}.
\PrintBackRefs{\CurrentBib}

\bibitem [\protect \citeauthoryear {%
Dendorfer%
\ \protect \BOthers {.}}{%
Dendorfer%
\ \protect \BOthers {.}}{%
{\protect \APACyear {2021}}%
}]{%
dendorfer2021motchallenge}
\APACinsertmetastar {%
dendorfer2021motchallenge}%
\begin{APACrefauthors}%
Dendorfer, P.%
, Osep, A.%
, Milan, A.%
, Schindler, K.%
, Cremers, D.%
, Reid, I.%
\BDBL {}Leal-Taix{\'e}, L.%
\end{APACrefauthors}%
\unskip\
\newblock
\APACrefYearMonthDay{2021}{}{}.
\newblock
{\BBOQ}\APACrefatitle {Motchallenge: A benchmark for single-camera multiple
  target tracking} {Motchallenge: A benchmark for single-camera multiple target
  tracking}.{\BBCQ}
\newblock
\APACjournalVolNumPages{Int. J. Comput. Vis.}{129}{4}{845--881,}
\newblock

\newblock

\PrintBackRefs{\CurrentBib}

\bibitem [\protect \citeauthoryear {%
Deng%
\ \protect \BOthers {.}}{%
Deng%
\ \protect \BOthers {.}}{%
{\protect \APACyear {2009}}%
}]{%
ImageNet}
\APACinsertmetastar {%
ImageNet}%
\begin{APACrefauthors}%
Deng, J.%
, Dong, W.%
, Socher, R.%
, Li, L\BHBI J.%
, Li, K.%
\BCBL {} Fei-Fei, L.%
\end{APACrefauthors}%
\unskip\
\newblock
\APACrefYearMonthDay{2009}{}{}.
\newblock
{\BBOQ}\APACrefatitle {ImageNet: A large-scale hierarchical image database}
  {Imagenet: A large-scale hierarchical image database}.{\BBCQ}
\newblock
 \APACrefbtitle {IEEE Conf. Comput. Vis. Pattern Recog.} {Ieee conf. comput.
  vis. pattern recog.}\ (\BPG~248-255).
\PrintBackRefs{\CurrentBib}

\bibitem [\protect \citeauthoryear {%
Dunn%
\ \protect \BOthers {.}}{%
Dunn%
\ \protect \BOthers {.}}{%
{\protect \APACyear {2021}}%
}]{%
DANNCE}
\APACinsertmetastar {%
DANNCE}%
\begin{APACrefauthors}%
Dunn, T.W.%
, Marshall, J.D.%
, Severson, K.S.%
, Aldarondo, D.E.%
, Hildebrand, D.G.%
, Chettih, S.N.%
\BDBL {}others%
\end{APACrefauthors}%
\unskip\
\newblock
\APACrefYearMonthDay{2021}{}{}.
\newblock
{\BBOQ}\APACrefatitle {Geometric deep learning enables 3D kinematic profiling
  across species and environments} {Geometric deep learning enables 3d
  kinematic profiling across species and environments}.{\BBCQ}
\newblock
\APACjournalVolNumPages{Nat. Methods}{18}{5}{564--573,}
\newblock

\newblock

\PrintBackRefs{\CurrentBib}

\bibitem [\protect \citeauthoryear {%
Duporge%
, Isupova%
, Reece%
, Macdonald%
\BCBL {}\ \BBA {} Wang%
}{%
Duporge%
\ \protect \BOthers {.}}{%
{\protect \APACyear {2021}}%
}]{%
Duporge2021us}
\APACinsertmetastar {%
Duporge2021us}%
\begin{APACrefauthors}%
Duporge, I.%
, Isupova, O.%
, Reece, S.%
, Macdonald, D.W.%
\BCBL {} Wang, T.%
\end{APACrefauthors}%
\unskip\
\newblock
\APACrefYearMonthDay{2021}{}{}.
\newblock
{\BBOQ}\APACrefatitle {Using very-high-resolution satellite imagery and deep
  learning to detect and count African elephants in heterogeneous landscapes}
  {Using very-high-resolution satellite imagery and deep learning to detect and
  count african elephants in heterogeneous landscapes}.{\BBCQ}
\newblock
\APACjournalVolNumPages{Remote Sensing in Ecology and
  Conservation}{7}{3}{369-381,}
\newblock

\newblock

\PrintBackRefs{\CurrentBib}

\bibitem [\protect \citeauthoryear {%
Ebrahimi%
\ \protect \BOthers {.}}{%
Ebrahimi%
\ \protect \BOthers {.}}{%
{\protect \APACyear {2023}}%
}]{%
3D-UPPER}
\APACinsertmetastar {%
3D-UPPER}%
\begin{APACrefauthors}%
Ebrahimi, A.S.%
, Orlowska-Feuer, P.%
, Huang, Q.%
, Zippo, A.G.%
, Martial, F.P.%
, Petersen, R.S.%
\BCBL {} Storchi, R.%
\end{APACrefauthors}%
\unskip\
\newblock
\APACrefYearMonthDay{2023}{}{}.
\newblock
{\BBOQ}\APACrefatitle {Three-dimensional unsupervised probabilistic pose
  reconstruction (3D-UPPER) for freely moving animals} {Three-dimensional
  unsupervised probabilistic pose reconstruction (3d-upper) for freely moving
  animals}.{\BBCQ}
\newblock
\APACjournalVolNumPages{Scientific Reports}{13}{1}{155,}
\newblock

\newblock

\PrintBackRefs{\CurrentBib}

\bibitem [\protect \citeauthoryear {%
Ferreira%
\ \protect \BOthers {.}}{%
Ferreira%
\ \protect \BOthers {.}}{%
{\protect \APACyear {2020}}%
}]{%
ferreira2020deep}
\APACinsertmetastar {%
ferreira2020deep}%
\begin{APACrefauthors}%
Ferreira, A.C.%
, Silva, L.R.%
, Renna, F.%
, Brandl, H.B.%
, Renoult, J.P.%
, Farine, D.R.%
\BDBL {}Doutrelant, C.%
\end{APACrefauthors}%
\unskip\
\newblock
\APACrefYearMonthDay{2020}{}{}.
\newblock
{\BBOQ}\APACrefatitle {Deep learning-based methods for individual recognition
  in small birds} {Deep learning-based methods for individual recognition in
  small birds}.{\BBCQ}
\newblock
\APACjournalVolNumPages{Methods in Ecology and Evolution}{11}{9}{1072--1085,}
\newblock

\newblock

\PrintBackRefs{\CurrentBib}

\bibitem [\protect \citeauthoryear {%
Ferrero%
\ \protect \BOthers {.}}{%
Ferrero%
\ \protect \BOthers {.}}{%
{\protect \APACyear {2017}}%
}]{%
idtrackerai}
\APACinsertmetastar {%
idtrackerai}%
\begin{APACrefauthors}%
Ferrero, F.R.%
, Bergomi, M.G.%
, Heras, F.J.%
, Hinz, R.%
, de Polavieja, G.G.%
\BCBL {} the Champalimaud~Foundation.%
\end{APACrefauthors}%
\unskip\
\newblock
\APACrefYearMonthDay{2017}{}{}.
\newblock
\APACrefbtitle {idtracker.ai.} {idtracker.ai.}
\newblock
\APACrefnote{\url{https://idtrackerai.readthedocs.io/en/latest}}
\PrintBackRefs{\CurrentBib}

\bibitem [\protect \citeauthoryear {%
Giebenhain%
, Waldmann%
, Johannsen%
\BCBL {}\ \BBA {} Goldluecke%
}{%
Giebenhain%
\ \protect \BOthers {.}}{%
{\protect \APACyear {2022}}%
}]{%
NePu}
\APACinsertmetastar {%
NePu}%
\begin{APACrefauthors}%
Giebenhain, S.%
, Waldmann, U.%
, Johannsen, O.%
\BCBL {} Goldluecke, B.%
\end{APACrefauthors}%
\unskip\
\newblock
\APACrefYearMonthDay{2022}{December}{}.
\newblock
{\BBOQ}\APACrefatitle {Neural Puppeteer: Keypoint-Based Neural Rendering of
  Dynamic Shapes} {Neural puppeteer: Keypoint-based neural rendering of dynamic
  shapes}.{\BBCQ}
\newblock
 \APACrefbtitle {Proceedings of the Asian Conference on Computer Vision (ACCV)}
  {Proceedings of the asian conference on computer vision (accv)}\ (\BPGS\
  2830--2847).
\PrintBackRefs{\CurrentBib}

\bibitem [\protect \citeauthoryear {%
Gomez-Marin%
, Kampff%
, Costa%
\BCBL {}\ \BBA {} Mainen%
}{%
Gomez-Marin%
\ \protect \BOthers {.}}{%
{\protect \APACyear {2014}}%
}]{%
gomez2014bi}
\APACinsertmetastar {%
gomez2014bi}%
\begin{APACrefauthors}%
Gomez-Marin, J.J., Alex an~Paton%
, Kampff, A.R.%
, Costa, R.M.%
\BCBL {} Mainen, Z.F.%
\end{APACrefauthors}%
\unskip\
\newblock
\APACrefYearMonthDay{2014}{}{}.
\newblock
{\BBOQ}\APACrefatitle {Big behavioral data: psychology, ethology and the
  foundations of neuroscience} {Big behavioral data: psychology, ethology and
  the foundations of neuroscience}.{\BBCQ}
\newblock
\APACjournalVolNumPages{Nat. Neurosci.}{17}{}{1455--1462,}
\newblock

\newblock

\PrintBackRefs{\CurrentBib}

\bibitem [\protect \citeauthoryear {%
Gosztolai%
\ \protect \BOthers {.}}{%
Gosztolai%
\ \protect \BOthers {.}}{%
{\protect \APACyear {2021}}%
}]{%
gosztolai2021li}
\APACinsertmetastar {%
gosztolai2021li}%
\begin{APACrefauthors}%
Gosztolai, A.%
, Günel, S.%
, Lobato-Ríos, V.%
, Pietro~Abrate, M.%
, Morales, D.%
, Rhodin, H.%
\BDBL {}Ramdya, P.%
\end{APACrefauthors}%
\unskip\
\newblock
\APACrefYearMonthDay{2021}{}{}.
\newblock
{\BBOQ}\APACrefatitle {LiftPose3D, a deep learning-based approach for
  transforming two-dimensional to three-dimensional poses in laboratory
  animals} {Liftpose3d, a deep learning-based approach for transforming
  two-dimensional to three-dimensional poses in laboratory animals}.{\BBCQ}
\newblock
\APACjournalVolNumPages{Nat. Methods}{18}{}{975--981,}
\newblock

\newblock

\PrintBackRefs{\CurrentBib}

\bibitem [\protect \citeauthoryear {%
Graving%
\ \protect \BOthers {.}}{%
Graving%
\ \protect \BOthers {.}}{%
{\protect \APACyear {2019}}%
}]{%
DPK}
\APACinsertmetastar {%
DPK}%
\begin{APACrefauthors}%
Graving, J.M.%
, Chae, D.%
, Naik, H.%
, Li, L.%
, Koger, B.%
, Costelloe, B.R.%
\BCBL {} Couzin, I.D.%
\end{APACrefauthors}%
\unskip\
\newblock
\APACrefYearMonthDay{2019}{oct}{}.
\newblock
{\BBOQ}\APACrefatitle {DeepPoseKit, a software toolkit for fast and robust
  animal pose estimation using deep learning} {Deepposekit, a software toolkit
  for fast and robust animal pose estimation using deep learning}.{\BBCQ}
\newblock
\APACjournalVolNumPages{eLife}{8}{}{e47994,}
\newblock
\begin{APACrefDOI} \doi{10.7554/eLife.47994} \end{APACrefDOI}
\newblock

\newblock

\PrintBackRefs{\CurrentBib}

\bibitem [\protect \citeauthoryear {%
Günel%
\ \protect \BOthers {.}}{%
Günel%
\ \protect \BOthers {.}}{%
{\protect \APACyear {2019}}%
}]{%
DeepFly3D}
\APACinsertmetastar {%
DeepFly3D}%
\begin{APACrefauthors}%
Günel, S.%
, Rhodin, H.%
, Morales, D.%
, Campagnolo, J.%
, Ramdya, P.%
\BCBL {} Fua, P.%
\end{APACrefauthors}%
\unskip\
\newblock
\APACrefYearMonthDay{2019}{}{}.
\newblock
{\BBOQ}\APACrefatitle {DeepFly3D, a deep learning-based approach for 3D limb
  and appendage tracking in tethered, adult \textit{Drosophila}} {Deepfly3d, a
  deep learning-based approach for 3d limb and appendage tracking in tethered,
  adult \textit{Drosophila}}.{\BBCQ}
\newblock
\APACjournalVolNumPages{eLife}{8}{}{e48571,}
\newblock

\newblock

\PrintBackRefs{\CurrentBib}

\bibitem [\protect \citeauthoryear {%
Han%
\ \protect \BOthers {.}}{%
Han%
\ \protect \BOthers {.}}{%
{\protect \APACyear {2023}}%
}]{%
han2023social}
\APACinsertmetastar {%
han2023social}%
\begin{APACrefauthors}%
Han, Y.%
, Chen, K.%
, Wang, Y.%
, Liu, W.%
, Wang, X.%
, Liao, J.%
\BDBL {}others%
\end{APACrefauthors}%
\unskip\
\newblock
\APACrefYearMonthDay{2023}{}{}.
\newblock
{\BBOQ}\APACrefatitle {Social Behavior Atlas: A computational framework for
  tracking and mapping 3D close interactions of free-moving animals} {Social
  behavior atlas: A computational framework for tracking and mapping 3d close
  interactions of free-moving animals}.{\BBCQ}
\newblock
\APACjournalVolNumPages{bioRxiv}{}{}{2023--03,}
\newblock

\newblock

\PrintBackRefs{\CurrentBib}

\bibitem [\protect \citeauthoryear {%
He%
, Gkioxari%
, Dollar%
\BCBL {}\ \BBA {} Girshick%
}{%
He%
\ \protect \BOthers {.}}{%
{\protect \APACyear {2017}}%
}]{%
MaskRCNN}
\APACinsertmetastar {%
MaskRCNN}%
\begin{APACrefauthors}%
He, K.%
, Gkioxari, G.%
, Dollar, P.%
\BCBL {} Girshick, R.%
\end{APACrefauthors}%
\unskip\
\newblock
\APACrefYearMonthDay{2017}{}{}.
\newblock
{\BBOQ}\APACrefatitle {Mask R-CNN} {Mask r-cnn}.{\BBCQ}
\newblock
 \APACrefbtitle {Int. Conf. Comput. Vis.} {Int. conf. comput. vis.}
\PrintBackRefs{\CurrentBib}

\bibitem [\protect \citeauthoryear {%
He%
, Zhang%
, Ren%
\BCBL {}\ \BBA {} Sun%
}{%
He%
\ \protect \BOthers {.}}{%
{\protect \APACyear {2016}}%
}]{%
ResNet}
\APACinsertmetastar {%
ResNet}%
\begin{APACrefauthors}%
He, K.%
, Zhang, X.%
, Ren, S.%
\BCBL {} Sun, J.%
\end{APACrefauthors}%
\unskip\
\newblock
\APACrefYearMonthDay{2016}{}{}.
\newblock
{\BBOQ}\APACrefatitle {Deep Residual Learning for Image Recognition} {Deep
  residual learning for image recognition}.{\BBCQ}
\newblock
 \APACrefbtitle {IEEE Conf. Comput. Vis. Pattern Recog.} {Ieee conf. comput.
  vis. pattern recog.}
\PrintBackRefs{\CurrentBib}

\bibitem [\protect \citeauthoryear {%
Heras%
, Romero-Ferrero%
, Hinz%
\BCBL {}\ \BBA {} de Polavieja%
}{%
Heras%
\ \protect \BOthers {.}}{%
{\protect \APACyear {2019}}%
}]{%
Zebrafish}
\APACinsertmetastar {%
Zebrafish}%
\begin{APACrefauthors}%
Heras, F.J.H.%
, Romero-Ferrero, F.%
, Hinz, R.C.%
\BCBL {} de Polavieja, G.G.%
\end{APACrefauthors}%
\unskip\
\newblock
\APACrefYearMonthDay{2019}{}{}.
\newblock
{\BBOQ}\APACrefatitle {Deep attention networks reveal the rules of collective
  motion in zebrafish} {Deep attention networks reveal the rules of collective
  motion in zebrafish}.{\BBCQ}
\newblock
\APACjournalVolNumPages{PLOS Computational Biology}{15}{9}{1-23,}
\newblock

\newblock

\PrintBackRefs{\CurrentBib}

\bibitem [\protect \citeauthoryear {%
Huang%
\ \protect \BOthers {.}}{%
Huang%
\ \protect \BOthers {.}}{%
{\protect \APACyear {2020}}%
}]{%
huang2020end}
\APACinsertmetastar {%
huang2020end}%
\begin{APACrefauthors}%
Huang, C.%
, Jiang, S.%
, Li, Y.%
, Zhang, Z.%
, Traish, J.%
, Deng, C.%
\BDBL {}Da~Xu, R.Y.%
\end{APACrefauthors}%
\unskip\
\newblock
\APACrefYearMonthDay{2020}{}{}.
\newblock
{\BBOQ}\APACrefatitle {End-to-end dynamic matching network for multi-view
  multi-person 3d pose estimation} {End-to-end dynamic matching network for
  multi-view multi-person 3d pose estimation}.{\BBCQ}
\newblock
 \APACrefbtitle {Computer Vision--ECCV 2020: 16th European Conference, Glasgow,
  UK, August 23--28, 2020, Proceedings, Part XXVIII 16} {Computer vision--eccv
  2020: 16th european conference, glasgow, uk, august 23--28, 2020,
  proceedings, part xxviii 16}\ (\BPGS\ 477--493).
\PrintBackRefs{\CurrentBib}

\bibitem [\protect \citeauthoryear {%
Ionescu%
, Papava%
, Olaru%
\BCBL {}\ \BBA {} Sminchisescu%
}{%
Ionescu%
\ \protect \BOthers {.}}{%
{\protect \APACyear {2014}}%
}]{%
H36M}
\APACinsertmetastar {%
H36M}%
\begin{APACrefauthors}%
Ionescu, C.%
, Papava, D.%
, Olaru, V.%
\BCBL {} Sminchisescu, C.%
\end{APACrefauthors}%
\unskip\
\newblock
\APACrefYearMonthDay{2014}{}{}.
\newblock
{\BBOQ}\APACrefatitle {Human3.6M: Large Scale Datasets and Predictive Methods
  for 3D Human Sensing in Natural Environments} {Human3.6m: Large scale
  datasets and predictive methods for 3d human sensing in natural
  environments}.{\BBCQ}
\newblock
\APACjournalVolNumPages{IEEE Trans. Pattern Anal. Mach.
  Intell.}{36}{7}{1325-1339,}
\newblock

\newblock

\PrintBackRefs{\CurrentBib}

\bibitem [\protect \citeauthoryear {%
Iskakov%
, Burkov%
, Lempitsky%
\BCBL {}\ \BBA {} Malkov%
}{%
Iskakov%
\ \protect \BOthers {.}}{%
{\protect \APACyear {2019}}%
}]{%
LToHP}
\APACinsertmetastar {%
LToHP}%
\begin{APACrefauthors}%
Iskakov, K.%
, Burkov, E.%
, Lempitsky, V.%
\BCBL {} Malkov, Y.%
\end{APACrefauthors}%
\unskip\
\newblock
\APACrefYearMonthDay{2019}{}{}.
\newblock
{\BBOQ}\APACrefatitle {Learnable Triangulation of Human Pose} {Learnable
  triangulation of human pose}.{\BBCQ}
\newblock
 \APACrefbtitle {Int. Conf. Comput. Vis.} {Int. conf. comput. vis.}
\PrintBackRefs{\CurrentBib}

\bibitem [\protect \citeauthoryear {%
Itahara%
\ \BBA {} Kano%
}{%
Itahara%
\ \BBA {} Kano%
}{%
{\protect \APACyear {2022}}%
}]{%
itahara2022corvid}
\APACinsertmetastar {%
itahara2022corvid}%
\begin{APACrefauthors}%
Itahara, A.%
\BCBT {}\ \BBA {} Kano, F.%
\end{APACrefauthors}%
\unskip\
\newblock
\APACrefYearMonthDay{2022}{}{}.
\newblock
{\BBOQ}\APACrefatitle {``Corvid Tracking Studio'': A custom-built motion
  capture system to track head movements of corvids.} {``corvid tracking
  studio'': A custom-built motion capture system to track head movements of
  corvids.}{\BBCQ}
\newblock
\APACjournalVolNumPages{Japanese Journal of Animal Psychology}{72}{1}{1--16,}
\newblock

\newblock

\PrintBackRefs{\CurrentBib}

\bibitem [\protect \citeauthoryear {%
Itahara%
\ \BBA {} Kano%
}{%
Itahara%
\ \BBA {} Kano%
}{%
{\protect \APACyear {2023}}%
}]{%
Itahara2023}
\APACinsertmetastar {%
Itahara2023}%
\begin{APACrefauthors}%
Itahara, A.%
\BCBT {}\ \BBA {} Kano, F.%
\end{APACrefauthors}%
\unskip\
\newblock
\APACrefYearMonthDay{2023}{}{}.
\newblock
{\BBOQ}\APACrefatitle {Gaze tracking of large-billed crows (Corvus
  macrorhynchos) in a motion-capture system} {Gaze tracking of large-billed
  crows (corvus macrorhynchos) in a motion-capture system}.{\BBCQ}
\newblock
\APACjournalVolNumPages{bioRxiv}{}{}{,}
\newblock
\begin{APACrefDOI} \doi{10.1101/2023.08.10.552747} \end{APACrefDOI}
\newblock

\newblock

\PrintBackRefs{\CurrentBib}

\bibitem [\protect \citeauthoryear {%
Jocher%
, Chaurasia%
\BCBL {}\ \BBA {} Qiu%
}{%
Jocher%
\ \protect \BOthers {.}}{%
{\protect \APACyear {2023}}%
}]{%
YOLOv8}
\APACinsertmetastar {%
YOLOv8}%
\begin{APACrefauthors}%
Jocher, G.%
, Chaurasia, A.%
\BCBL {} Qiu, J.%
\end{APACrefauthors}%
\unskip\
\newblock
\APACrefYearMonthDay{2023}{jan}{}.
\newblock
\APACrefbtitle {YOLO by Ultralytics.} {Yolo by ultralytics.}
\newblock
\begin{APACrefURL} {https://github.com/ultralytics/ultralytics}
  \end{APACrefURL}
\PrintBackRefs{\CurrentBib}

\bibitem [\protect \citeauthoryear {%
Joska%
\ \protect \BOthers {.}}{%
Joska%
\ \protect \BOthers {.}}{%
{\protect \APACyear {2021}}%
}]{%
AcinoSet}
\APACinsertmetastar {%
AcinoSet}%
\begin{APACrefauthors}%
Joska, D.%
, Clark, L.%
, Muramatsu, N.%
, Jericevich, R.%
, Nicolls, F.%
, Mathis, A.%
\BDBL {}Patel, A.%
\end{APACrefauthors}%
\unskip\
\newblock
\APACrefYearMonthDay{2021}{}{}.
\newblock
{\BBOQ}\APACrefatitle {AcinoSet: A 3D Pose Estimation Dataset and Baseline
  Models for Cheetahs in the Wild} {Acinoset: A 3d pose estimation dataset and
  baseline models for cheetahs in the wild}.{\BBCQ}
\newblock
 \APACrefbtitle {2021 IEEE International Conference on Robotics and Automation
  (ICRA)} {2021 ieee international conference on robotics and automation
  (icra)}\ (\BPG~13901-13908).
\PrintBackRefs{\CurrentBib}

\bibitem [\protect \citeauthoryear {%
Kalman%
}{%
Kalman%
}{%
{\protect \APACyear {1960}}%
}]{%
KalmanFilter}
\APACinsertmetastar {%
KalmanFilter}%
\begin{APACrefauthors}%
Kalman, R.E.%
\end{APACrefauthors}%
\unskip\
\newblock
\APACrefYearMonthDay{1960}{}{}.
\newblock
{\BBOQ}\APACrefatitle {{A New Approach to Linear Filtering and Prediction
  Problems}} {{A New Approach to Linear Filtering and Prediction
  Problems}}.{\BBCQ}
\newblock
\APACjournalVolNumPages{Journal of Basic Engineering}{82}{1}{35-45,}
\newblock

\newblock

\PrintBackRefs{\CurrentBib}

\bibitem [\protect \citeauthoryear {%
Kane%
, Lopes%
, Saunders%
, Mathis%
\BCBL {}\ \BBA {} Mathis%
}{%
Kane%
\ \protect \BOthers {.}}{%
{\protect \APACyear {2020}}%
}]{%
DLC-live}
\APACinsertmetastar {%
DLC-live}%
\begin{APACrefauthors}%
Kane, G.A.%
, Lopes, G.%
, Saunders, J.L.%
, Mathis, A.%
\BCBL {} Mathis, M.W.%
\end{APACrefauthors}%
\unskip\
\newblock
\APACrefYearMonthDay{2020}{}{}.
\newblock
{\BBOQ}\APACrefatitle {Real-time, low-latency closed-loop feedback using
  markerless posture tracking} {Real-time, low-latency closed-loop feedback
  using markerless posture tracking}.{\BBCQ}
\newblock
\APACjournalVolNumPages{Elife}{9}{}{e61909,}
\newblock

\newblock

\PrintBackRefs{\CurrentBib}

\bibitem [\protect \citeauthoryear {%
Kano%
, Naik%
, Keskin%
, Couzin%
\BCBL {}\ \BBA {} Nagy%
}{%
Kano%
\ \protect \BOthers {.}}{%
{\protect \APACyear {2022}}%
}]{%
kano2022head}
\APACinsertmetastar {%
kano2022head}%
\begin{APACrefauthors}%
Kano, F.%
, Naik, H.%
, Keskin, G.%
, Couzin, I.D.%
\BCBL {} Nagy, M.%
\end{APACrefauthors}%
\unskip\
\newblock
\APACrefYearMonthDay{2022}{}{}.
\newblock
{\BBOQ}\APACrefatitle {Head-tracking of freely-behaving pigeons in a
  motion-capture system reveals the selective use of visual field regions}
  {Head-tracking of freely-behaving pigeons in a motion-capture system reveals
  the selective use of visual field regions}.{\BBCQ}
\newblock
\APACjournalVolNumPages{Scientific Reports}{12}{1}{19113,}
\newblock

\newblock

\PrintBackRefs{\CurrentBib}

\bibitem [\protect \citeauthoryear {%
Karaev%
\ \protect \BOthers {.}}{%
Karaev%
\ \protect \BOthers {.}}{%
{\protect \APACyear {2023}}%
}]{%
karaev2023cotracker}
\APACinsertmetastar {%
karaev2023cotracker}%
\begin{APACrefauthors}%
Karaev, N.%
, Rocco, I.%
, Graham, B.%
, Neverova, N.%
, Vedaldi, A.%
\BCBL {} Rupprecht, C.%
\end{APACrefauthors}%
\unskip\
\newblock
\APACrefYearMonthDay{2023}{}{}.
\newblock
{\BBOQ}\APACrefatitle {Cotracker: It is better to track together} {Cotracker:
  It is better to track together}.{\BBCQ}
\newblock
\APACjournalVolNumPages{arXiv preprint arXiv:2307.07635}{}{}{,}
\newblock

\newblock

\PrintBackRefs{\CurrentBib}

\bibitem [\protect \citeauthoryear {%
Karashchuk%
\ \protect \BOthers {.}}{%
Karashchuk%
\ \protect \BOthers {.}}{%
{\protect \APACyear {2021}}%
}]{%
Anipose}
\APACinsertmetastar {%
Anipose}%
\begin{APACrefauthors}%
Karashchuk, P.%
, Rupp, K.L.%
, Dickinson, E.S.%
, Walling-Bell, S.%
, Sanders, E.%
, Azim, E.%
\BDBL {}Tuthill, J.C.%
\end{APACrefauthors}%
\unskip\
\newblock
\APACrefYearMonthDay{2021}{}{}.
\newblock
{\BBOQ}\APACrefatitle {Anipose: A toolkit for robust markerless 3D pose
  estimation} {Anipose: A toolkit for robust markerless 3d pose
  estimation}.{\BBCQ}
\newblock
\APACjournalVolNumPages{Cell Reports}{36}{13}{109730,}
\newblock

\newblock

\PrintBackRefs{\CurrentBib}

\bibitem [\protect \citeauthoryear {%
Kays%
, Crofoot%
, Jetz%
\BCBL {}\ \BBA {} Wikelski%
}{%
Kays%
\ \protect \BOthers {.}}{%
{\protect \APACyear {2015}}%
}]{%
Kays2015te}
\APACinsertmetastar {%
Kays2015te}%
\begin{APACrefauthors}%
Kays, R.%
, Crofoot, M.C.%
, Jetz, W.%
\BCBL {} Wikelski, M.%
\end{APACrefauthors}%
\unskip\
\newblock
\APACrefYearMonthDay{2015}{}{}.
\newblock
{\BBOQ}\APACrefatitle {Terrestrial animal tracking as an eye on life and
  planet} {Terrestrial animal tracking as an eye on life and planet}.{\BBCQ}
\newblock
\APACjournalVolNumPages{Science}{348}{6240}{aaa2478,}
\newblock

\newblock

\PrintBackRefs{\CurrentBib}

\bibitem [\protect \citeauthoryear {%
Kirillov%
\ \protect \BOthers {.}}{%
Kirillov%
\ \protect \BOthers {.}}{%
{\protect \APACyear {2023}}%
}]{%
SAM}
\APACinsertmetastar {%
SAM}%
\begin{APACrefauthors}%
Kirillov, A.%
, Mintun, E.%
, Ravi, N.%
, Mao, H.%
, Rolland, C.%
, Gustafson, L.%
\BDBL {}others%
\end{APACrefauthors}%
\unskip\
\newblock
\APACrefYearMonthDay{2023}{}{}.
\newblock
{\BBOQ}\APACrefatitle {Segment anything} {Segment anything}.{\BBCQ}
\newblock
\APACjournalVolNumPages{arXiv preprint arXiv:2304.02643}{}{}{,}
\newblock

\newblock

\PrintBackRefs{\CurrentBib}

\bibitem [\protect \citeauthoryear {%
Koger%
\ \protect \BOthers {.}}{%
Koger%
\ \protect \BOthers {.}}{%
{\protect \APACyear {2023}}%
}]{%
koger2023quantifying}
\APACinsertmetastar {%
koger2023quantifying}%
\begin{APACrefauthors}%
Koger, B.%
, Deshpande, A.%
, Kerby, J.T.%
, Graving, J.M.%
, Costelloe, B.R.%
\BCBL {} Couzin, I.D.%
\end{APACrefauthors}%
\unskip\
\newblock
\APACrefYearMonthDay{2023}{}{}.
\newblock
{\BBOQ}\APACrefatitle {Quantifying the movement, behaviour and environmental
  context of group-living animals using drones and computer vision}
  {Quantifying the movement, behaviour and environmental context of
  group-living animals using drones and computer vision}.{\BBCQ}
\newblock
\APACjournalVolNumPages{Journal of Animal Ecology}{}{}{,}
\newblock

\newblock

\PrintBackRefs{\CurrentBib}

\bibitem [\protect \citeauthoryear {%
Labuguen%
\ \protect \BOthers {.}}{%
Labuguen%
\ \protect \BOthers {.}}{%
{\protect \APACyear {2021}}%
}]{%
labuguen2021ma}
\APACinsertmetastar {%
labuguen2021ma}%
\begin{APACrefauthors}%
Labuguen, R.%
, Matsumoto, J.%
, Negrete, S.B.%
, Nishimaru, H.%
, Nishijo, H.%
, Takada, M.%
\BDBL {}Shibata, T.%
\end{APACrefauthors}%
\unskip\
\newblock
\APACrefYearMonthDay{2021}{}{}.
\newblock
{\BBOQ}\APACrefatitle {MacaquePose: A Novel “In the Wild” Macaque Monkey
  Pose Dataset for Markerless Motion Capture} {Macaquepose: A novel “in the
  wild” macaque monkey pose dataset for markerless motion capture}.{\BBCQ}
\newblock
\APACjournalVolNumPages{Frontiers in Behavioral Neuroscience}{14}{}{268,}
\newblock

\newblock

\PrintBackRefs{\CurrentBib}

\bibitem [\protect \citeauthoryear {%
Lauer%
\ \protect \BOthers {.}}{%
Lauer%
\ \protect \BOthers {.}}{%
{\protect \APACyear {2022}}%
}]{%
maDLC}
\APACinsertmetastar {%
maDLC}%
\begin{APACrefauthors}%
Lauer, J.%
, Zhou, M.%
, Ye, S.%
, Menegas, W.%
, Schneider, S.%
, Nath, T.%
\BDBL {}Mathis, A.%
\end{APACrefauthors}%
\unskip\
\newblock
\APACrefYearMonthDay{2022}{}{}.
\newblock
{\BBOQ}\APACrefatitle {Multi-animal pose estimation, identification and
  tracking with DeepLabCut} {Multi-animal pose estimation, identification and
  tracking with deeplabcut}.{\BBCQ}
\newblock
\APACjournalVolNumPages{Nat. Methods}{19}{}{496–504,}
\newblock

\newblock

\PrintBackRefs{\CurrentBib}

\bibitem [\protect \citeauthoryear {%
Li%
, Huang%
\BCBL {}\ \BBA {} Nevatia%
}{%
Li%
\ \protect \BOthers {.}}{%
{\protect \APACyear {2009}}%
}]{%
nevatia2009learning}
\APACinsertmetastar {%
nevatia2009learning}%
\begin{APACrefauthors}%
Li, Y.%
, Huang, C.%
\BCBL {} Nevatia, R.%
\end{APACrefauthors}%
\unskip\
\newblock
\APACrefYearMonthDay{2009}{}{}.
\newblock
{\BBOQ}\APACrefatitle {Learning to associate: HybridBoosted multi-target
  tracker for crowded scene} {Learning to associate: Hybridboosted multi-target
  tracker for crowded scene}.{\BBCQ}
\newblock
 \APACrefbtitle {IEEE Conf. Comput. Vis. Pattern Recog.} {Ieee conf. comput.
  vis. pattern recog.}\ (\BPG~2953-2960).
\PrintBackRefs{\CurrentBib}

\bibitem [\protect \citeauthoryear {%
Lin%
\ \protect \BOthers {.}}{%
Lin%
\ \protect \BOthers {.}}{%
{\protect \APACyear {2017}}%
}]{%
FPN}
\APACinsertmetastar {%
FPN}%
\begin{APACrefauthors}%
Lin, T\BHBI Y.%
, Dollar, P.%
, Girshick, R.%
, He, K.%
, Hariharan, B.%
\BCBL {} Belongie, S.%
\end{APACrefauthors}%
\unskip\
\newblock
\APACrefYearMonthDay{2017}{}{}.
\newblock
{\BBOQ}\APACrefatitle {Feature Pyramid Networks for Object Detection} {Feature
  pyramid networks for object detection}.{\BBCQ}
\newblock
 \APACrefbtitle {IEEE Conf. Comput. Vis. Pattern Recog.} {Ieee conf. comput.
  vis. pattern recog.}
\PrintBackRefs{\CurrentBib}

\bibitem [\protect \citeauthoryear {%
Luiten%
\ \BBA {} Hoffhues%
}{%
Luiten%
\ \BBA {} Hoffhues%
}{%
{\protect \APACyear {2020}}%
}]{%
luiten2020trackeval}
\APACinsertmetastar {%
luiten2020trackeval}%
\begin{APACrefauthors}%
Luiten, J.%
\BCBT {}\ \BBA {} Hoffhues, A.%
\end{APACrefauthors}%
\unskip\
\newblock
\APACrefYearMonthDay{2020}{}{}.
\newblock
\APACrefbtitle {TrackEval.} {Trackeval.}
\newblock
\APAChowpublished {\url{https://github.com/JonathonLuiten/TrackEval}}.
\PrintBackRefs{\CurrentBib}

\bibitem [\protect \citeauthoryear {%
Luiten%
\ \protect \BOthers {.}}{%
Luiten%
\ \protect \BOthers {.}}{%
{\protect \APACyear {2021}}%
}]{%
HOTA}
\APACinsertmetastar {%
HOTA}%
\begin{APACrefauthors}%
Luiten, J.%
, Osep, A.%
, Dendorfer, P.%
, Torr, P.%
, Geiger, A.%
, Leal-Taix{\'e}, L.%
\BCBL {} Leibe, B.%
\end{APACrefauthors}%
\unskip\
\newblock
\APACrefYearMonthDay{2021}{}{}.
\newblock
{\BBOQ}\APACrefatitle {Hota: A higher order metric for evaluating multi-object
  tracking} {Hota: A higher order metric for evaluating multi-object
  tracking}.{\BBCQ}
\newblock
\APACjournalVolNumPages{Int. J. Comput. Vis.}{129}{2}{548--578,}
\newblock

\newblock

\PrintBackRefs{\CurrentBib}

\bibitem [\protect \citeauthoryear {%
Marshall%
\ \protect \BOthers {.}}{%
Marshall%
\ \protect \BOthers {.}}{%
{\protect \APACyear {2021}}%
}]{%
marshall2021pair}
\APACinsertmetastar {%
marshall2021pair}%
\begin{APACrefauthors}%
Marshall, J.D.%
, Klibaite, U.%
, Gellis, A.%
, Aldarondo, D.E.%
, {\"O}lveczky, B.P.%
\BCBL {} Dunn, T.W.%
\end{APACrefauthors}%
\unskip\
\newblock
\APACrefYearMonthDay{2021}{}{}.
\newblock
{\BBOQ}\APACrefatitle {The pair-r24m dataset for multi-animal 3d pose
  estimation} {The pair-r24m dataset for multi-animal 3d pose
  estimation}.{\BBCQ}
\newblock
\APACjournalVolNumPages{bioRxiv}{}{}{2021--11,}
\newblock

\newblock

\PrintBackRefs{\CurrentBib}

\bibitem [\protect \citeauthoryear {%
Mathis%
\ \protect \BOthers {.}}{%
Mathis%
\ \protect \BOthers {.}}{%
{\protect \APACyear {2018}}%
}]{%
DLC}
\APACinsertmetastar {%
DLC}%
\begin{APACrefauthors}%
Mathis, A.%
, Mamidanna, P.%
, Cury, K.M.%
, Abe, T.%
, Murthy, V.N.%
, Mathis, M.W.%
\BCBL {} Bethge, M.%
\end{APACrefauthors}%
\unskip\
\newblock
\APACrefYearMonthDay{2018}{}{}.
\newblock
{\BBOQ}\APACrefatitle {DeepLabCut: markerless pose estimation of user-defined
  body parts with deep learning} {Deeplabcut: markerless pose estimation of
  user-defined body parts with deep learning}.{\BBCQ}
\newblock
\APACjournalVolNumPages{Nat. Neurosci.}{21}{}{1281--1289,}
\newblock

\newblock

\PrintBackRefs{\CurrentBib}

\bibitem [\protect \citeauthoryear {%
Mi{\~n}ano%
, Golodetz%
, Cavallari%
\BCBL {}\ \BBA {} Taylor%
}{%
Mi{\~n}ano%
\ \protect \BOthers {.}}{%
{\protect \APACyear {2023}}%
}]{%
minano2023through}
\APACinsertmetastar {%
minano2023through}%
\begin{APACrefauthors}%
Mi{\~n}ano, S.%
, Golodetz, S.%
, Cavallari, T.%
\BCBL {} Taylor, G.K.%
\end{APACrefauthors}%
\unskip\
\newblock
\APACrefYearMonthDay{2023}{}{}.
\newblock
{\BBOQ}\APACrefatitle {Through hawks’ eyes: synthetically reconstructing the
  visual field of a bird in flight} {Through hawks’ eyes: synthetically
  reconstructing the visual field of a bird in flight}.{\BBCQ}
\newblock
\APACjournalVolNumPages{International Journal of Computer
  Vision}{131}{6}{1497--1531,}
\newblock

\newblock

\PrintBackRefs{\CurrentBib}

\bibitem [\protect \citeauthoryear {%
Nagy%
, {\'A}kos%
, Biro%
\BCBL {}\ \BBA {} Vicsek%
}{%
Nagy%
\ \protect \BOthers {.}}{%
{\protect \APACyear {2010}}%
}]{%
nagy2010hierarchical}
\APACinsertmetastar {%
nagy2010hierarchical}%
\begin{APACrefauthors}%
Nagy, M.%
, {\'A}kos, Z.%
, Biro, D.%
\BCBL {} Vicsek, T.%
\end{APACrefauthors}%
\unskip\
\newblock
\APACrefYearMonthDay{2010}{}{}.
\newblock
{\BBOQ}\APACrefatitle {Hierarchical group dynamics in pigeon flocks}
  {Hierarchical group dynamics in pigeon flocks}.{\BBCQ}
\newblock
\APACjournalVolNumPages{Nature}{464}{7290}{890--893,}
\newblock

\newblock

\PrintBackRefs{\CurrentBib}

\bibitem [\protect \citeauthoryear {%
Nagy%
\ \protect \BOthers {.}}{%
Nagy%
\ \protect \BOthers {.}}{%
{\protect \APACyear {2023}}%
}]{%
SMART-BARN}
\APACinsertmetastar {%
SMART-BARN}%
\begin{APACrefauthors}%
Nagy, M.%
, Naik, H.%
, Fumihiro, K.%
, Nora, C.V.%
, Koblitz, J.C.%
, Wikelski, M.%
\BCBL {} Couzin, I.D.%
\end{APACrefauthors}%
\unskip\
\newblock
\APACrefYearMonthDay{2023}{}{}.
\newblock
{\BBOQ}\APACrefatitle {SMART-BARN: Scalable Multimodal Arena for Real-time
  Tracking Behavior of Animals in Large Numbers} {Smart-barn: Scalable
  multimodal arena for real-time tracking behavior of animals in large
  numbers}.{\BBCQ}
\newblock
\APACjournalVolNumPages{(in press) Science Advances}{}{}{,}
\newblock

\newblock

\PrintBackRefs{\CurrentBib}

\bibitem [\protect \citeauthoryear {%
Nagy%
\ \protect \BOthers {.}}{%
Nagy%
\ \protect \BOthers {.}}{%
{\protect \APACyear {2013}}%
}]{%
nagy2013context}
\APACinsertmetastar {%
nagy2013context}%
\begin{APACrefauthors}%
Nagy, M.%
, V{\'a}s{\'a}rhelyi, G.%
, Pettit, B.%
, Roberts-Mariani, I.%
, Vicsek, T.%
\BCBL {} Biro, D.%
\end{APACrefauthors}%
\unskip\
\newblock
\APACrefYearMonthDay{2013}{}{}.
\newblock
{\BBOQ}\APACrefatitle {Context-dependent hierarchies in pigeons}
  {Context-dependent hierarchies in pigeons}.{\BBCQ}
\newblock
\APACjournalVolNumPages{Proceedings of the National Academy of
  Sciences}{110}{32}{13049--13054,}
\newblock

\newblock

\PrintBackRefs{\CurrentBib}

\bibitem [\protect \citeauthoryear {%
Naik%
}{%
Naik%
}{%
{\protect \APACyear {2021}}%
}]{%
naik2021xrforall}
\APACinsertmetastar {%
naik2021xrforall}%
\begin{APACrefauthors}%
Naik, H.%
\end{APACrefauthors}%
\unskip\
\newblock
\APACrefYear{2021}.
\unskip\
\newblock
\APACrefbtitle {XR For All: Closed-loop Visual Stimulation Techniques for Human
  and Non-Human Animals} {Xr for all: Closed-loop visual stimulation techniques
  for human and non-human animals}\ \APACtypeAddressSchool {Dissertation}{}{}.
\unskip\
\newblock
\APACaddressSchool {München}{Technische Universität München}.
\PrintBackRefs{\CurrentBib}

\bibitem [\protect \citeauthoryear {%
Naik%
, Bastien%
, Navab%
\BCBL {}\ \BBA {} Couzin%
}{%
Naik%
\ \protect \BOthers {.}}{%
{\protect \APACyear {2020}}%
}]{%
naik2020animals}
\APACinsertmetastar {%
naik2020animals}%
\begin{APACrefauthors}%
Naik, H.%
, Bastien, R.%
, Navab, N.%
\BCBL {} Couzin, I.D.%
\end{APACrefauthors}%
\unskip\
\newblock
\APACrefYearMonthDay{2020}{}{}.
\newblock
{\BBOQ}\APACrefatitle {Animals in virtual environments} {Animals in virtual
  environments}.{\BBCQ}
\newblock
\APACjournalVolNumPages{IEEE Transactions on Visualization and Computer
  Graphics}{26}{5}{2073--2083,}
\newblock

\newblock

\PrintBackRefs{\CurrentBib}

\bibitem [\protect \citeauthoryear {%
Naik%
\ \protect \BOthers {.}}{%
Naik%
\ \protect \BOthers {.}}{%
{\protect \APACyear {2023}}%
}]{%
3D-POP}
\APACinsertmetastar {%
3D-POP}%
\begin{APACrefauthors}%
Naik, H.%
, Chan, A.H.H.%
, Yang, J.%
, Delacoux, M.%
, Couzin, I.D.%
, Kano, F.%
\BCBL {} Nagy, M.%
\end{APACrefauthors}%
\unskip\
\newblock
\APACrefYearMonthDay{2023}{June}{}.
\newblock
{\BBOQ}\APACrefatitle {3D-POP - An Automated Annotation Approach to Facilitate
  Markerless 2D-3D Tracking of Freely Moving Birds With Marker-Based Motion
  Capture} {3d-pop - an automated annotation approach to facilitate markerless
  2d-3d tracking of freely moving birds with marker-based motion
  capture}.{\BBCQ}
\newblock
 \APACrefbtitle {IEEE Conf. Comput. Vis. Pattern Recog.} {Ieee conf. comput.
  vis. pattern recog.}\ (\BPG~21274-21284).
\PrintBackRefs{\CurrentBib}

\bibitem [\protect \citeauthoryear {%
Nath%
\ \protect \BOthers {.}}{%
Nath%
\ \protect \BOthers {.}}{%
{\protect \APACyear {2019}}%
}]{%
3DDLC}
\APACinsertmetastar {%
3DDLC}%
\begin{APACrefauthors}%
Nath, T.%
, Mathis, A.%
, Chen, A.C.%
, Patel, A.%
, Bethge, M.%
\BCBL {} Mathis, M.W.%
\end{APACrefauthors}%
\unskip\
\newblock
\APACrefYearMonthDay{2019}{}{}.
\newblock
{\BBOQ}\APACrefatitle {Using DeepLabCut for 3D markerless pose estimation
  across species and behaviors} {Using deeplabcut for 3d markerless pose
  estimation across species and behaviors}.{\BBCQ}
\newblock
\APACjournalVolNumPages{Nat. Protoc.}{14}{}{2152--2176,}
\newblock

\newblock

\PrintBackRefs{\CurrentBib}

\bibitem [\protect \citeauthoryear {%
Newell%
, Yang%
\BCBL {}\ \BBA {} Deng%
}{%
Newell%
\ \protect \BOthers {.}}{%
{\protect \APACyear {2016}}%
}]{%
StackedHourglass}
\APACinsertmetastar {%
StackedHourglass}%
\begin{APACrefauthors}%
Newell, A.%
, Yang, K.%
\BCBL {} Deng, J.%
\end{APACrefauthors}%
\unskip\
\newblock
\APACrefYearMonthDay{2016}{}{}.
\newblock
{\BBOQ}\APACrefatitle {Stacked Hourglass Networks for Human Pose Estimation}
  {Stacked hourglass networks for human pose estimation}.{\BBCQ}
\newblock
 \APACrefbtitle {Computer Vision--ECCV 2016: 14th European Conference,
  Amsterdam, The Netherlands, October 11-14, 2016, Proceedings, Part VIII 14}
  {Computer vision--eccv 2016: 14th european conference, amsterdam, the
  netherlands, october 11-14, 2016, proceedings, part viii 14}\ (\BPGS\
  483--499).
\PrintBackRefs{\CurrentBib}

\bibitem [\protect \citeauthoryear {%
Nourizonoz%
\ \protect \BOthers {.}}{%
Nourizonoz%
\ \protect \BOthers {.}}{%
{\protect \APACyear {2020}}%
}]{%
EthoLoop}
\APACinsertmetastar {%
EthoLoop}%
\begin{APACrefauthors}%
Nourizonoz, A.%
, Zimmermann, R.%
, Ho, C.L.A.%
, Pellat, S.%
, Ormen, Y.%
, Prévost-Solié, C.%
\BDBL {}Huber, D.%
\end{APACrefauthors}%
\unskip\
\newblock
\APACrefYearMonthDay{2020}{}{}.
\newblock
{\BBOQ}\APACrefatitle {EthoLoop: automated closed-loop neuroethology in
  naturalistic environments} {Etholoop: automated closed-loop neuroethology in
  naturalistic environments}.{\BBCQ}
\newblock
\APACjournalVolNumPages{Nat. Methods}{17}{}{1052--1059,}
\newblock

\newblock

\PrintBackRefs{\CurrentBib}

\bibitem [\protect \citeauthoryear {%
Papadopoulou%
, Hildenbrandt%
, Sankey%
, Portugal%
\BCBL {}\ \BBA {} Hemelrijk%
}{%
Papadopoulou%
\ \protect \BOthers {.}}{%
{\protect \APACyear {2022}}%
}]{%
papadopoulou2022self}
\APACinsertmetastar {%
papadopoulou2022self}%
\begin{APACrefauthors}%
Papadopoulou, M.%
, Hildenbrandt, H.%
, Sankey, D.W.%
, Portugal, S.J.%
\BCBL {} Hemelrijk, C.K.%
\end{APACrefauthors}%
\unskip\
\newblock
\APACrefYearMonthDay{2022}{}{}.
\newblock
{\BBOQ}\APACrefatitle {Self-organization of collective escape in pigeon flocks}
  {Self-organization of collective escape in pigeon flocks}.{\BBCQ}
\newblock
\APACjournalVolNumPages{PLoS computational biology}{18}{1}{e1009772,}
\newblock

\newblock

\PrintBackRefs{\CurrentBib}

\bibitem [\protect \citeauthoryear {%
Paszke%
\ \protect \BOthers {.}}{%
Paszke%
\ \protect \BOthers {.}}{%
{\protect \APACyear {2019}}%
}]{%
pytorch}
\APACinsertmetastar {%
pytorch}%
\begin{APACrefauthors}%
Paszke, A.%
, Gross, S.%
, Massa, F.%
, Lerer, A.%
, Bradbury, J.%
, Chanan, G.%
\BDBL {}Chintala, S.%
\end{APACrefauthors}%
\unskip\
\newblock
\APACrefYearMonthDay{2019}{}{}.
\newblock
{\BBOQ}\APACrefatitle {PyTorch: An Imperative Style, High-Performance Deep
  Learning Library} {Pytorch: An imperative style, high-performance deep
  learning library}.{\BBCQ}
\newblock
 \APACrefbtitle {Adv. Neural Inform. Process. Syst.} {Adv. neural inform.
  process. syst.}
\PrintBackRefs{\CurrentBib}

\bibitem [\protect \citeauthoryear {%
Pedersen%
, Haurum%
, Bengtson%
\BCBL {}\ \BBA {} Moeslund%
}{%
Pedersen%
\ \protect \BOthers {.}}{%
{\protect \APACyear {2020}}%
}]{%
Pedersen20203d}
\APACinsertmetastar {%
Pedersen20203d}%
\begin{APACrefauthors}%
Pedersen, M.%
, Haurum, J.B.%
, Bengtson, S.H.%
\BCBL {} Moeslund, T.B.%
\end{APACrefauthors}%
\unskip\
\newblock
\APACrefYearMonthDay{2020}{}{}.
\newblock
{\BBOQ}\APACrefatitle {3D-ZeF: A 3D Zebrafish Tracking Benchmark Dataset}
  {3d-zef: A 3d zebrafish tracking benchmark dataset}.{\BBCQ}
\newblock
 \APACrefbtitle {IEEE Conf. Comput. Vis. Pattern Recog.} {Ieee conf. comput.
  vis. pattern recog.}
\PrintBackRefs{\CurrentBib}

\bibitem [\protect \citeauthoryear {%
Pereira%
\ \protect \BOthers {.}}{%
Pereira%
\ \protect \BOthers {.}}{%
{\protect \APACyear {2019}}%
}]{%
LEAP}
\APACinsertmetastar {%
LEAP}%
\begin{APACrefauthors}%
Pereira, T.D.%
, Aldarondo, D.E.%
, Willmore, L.%
, Kislin, M.%
, Wang, S.S\BHBI H.%
, Murthy, M.%
\BCBL {} Shaevitz, J.W.%
\end{APACrefauthors}%
\unskip\
\newblock
\APACrefYearMonthDay{2019}{}{}.
\newblock
{\BBOQ}\APACrefatitle {Fast animal pose estimation using deep neural networks}
  {Fast animal pose estimation using deep neural networks}.{\BBCQ}
\newblock
\APACjournalVolNumPages{Nat. Methods}{16}{}{117--125,}
\newblock

\newblock

\PrintBackRefs{\CurrentBib}

\bibitem [\protect \citeauthoryear {%
Pereira%
\ \protect \BOthers {.}}{%
Pereira%
\ \protect \BOthers {.}}{%
{\protect \APACyear {2022}}%
}]{%
SLEAP}
\APACinsertmetastar {%
SLEAP}%
\begin{APACrefauthors}%
Pereira, T.D.%
, Tabris, N.%
, Matsliah, A.%
, Turner, D.M.%
, Li, J.%
, Ravindranath, S.%
\BDBL {}Murthy, M.%
\end{APACrefauthors}%
\unskip\
\newblock
\APACrefYearMonthDay{2022}{}{}.
\newblock
{\BBOQ}\APACrefatitle {SLEAP: A deep learning system for multi-animal pose
  tracking} {Sleap: A deep learning system for multi-animal pose
  tracking}.{\BBCQ}
\newblock
\APACjournalVolNumPages{Nat. Methods}{19}{}{486–495,}
\newblock

\newblock

\PrintBackRefs{\CurrentBib}

\bibitem [\protect \citeauthoryear {%
Ristani%
, Solera%
, Zou%
, Cucchiara%
\BCBL {}\ \BBA {} Tomasi%
}{%
Ristani%
\ \protect \BOthers {.}}{%
{\protect \APACyear {2016}}%
}]{%
ristani2016performance}
\APACinsertmetastar {%
ristani2016performance}%
\begin{APACrefauthors}%
Ristani, E.%
, Solera, F.%
, Zou, R.%
, Cucchiara, R.%
\BCBL {} Tomasi, C.%
\end{APACrefauthors}%
\unskip\
\newblock
\APACrefYearMonthDay{2016}{}{}.
\newblock
{\BBOQ}\APACrefatitle {Performance measures and a data set for multi-target,
  multi-camera tracking} {Performance measures and a data set for multi-target,
  multi-camera tracking}.{\BBCQ}
\newblock
 \APACrefbtitle {Eur. Conf. Comput. Vis.} {Eur. conf. comput. vis.}\ (\BPGS\
  17--35).
\PrintBackRefs{\CurrentBib}

\bibitem [\protect \citeauthoryear {%
Romero-Ferrero%
, Bergomi%
, Hinz%
, Heras%
\BCBL {}\ \BBA {} de Polavieja%
}{%
Romero-Ferrero%
\ \protect \BOthers {.}}{%
{\protect \APACyear {2019}}%
}]{%
Paperidtrackerai}
\APACinsertmetastar {%
Paperidtrackerai}%
\begin{APACrefauthors}%
Romero-Ferrero, F.%
, Bergomi, M.G.%
, Hinz, R.C.%
, Heras, F.J.H.%
\BCBL {} de Polavieja, G.G.%
\end{APACrefauthors}%
\unskip\
\newblock
\APACrefYearMonthDay{2019}{}{}.
\newblock
{\BBOQ}\APACrefatitle {idtracker.ai: tracking all individuals in small or large
  collectives of unmarked animals} {idtracker.ai: tracking all individuals in
  small or large collectives of unmarked animals}.{\BBCQ}
\newblock
\APACjournalVolNumPages{Nat. Methods}{16}{}{179--182,}
\newblock

\newblock

\PrintBackRefs{\CurrentBib}

\bibitem [\protect \citeauthoryear {%
Sanakoyeu%
, Khalidov%
, McCarthy%
, Vedaldi%
\BCBL {}\ \BBA {} Neverova%
}{%
Sanakoyeu%
\ \protect \BOthers {.}}{%
{\protect \APACyear {2020}}%
}]{%
Sanakoyeu_2020_CVPR}
\APACinsertmetastar {%
Sanakoyeu_2020_CVPR}%
\begin{APACrefauthors}%
Sanakoyeu, A.%
, Khalidov, V.%
, McCarthy, M.S.%
, Vedaldi, A.%
\BCBL {} Neverova, N.%
\end{APACrefauthors}%
\unskip\
\newblock
\APACrefYearMonthDay{2020}{June}{}.
\newblock
{\BBOQ}\APACrefatitle {Transferring Dense Pose to Proximal Animal Classes}
  {Transferring dense pose to proximal animal classes}.{\BBCQ}
\newblock
 \APACrefbtitle {Proceedings of the IEEE/CVF Conference on Computer Vision and
  Pattern Recognition (CVPR).} {Proceedings of the ieee/cvf conference on
  computer vision and pattern recognition (cvpr).}
\PrintBackRefs{\CurrentBib}

\bibitem [\protect \citeauthoryear {%
Sasaki%
\ \BBA {} Biro%
}{%
Sasaki%
\ \BBA {} Biro%
}{%
{\protect \APACyear {2017}}%
}]{%
sasaki2017cumulative}
\APACinsertmetastar {%
sasaki2017cumulative}%
\begin{APACrefauthors}%
Sasaki, T.%
\BCBT {}\ \BBA {} Biro, D.%
\end{APACrefauthors}%
\unskip\
\newblock
\APACrefYearMonthDay{2017}{}{}.
\newblock
{\BBOQ}\APACrefatitle {Cumulative culture can emerge from collective
  intelligence in animal groups} {Cumulative culture can emerge from collective
  intelligence in animal groups}.{\BBCQ}
\newblock
\APACjournalVolNumPages{Nature communications}{8}{1}{15049,}
\newblock

\newblock

\PrintBackRefs{\CurrentBib}

\bibitem [\protect \citeauthoryear {%
Sun%
\ \protect \BOthers {.}}{%
Sun%
\ \protect \BOthers {.}}{%
{\protect \APACyear {2023}}%
}]{%
BKinD-3D}
\APACinsertmetastar {%
BKinD-3D}%
\begin{APACrefauthors}%
Sun, J.J.%
, Karashchuk, L.%
, Dravid, A.%
, Ryou, S.%
, Fereidooni, S.%
, Tuthill, J.C.%
\BDBL {}others%
\end{APACrefauthors}%
\unskip\
\newblock
\APACrefYearMonthDay{2023}{}{}.
\newblock
{\BBOQ}\APACrefatitle {BKinD-3D: Self-Supervised 3D Keypoint Discovery from
  Multi-View Videos} {Bkind-3d: Self-supervised 3d keypoint discovery from
  multi-view videos}.{\BBCQ}
\newblock
 \APACrefbtitle {IEEE Conf. Comput. Vis. Pattern Recog.} {Ieee conf. comput.
  vis. pattern recog.}\ (\BPGS\ 9001--9010).
\PrintBackRefs{\CurrentBib}

\bibitem [\protect \citeauthoryear {%
Van~Horn%
\ \protect \BOthers {.}}{%
Van~Horn%
\ \protect \BOthers {.}}{%
{\protect \APACyear {2015}}%
}]{%
NAbirds}
\APACinsertmetastar {%
NAbirds}%
\begin{APACrefauthors}%
Van~Horn, G.%
, Branson, S.%
, Farrell, R.%
, Haber, S.%
, Barry, J.%
, Ipeirotis, P.%
\BDBL {}Belongie, S.%
\end{APACrefauthors}%
\unskip\
\newblock
\APACrefYearMonthDay{2015}{}{}.
\newblock
{\BBOQ}\APACrefatitle {Building a Bird Recognition App and Large Scale Dataset
  With Citizen Scientists: The Fine Print in Fine-Grained Dataset Collection}
  {Building a bird recognition app and large scale dataset with citizen
  scientists: The fine print in fine-grained dataset collection}.{\BBCQ}
\newblock
 \APACrefbtitle {IEEE Conf. Comput. Vis. Pattern Recog.} {Ieee conf. comput.
  vis. pattern recog.}
\PrintBackRefs{\CurrentBib}

\bibitem [\protect \citeauthoryear {%
Waldmann%
, Bamberger%
, Johannsen%
, Deussen%
\BCBL {}\ \BBA {} Goldl\"{u}cke%
}{%
Waldmann%
, Bamberger%
\BCBL {}\ \protect \BOthers {.}}{%
{\protect \APACyear {2022}}%
}]{%
unlabprop}
\APACinsertmetastar {%
unlabprop}%
\begin{APACrefauthors}%
Waldmann, U.%
, Bamberger, J.%
, Johannsen, O.%
, Deussen, O.%
\BCBL {} Goldl\"{u}cke, B.%
\end{APACrefauthors}%
\unskip\
\newblock
\APACrefYearMonthDay{2022}{}{}.
\newblock
{\BBOQ}\APACrefatitle {Improving Unsupervised Label Propagation for Pose
  Tracking and Video Object Segmentation} {Improving unsupervised label
  propagation for pose tracking and video object segmentation}.{\BBCQ}
\newblock
 \APACrefbtitle {DAGM German Conference on Pattern Recognition} {Dagm german
  conference on pattern recognition}\ (\BPGS\ 230--245).
\PrintBackRefs{\CurrentBib}

\bibitem [\protect \citeauthoryear {%
Waldmann%
, Johannsen%
\BCBL {}\ \BBA {} Goldluecke%
}{%
Waldmann%
\ \protect \BOthers {.}}{%
{\protect \APACyear {2023}}%
}]{%
NeTePu}
\APACinsertmetastar {%
NeTePu}%
\begin{APACrefauthors}%
Waldmann, U.%
, Johannsen, O.%
\BCBL {} Goldluecke, B.%
\end{APACrefauthors}%
\unskip\
\newblock
\APACrefYearMonthDay{2023}{}{}.
\newblock
{\BBOQ}\APACrefatitle {Neural Texture Puppeteer: A Framework for Neural
  Geometry and Texture Rendering of Articulated Shapes, Enabling
  Re-Identification at Interactive Speed} {Neural texture puppeteer: A
  framework for neural geometry and texture rendering of articulated shapes,
  enabling re-identification at interactive speed}.{\BBCQ}
\newblock
\APACjournalVolNumPages{arXiv preprint arXiv:2311.17109}{}{}{,}
\newblock

\newblock

\PrintBackRefs{\CurrentBib}

\bibitem [\protect \citeauthoryear {%
Waldmann%
, Naik%
\BCBL {}\ \protect \BOthers {.}}{%
Waldmann%
, Naik%
\BCBL {}\ \protect \BOthers {.}}{%
{\protect \APACyear {2022}}%
}]{%
I-MuPPET}
\APACinsertmetastar {%
I-MuPPET}%
\begin{APACrefauthors}%
Waldmann, U.%
, Naik, H.%
, M{\'a}t{\'e}, N.%
, Kano, F.%
, Couzin, I.D.%
, Deussen, O.%
\BCBL {} Goldl{\"u}cke, B.%
\end{APACrefauthors}%
\unskip\
\newblock
\APACrefYearMonthDay{2022}{}{}.
\newblock
{\BBOQ}\APACrefatitle {I-MuPPET: Interactive multi-pigeon pose estimation and
  tracking} {I-muppet: Interactive multi-pigeon pose estimation and
  tracking}.{\BBCQ}
\newblock
 \APACrefbtitle {DAGM German Conference on Pattern Recognition} {Dagm german
  conference on pattern recognition}\ (\BPGS\ 513--528).
\PrintBackRefs{\CurrentBib}

\bibitem [\protect \citeauthoryear {%
Walter%
\ \BBA {} Couzin%
}{%
Walter%
\ \BBA {} Couzin%
}{%
{\protect \APACyear {2021}}%
}]{%
TRex}
\APACinsertmetastar {%
TRex}%
\begin{APACrefauthors}%
Walter, T.%
\BCBT {}\ \BBA {} Couzin, I.D.%
\end{APACrefauthors}%
\unskip\
\newblock
\APACrefYearMonthDay{2021}{}{}.
\newblock
{\BBOQ}\APACrefatitle {TRex, a fast multi-animal tracking system with
  markerless identification, and 2D estimation of posture and visual fields}
  {Trex, a fast multi-animal tracking system with markerless identification,
  and 2d estimation of posture and visual fields}.{\BBCQ}
\newblock
\APACjournalVolNumPages{eLife}{10}{}{e64000,}
\newblock

\newblock

\PrintBackRefs{\CurrentBib}

\bibitem [\protect \citeauthoryear {%
J.~Wang%
\ \BBA {} Yuille%
}{%
J.~Wang%
\ \BBA {} Yuille%
}{%
{\protect \APACyear {2015}}%
}]{%
wang2015se}
\APACinsertmetastar {%
wang2015se}%
\begin{APACrefauthors}%
Wang, J.%
\BCBT {}\ \BBA {} Yuille, A.L.%
\end{APACrefauthors}%
\unskip\
\newblock
\APACrefYearMonthDay{2015}{}{}.
\newblock
{\BBOQ}\APACrefatitle {Semantic Part Segmentation Using Compositional Model
  Combining Shape and Appearance} {Semantic part segmentation using
  compositional model combining shape and appearance}.{\BBCQ}
\newblock
 \APACrefbtitle {IEEE Conf. Comput. Vis. Pattern Recog.} {Ieee conf. comput.
  vis. pattern recog.}
\PrintBackRefs{\CurrentBib}

\bibitem [\protect \citeauthoryear {%
P.~Wang%
\ \protect \BOthers {.}}{%
P.~Wang%
\ \protect \BOthers {.}}{%
{\protect \APACyear {2015}}%
}]{%
wang2015jo}
\APACinsertmetastar {%
wang2015jo}%
\begin{APACrefauthors}%
Wang, P.%
, Shen, X.%
, Lin, Z.%
, Cohen, S.%
, Price, B.%
\BCBL {} Yuille, A.L.%
\end{APACrefauthors}%
\unskip\
\newblock
\APACrefYearMonthDay{2015}{}{}.
\newblock
{\BBOQ}\APACrefatitle {Joint Object and Part Segmentation Using Deep Learned
  Potentials} {Joint object and part segmentation using deep learned
  potentials}.{\BBCQ}
\newblock
 \APACrefbtitle {Int. Conf. Comput. Vis.} {Int. conf. comput. vis.}
\PrintBackRefs{\CurrentBib}

\bibitem [\protect \citeauthoryear {%
Welinder%
\ \protect \BOthers {.}}{%
Welinder%
\ \protect \BOthers {.}}{%
{\protect \APACyear {2010}}%
}]{%
CaltechBirdDataset}
\APACinsertmetastar {%
CaltechBirdDataset}%
\begin{APACrefauthors}%
Welinder, P.%
, Branson, S.%
, Mita, T.%
, Wah, C.%
, Schroff, F.%
, Belongie, S.%
\BCBL {} Perona, P.%
\end{APACrefauthors}%
\unskip\
\newblock
\APACrefYearMonthDay{2010}{}{}.
\newblock
\APACrefbtitle {{Caltech-UCSD Birds 200}} {{Caltech-UCSD Birds 200}}\
  \APACbVolEdTR{}{\BTR{}\ \BNUM\ CNS-TR-2010-001}.
\newblock
\APACaddressInstitution{}{California Institute of Technology}.
\PrintBackRefs{\CurrentBib}

\bibitem [\protect \citeauthoryear {%
Wojke%
\ \BBA {} Bewley%
}{%
Wojke%
\ \BBA {} Bewley%
}{%
{\protect \APACyear {2018}}%
}]{%
Wojke2018deep}
\APACinsertmetastar {%
Wojke2018deep}%
\begin{APACrefauthors}%
Wojke, N.%
\BCBT {}\ \BBA {} Bewley, A.%
\end{APACrefauthors}%
\unskip\
\newblock
\APACrefYearMonthDay{2018}{}{}.
\newblock
{\BBOQ}\APACrefatitle {Deep Cosine Metric Learning for Person
  Re-identification} {Deep cosine metric learning for person
  re-identification}.{\BBCQ}
\newblock
 \APACrefbtitle {2018 IEEE Winter Conference on Applications of Computer Vision
  (WACV)} {2018 ieee winter conference on applications of computer vision
  (wacv)}\ (\BPGS\ 748--756).
\PrintBackRefs{\CurrentBib}

\bibitem [\protect \citeauthoryear {%
Xiao%
, Wu%
\BCBL {}\ \BBA {} Wei%
}{%
Xiao%
\ \protect \BOthers {.}}{%
{\protect \APACyear {2018}}%
}]{%
xiao2018si}
\APACinsertmetastar {%
xiao2018si}%
\begin{APACrefauthors}%
Xiao, B.%
, Wu, H.%
\BCBL {} Wei, Y.%
\end{APACrefauthors}%
\unskip\
\newblock
\APACrefYearMonthDay{2018}{}{}.
\newblock
{\BBOQ}\APACrefatitle {Simple Baselines for Human Pose Estimation and Tracking}
  {Simple baselines for human pose estimation and tracking}.{\BBCQ}
\newblock
 \APACrefbtitle {Eur. Conf. Comput. Vis.} {Eur. conf. comput. vis.}
\PrintBackRefs{\CurrentBib}

\bibitem [\protect \citeauthoryear {%
Xu%
, Zhang%
, Zhang%
\BCBL {}\ \BBA {} Tao%
}{%
Xu%
\ \protect \BOthers {.}}{%
{\protect \APACyear {2022}}%
}]{%
ViTPose}
\APACinsertmetastar {%
ViTPose}%
\begin{APACrefauthors}%
Xu, Y.%
, Zhang, J.%
, Zhang, Q.%
\BCBL {} Tao, D.%
\end{APACrefauthors}%
\unskip\
\newblock
\APACrefYearMonthDay{2022}{}{}.
\newblock
{\BBOQ}\APACrefatitle {Vi{TP}ose: Simple Vision Transformer Baselines for Human
  Pose Estimation} {Vi{TP}ose: Simple vision transformer baselines for human
  pose estimation}.{\BBCQ}
\newblock
 \APACrefbtitle {Advances in Neural Information Processing Systems.} {Advances
  in neural information processing systems.}
\PrintBackRefs{\CurrentBib}

\bibitem [\protect \citeauthoryear {%
J.~Yang%
\ \protect \BOthers {.}}{%
J.~Yang%
\ \protect \BOthers {.}}{%
{\protect \APACyear {2023}}%
}]{%
yang2023track}
\APACinsertmetastar {%
yang2023track}%
\begin{APACrefauthors}%
Yang, J.%
, Gao, M.%
, Li, Z.%
, Gao, S.%
, Wang, F.%
\BCBL {} Zheng, F.%
\end{APACrefauthors}%
\unskip\
\newblock
\APACrefYearMonthDay{2023}{}{}.
\newblock
{\BBOQ}\APACrefatitle {Track anything: Segment anything meets videos} {Track
  anything: Segment anything meets videos}.{\BBCQ}
\newblock
\APACjournalVolNumPages{arXiv preprint arXiv:2304.11968}{}{}{,}
\newblock

\newblock

\PrintBackRefs{\CurrentBib}

\bibitem [\protect \citeauthoryear {%
Y.~Yang%
\ \BBA {} Ramanan%
}{%
Y.~Yang%
\ \BBA {} Ramanan%
}{%
{\protect \APACyear {2013}}%
}]{%
yang2013ar}
\APACinsertmetastar {%
yang2013ar}%
\begin{APACrefauthors}%
Yang, Y.%
\BCBT {}\ \BBA {} Ramanan, D.%
\end{APACrefauthors}%
\unskip\
\newblock
\APACrefYearMonthDay{2013}{}{}.
\newblock
{\BBOQ}\APACrefatitle {Articulated Human Detection with Flexible Mixtures of
  Parts} {Articulated human detection with flexible mixtures of parts}.{\BBCQ}
\newblock
\APACjournalVolNumPages{IEEE Trans. Pattern Anal. Mach.
  Intell.}{35}{12}{2878-2890,}
\newblock

\newblock

\PrintBackRefs{\CurrentBib}

\bibitem [\protect \citeauthoryear {%
Yomosa%
, Mizuguchi%
, V{\'a}s{\'a}rhelyi%
\BCBL {}\ \BBA {} Nagy%
}{%
Yomosa%
\ \protect \BOthers {.}}{%
{\protect \APACyear {2015}}%
}]{%
yomosa2015coordinated}
\APACinsertmetastar {%
yomosa2015coordinated}%
\begin{APACrefauthors}%
Yomosa, M.%
, Mizuguchi, T.%
, V{\'a}s{\'a}rhelyi, G.%
\BCBL {} Nagy, M.%
\end{APACrefauthors}%
\unskip\
\newblock
\APACrefYearMonthDay{2015}{}{}.
\newblock
{\BBOQ}\APACrefatitle {Coordinated behaviour in pigeon flocks} {Coordinated
  behaviour in pigeon flocks}.{\BBCQ}
\newblock
\APACjournalVolNumPages{Plos one}{10}{10}{e0140558,}
\newblock

\newblock

\PrintBackRefs{\CurrentBib}

\bibitem [\protect \citeauthoryear {%
Zhang%
, Gao%
, Xiao%
\BCBL {}\ \BBA {} Fan%
}{%
Zhang%
\ \protect \BOthers {.}}{%
{\protect \APACyear {2023}}%
}]{%
AnimalTrack}
\APACinsertmetastar {%
AnimalTrack}%
\begin{APACrefauthors}%
Zhang, L.%
, Gao, J.%
, Xiao, Z.%
\BCBL {} Fan, H.%
\end{APACrefauthors}%
\unskip\
\newblock
\APACrefYearMonthDay{2023}{}{}.
\newblock
{\BBOQ}\APACrefatitle {AnimalTrack: A Benchmark for Multi-Animal Tracking in
  the Wild} {Animaltrack: A benchmark for multi-animal tracking in the
  wild}.{\BBCQ}
\newblock
\APACjournalVolNumPages{International Journal of Computer
  Vision}{131}{2}{496--513,}
\newblock

\newblock

\PrintBackRefs{\CurrentBib}

\bibitem [\protect \citeauthoryear {%
Zuffi%
\ \protect \BOthers {.}}{%
Zuffi%
\ \protect \BOthers {.}}{%
{\protect \APACyear {2023}}%
}]{%
CV4An2023}
\APACinsertmetastar {%
CV4An2023}%
\begin{APACrefauthors}%
Zuffi, S.%
, Rhodin, H.%
, Park, H.S.%
, Beery, S.%
, Kanazawa, A.%
, Nobuhara, S.%
\BCBL {} Zamansky, A.%
\end{APACrefauthors}%
\unskip\
\newblock
\APACrefYearMonthDay{2023}{}{}.
\newblock
\APACrefbtitle {CV4Animals: Computer Vision for Animal Behavior Tracking and
  Modeling.} {Cv4animals: Computer vision for animal behavior tracking and
  modeling.}
\newblock
\APACrefnote{\url{https://www.cv4animals.com/}}
\PrintBackRefs{\CurrentBib}

\end{thebibliography}

\end{document}